\ifdefined\pdftexversion
  \pdfoutput=1
\fi
\documentclass[10pt, logo, onecolumn, copyright]{nv}

\usepackage{times}

\usepackage[square,sort,comma,numbers]{natbib}
\titlespacing*{\paragraph}{0pt}{0.6ex plus 0.5ex minus 0.2ex}{0.8em}
\usepackage{subcaption}
\usepackage{float}
\usepackage{amsmath,amssymb}
\usepackage{booktabs,tabularx,colortbl}
\usepackage{graphicx}
\usepackage{adjustbox}
\usepackage{xspace}
\usepackage{enumitem}
\usepackage{pifont}
\usepackage{tikz}
\usetikzlibrary{arrows.meta,positioning,calc,fit,backgrounds,shapes.misc,decorations.pathreplacing,shadows,shadings,patterns}

\definecolor{nvidiagreen}{HTML}{76B900}
\definecolor{nvgreen}{HTML}{76B900}
\definecolor{nvgreendark}{HTML}{3F6A1A}
\definecolor{catgray}{gray}{0.92}

\newcommand{\ours}{SANA-Video~2.5\xspace}

\newcommand{\prev}{SANA-Video~2.0\xspace}
\newcommand{\attnres}{AttnRes\xspace}

\makeatletter\newcommand{\coltabtype}{\def\@captype{table}\renewenvironment{table}[1][]{}{}}\makeatother
\NewDocumentCommand{\coltabpar}{O{0.5}O{0.47}+m+m}{\par\medskip\noindent
  \begin{minipage}[t]{#1\linewidth}\vspace{0pt}\setlength{\parindent}{0pt}\setlength{\parskip}{4pt}#3\end{minipage}\hfill
  \begin{minipage}[t]{#2\linewidth}\vspace{0pt}\centering\coltabtype#4\end{minipage}\par\medskip}
\makeatletter
\DeclareRobustCommand\onedot{\futurelet\@let@token\@onedot}
\def\@onedot{\ifx\@let@token.\else.\null\fi\xspace}

\def\ie{\emph{i.e}\onedot}
\def\vs{\emph{vs}\onedot}

\makeatother

\newcommand{\T}{\mathcal{T}}
\newcommand{\J}{J}

\newcommand{\Hthree}{H3\xspace}
\newcommand{\GN}[1]{GN-#1\xspace}
\newcommand{\takeaway}[1]{\textbf{\textit{#1}}}
\newcommand{\keyres}[1]{\textbf{#1}}
\newcommand{\cmark}{\ding{51}}

\DeclareMathOperator*{\argmin}{arg\,min}

\usepackage{placeins}
\hypersetup{colorlinks,linkcolor=nvgreendark,citecolor=nvgreendark,urlcolor=nvgreendark}
\usepackage[capitalise,nameinlink]{cleveref}
\crefname{section}{Section}{Sections}
\Crefname{section}{Section}{Sections}
\crefname{table}{Table}{Tables}
\Crefname{table}{Table}{Tables}
\crefname{subtable}{Table}{Tables}
\Crefname{subtable}{Table}{Tables}
\crefname{figure}{Figure}{Figures}
\Crefname{figure}{Figure}{Figures}
\crefname{equation}{Eq.}{Eqs.}
\crefname{appendix}{App.}{Apps.}

\title{\ours{}: Scaling Hybrid Linear Attention \mbox{for Efficient Audio--Video Generation}}

\author{%
\parbox{\linewidth}{%
\centering
\setlength{\parskip}{0pt}
\vspace{-2pt}
{\bfseries\fontsize{9.6pt}{13.5pt}\selectfont
Junsong~Chen\textsuperscript{*},\quad Shuchen~Xue\textsuperscript{*},\quad Yitong~Li,\quad Jincheng~Yu,\quad Haopeng~Li,\par
Haozhe~Liu,\quad Jingyu~Xin,\quad Daquan~Zhou,\quad Ping~Luo,\quad Song~Han,\quad Enze~Xie\par}
\vspace{2mm}
{\normalsize NVIDIA \par}
\vspace{0.1em}
{\normalsize
\href{https://nvlabs.github.io/Sana/Video25/}{\raisebox{-0.35em}{\includegraphics[height=1.5em]{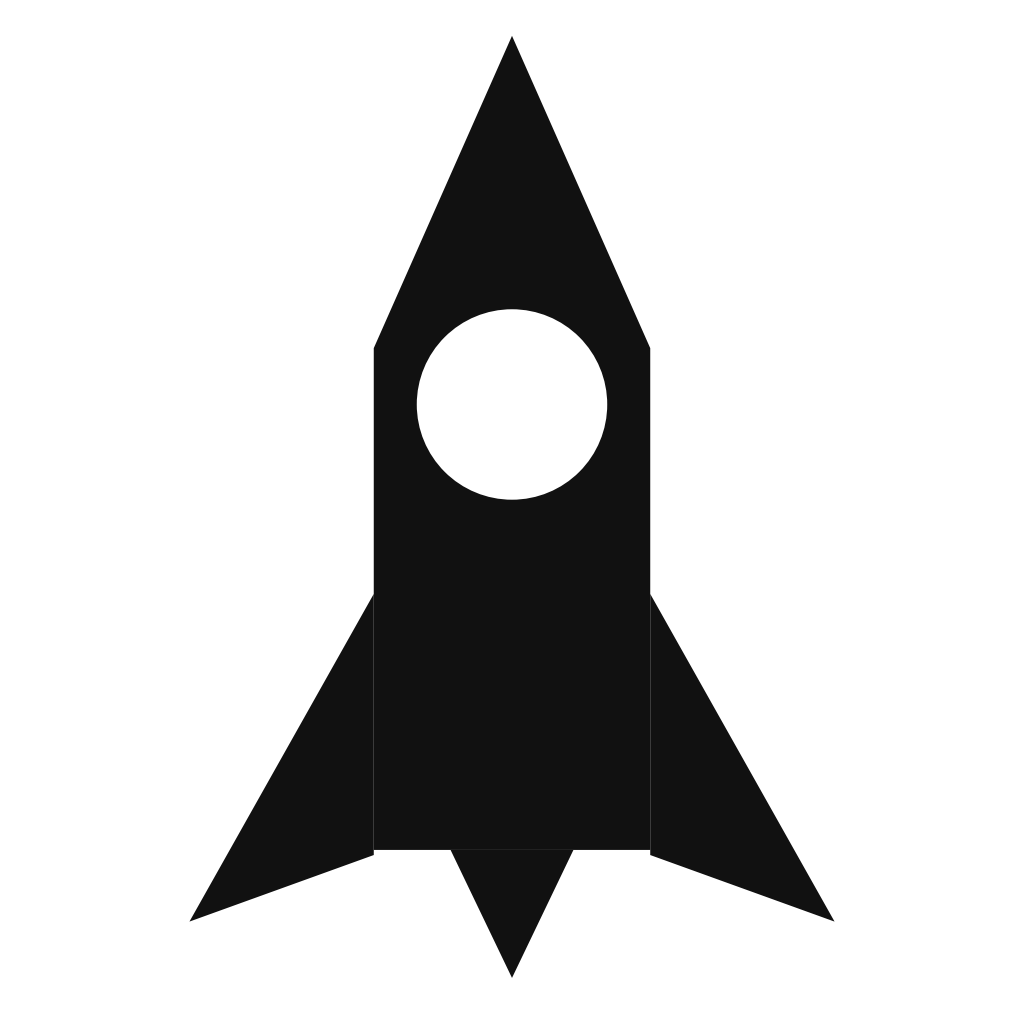}}\kern0.3em Project Page}\quad
\href{https://github.com/NVlabs/Sana}{\raisebox{-0.35em}{\includegraphics[height=1.2em,trim=256 223 256 249,clip]{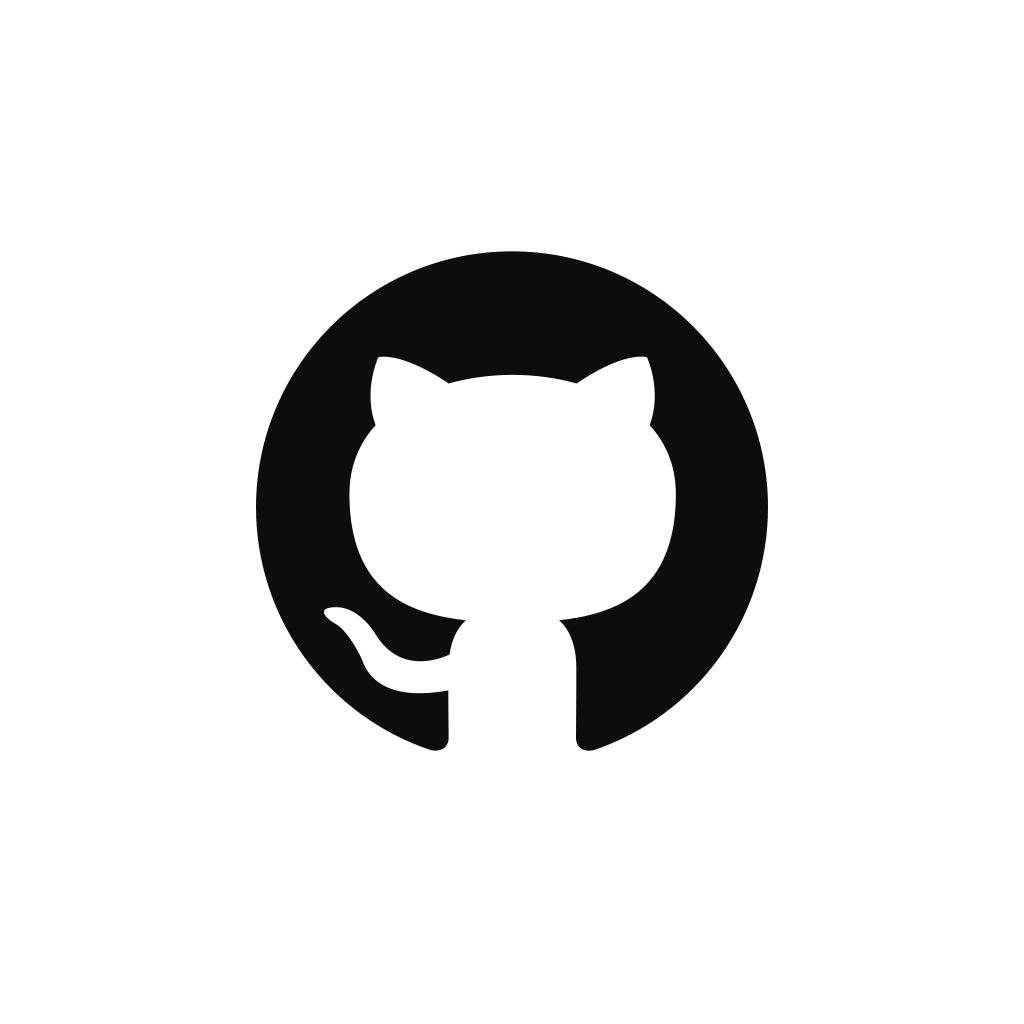}}\kern0.3em GitHub}\par}
}}
\correspondingauthor{Equal contribution.}

\begin{abstract}
We introduce \ours, a 16B model that generates 5--15\,s 768p videos with synchronised sound in seconds on a single GPU. Linear attention keeps a fixed-size state at any length, suiting long-horizon generation. \ours scales the hybrid linear-attention backbone of SANA-Video 2.0 to 16B parameters and 15\,s clips with sound. Core designs include:
(1) \textbf{Parallel audio stream:} a 2B audio stream exchanges information with every layer of a 14B hybrid video model through time-aligned cross-attention.
(2) \textbf{Staged pre-training:} audio is bootstrapped beside the frozen, LoRA-adapted video weights, then all 16B parameters train jointly.
(3) \textbf{Cross-latent few-step distillation:} 4- and 8-step students learn from the stronger 33B MiniMax-H3 in its own latent space, and a Gauss--Newton pullback through a learned adapter raises the realised cosine of the teacher's update from 0.33 to 0.88.
On one GB200, the 4-step student generates a 5\,s 768p clip with audio in \textbf{2.9\,s}, about 45$\times$ faster than the 50-step teacher and 6--7$\times$ faster than few-step H3 models, yet scores higher than both on HPSv3. The 8-step student generates a 15\,s clip in 14.7\,s, faster than 8-step LTX-2.5 distilled (20.8\,s), and in \textbf{9.0\,s} with Sol-Engine, MXFP8 and a tiny VAE decoder. Fine-tuned with read-only reference banks, the model further generates video and sound from multiple reference images, clips and sounds, following H3, with each asset bound to a subject in the prompt.
\end{abstract}

\let\sharedpaperabstract\theabstract
\long\gdef\theabstract{\vspace*{-16pt}\noindent\textbf{Abstract:} \sharedpaperabstract}

\begin{document}
\maketitle
\enlargethispage{2cm}

\begin{center}
\captionsetup{type=figure}
\vspace{-2pt}
\includegraphics[width=\linewidth]{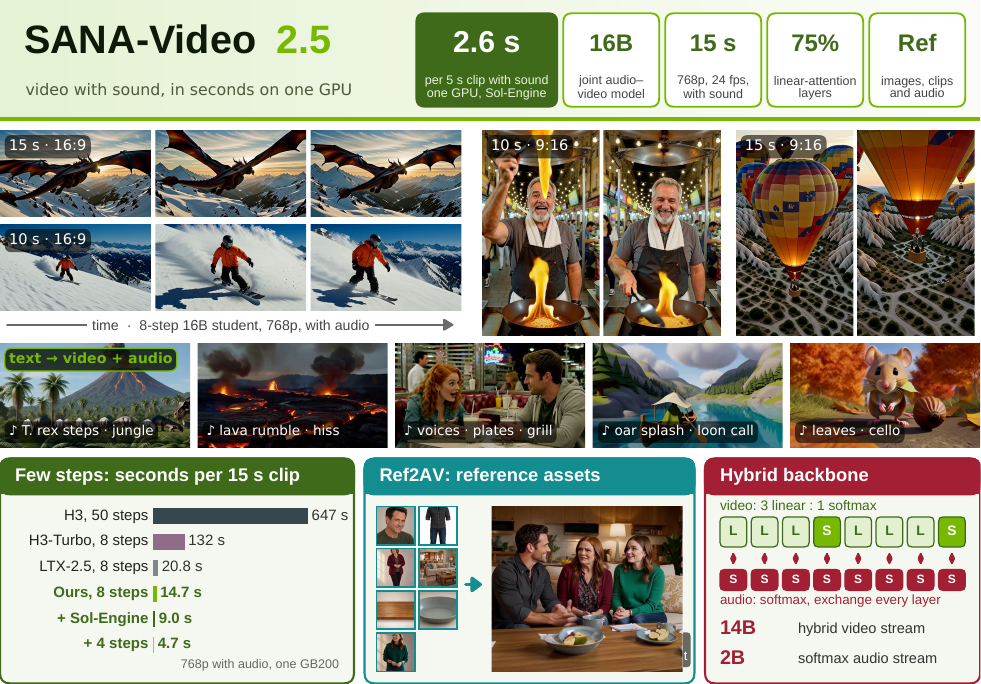}
\caption{\textbf{\ours generates video with synchronised sound from text in seconds on one GPU, and from reference images after fine-tuning.}
Top: 5 to 15\,s clips from one 8-step 16B student. Middle: text-to-audio--video samples, tagged with prompted sounds. Bottom: 15\,s latency with audio on one GB200 (+ Sol-Engine~\cite{solengine}, \cref{sec:engine}), Ref2AV and the hybrid backbone.}
\label{fig:teaser}
\end{center}
\newpage

\section{Introduction}
Video generation is moving from silent clips to audio--visual scenes. Closed systems such as Veo\,3~\cite{wiedemer2025veo3} and open models such as LTX-2~\cite{hacohen2026ltx2}, Ovi~\cite{low2025ovi} and MiniMax-H3~\cite{minimax2026h3} now generate video with synchronised speech, music and sound effects.
This makes generation even more token-hungry. A 15\,s 768p clip already spans over 40{,}000 video latent tokens, the audio track adds its own sequence, and every layer must keep the two in sync. With full softmax attention the cost grows quadratically with duration, and the 33B joint model MiniMax-H3 takes over two minutes for a 5\,s 768p clip with audio on one GB200 at 50 steps.
SANA-Video~2.0~\cite{chen2026sanavideo2} showed that a hybrid linear-attention backbone, 75\% gated linear attention with periodic softmax anchors and block attention residuals (\attnres)~\cite{kimiteam2026attnres}, matches full-softmax video models at a fraction of their cost, but it was trained on silent clips of up to 8\,s and used 40--50 sampling steps.
Linear attention also matters beyond speed: its fixed-size state keeps long-horizon memory far cheaper than a growing softmax cache~\cite{chen2025sanavideo}, which motivates scaling such backbones further, now with sound.
We ask: \emph{how far can a hybrid linear-attention backbone scale towards efficient generation of long clips with synchronised sound?}

We answer with \ours, a 16B joint audio--video model, \ie, a 14B hybrid video stream with a 2B audio stream beside it.
It generates 5--15\,s videos at 768p and 24\,fps with stereo audio from text, first and last frames, and, after fine-tuning, from reference images, clips and sounds (\cref{fig:teaser}).
Its 8-step student generates a 15\,s $1344{\times}768$ clip with audio in 14.7\,s on one GB200, faster than 8-step LTX-2.5 distilled in its fastest tested implementation (20.8\,s), and in 9.0\,s when deployed with Sol-Engine~\cite{solengine}, the 8-bit MXFP8 format~\cite{rouhani2023mx} and a tiny distilled VAE decoder~\cite{boerbohan2025taehv}.
We keep the hybrid backbone's mix of linear and softmax attention layers, so its efficiency carries over, and add three things to it: a parallel audio stream, staged audio--video pre-training, and few-step distillation from a stronger teacher.

\paragraph{A parallel audio stream on a hybrid backbone (\cref{sec:arch}).}
A 40-layer audio stream exchanges information with every video layer through bidirectional cross-attention aligned in physical time.
Among four designs compared with matched data and training steps, it gives the best text--audio agreement and is the only one that leaves the video stream's linear/softmax layout, kernels and RoPE~\cite{su2021roformer} untouched, so the joint model inherits the efficiency of the video model.

\paragraph{Staged audio--video pre-training (\cref{fig:overview,sec:pretrain}).}
We first train the audio stream and a video LoRA with the video backbone frozen on short 16\,fps clips, so audio is learned cheaply without disturbing the video model. We then train all 16B parameters jointly at up to 15\,s and 768p.

\paragraph{Few-step distillation from a stronger teacher (\cref{sec:fewstep}).}
The 40-step model is distilled into 4- and 8-step students by DMD~\cite{yin2023dmd,yin2024dmd2} from MiniMax-H3, a stronger 33B teacher in another VAE latent space, through a learned latent adapter, so students keep their efficient backbone and their own latent space.
On the 5B testbed, the student update obtained by backpropagation maps mostly to a global colour shift, with cosine similarity 0.22 to the teacher's correction. A damped Gauss--Newton pullback solves for a student update whose mapped direction better matches that correction, raising the cosine to 0.83 (0.33 to 0.88 for the 16B model).

\noindent In summary, our contributions are:
\begin{itemize}[leftmargin=*,itemsep=0pt,topsep=2pt]
\item \textbf{To the best of our knowledge, the first hybrid linear-attention model for joint audio--video generation}, scaled to 16B with a time-aligned parallel audio stream chosen by a controlled comparison of four designs.
\item \textbf{A staged recipe} that first trains a 2B audio stream cheaply beside a frozen, LoRA-adapted 14B video model, then trains all 16B parameters jointly with per-stream learning rates.
\item \textbf{Cross-latent few-step distillation} that lets a student learn from a stronger teacher in a foreign VAE latent space, with a fidelity diagnosis of the backpropagated pullback and a matrix-free Gauss--Newton fix, giving students that surpass their 40-step parent on both preference models, HPSv3 and VideoReward, at a fraction of its latency.
\item \textbf{Open release} of the training code and model weights for research on efficient audio--video generation.
\end{itemize}
\noindent As an application, Ref2AV (\cref{sec:ref2av}) follows \Hthree in taking multiple reference images, clips and sounds of people, voices, objects and places, binds them to prompt subjects and generates their video and sound from read-only banks.

\newcommand{\avtabcaption}{\caption{\textbf{Audio--video interaction designs.} Params: audio-side parameters. Quality: 20 prompts, 5\,s at 16\,fps. DS: DeSync (s). Cost: extra sampling time over video only at 15\,s, 24\,fps.}
\label{tab:avarch}}
\newcommand{\avtabbody}{{\fontsize{8.6}{10.4}\selectfont\setlength{\tabcolsep}{2.4pt}%
\begin{tabularx}{\linewidth}{@{}l *{5}{>{\centering\arraybackslash}X}@{}}
\toprule
\textbf{Design} & \textbf{Params} & \textbf{CLAP}$\uparrow$ & \textbf{PQ}$\uparrow$ & \textbf{DS}$\downarrow$ & \textbf{Cost}$\downarrow$ \\
\midrule
(a) Baseline & 2.64B & 0.156 & 5.81 & 1.01 & \textbf{+61\%} \\
\rowcolor{nvidiagreen!15}(b) Cross-attention (ours) & 2.60B & \textbf{0.229} & 5.94 & 0.94 & +68\% \\
(c) Hybrid & 2.56B & 0.193 & \textbf{5.98} & 1.04 & +84\% \\
(d) MoT & 2.15B & 0.180 & 5.85 & \textbf{0.77} & +66\% \\
\bottomrule
\end{tabularx}}}
\newlength{\avrowH}\setlength{\avrowH}{14pt}
\section{Model}
\label{sec:arch}

\paragraph{Video stream.}
The video stream follows SANA-Video~2.0~\cite{chen2026sanavideo2} without modification where it matters for efficiency: gated bidirectional linear attention~\cite{yang2025gdn,yang2025kimi} in three of every four layers and gated softmax anchors in the fourth, block \attnres~\cite{kimiteam2026attnres}, SwiGLU FFNs, and the LTX-2.5 VAE~\cite{hacohen2026ltx2} with $8{\times}32{\times}32$ compression and 128 channels.
We change three things (configuration in \cref{tab:config}).
(1)~\emph{Text \mbox{encoder.}} We use the vision--language model Qwen3-VL-4B~\cite{bai2025qwen3vl}, which reads long prompts describing picture, sound and music separately~\cite{minimax2026h3skill} and reads reference images inline with the prompt for Ref2AV (\cref{sec:ref2av}). Several of its hidden layers are concatenated and projected to the model width. The audio stream projects the same features with its own normalisation and projection, so the two modalities share the encoder but not the conditioning path.
SANA-Video~2.0 conditioned on Gemma-2-2B features truncated to 300 tokens, while we use up to 1024 tokens. The longer context accommodates detailed descriptions of both the scene and its sound.
(2)~\emph{Longer clips.} The model is continued to 480p--768p and 5, 10 and 15\,s at 24\,fps, up to 46k video tokens per clip, with the token-count-aware flow shift extended to the longer sequences.
(3)~\emph{Keyframes.} A first, last or first-and-last frame is encoded separately and enters as a clean prefix at noise level zero, so text-to-video and image-to-video share weights.

\subsection{Audio Stream and Audio--Video Interaction}

\paragraph{Audio latent.}
Audio is encoded by the frozen waveform VAE of MiniMax-H3~\cite{minimax2026h3} (32\,kHz stereo, 40 latents per second, 32 channels per side). The two stereo sides of one time step are stacked into one 64-dimensional token, so a 15\,s clip has only 600 audio tokens, against up to 46k video tokens.

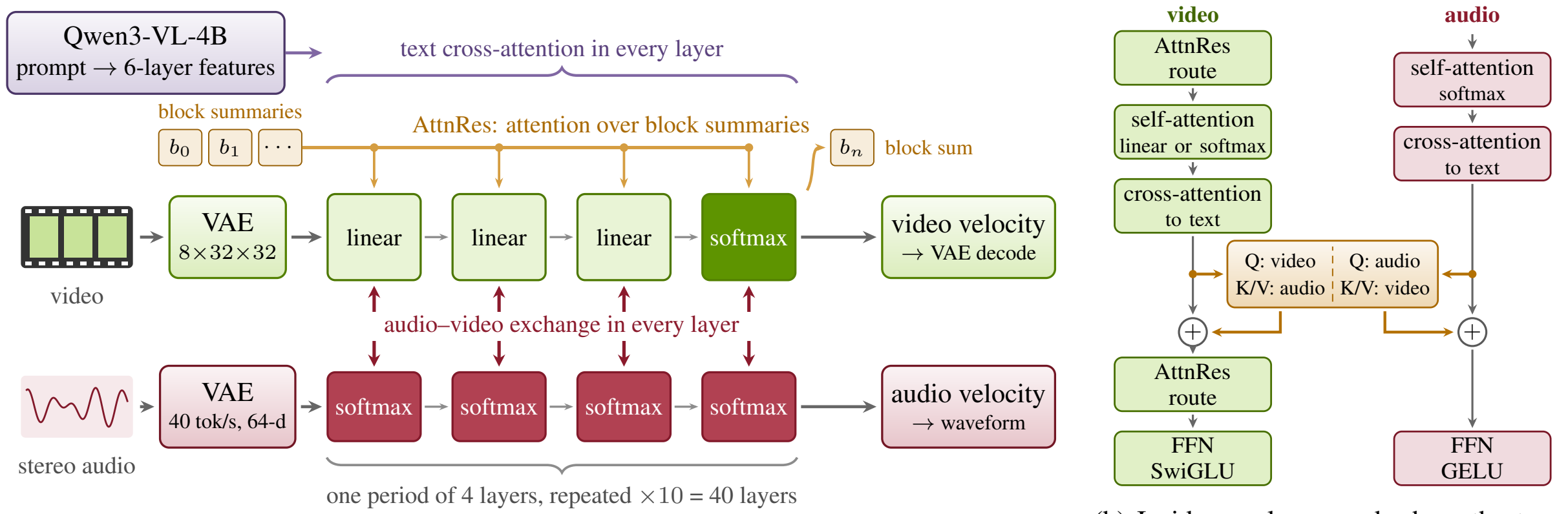
\begin{figure}[t]
\centering
\sbox0{\adjustbox{height=5.2cm}{\definecolor{vidc}{HTML}{76B900}
\definecolor{audc}{HTML}{A31F34}
\definecolor{txtc}{HTML}{6A4C93}
\definecolor{resc}{HTML}{C77C02}
\begin{tikzpicture}[
    font=\small, >={Stealth[length=4.5pt, width=4pt]},
    box/.style={draw=#1!70!black, rounded corners=3pt, top color=#1!5, bottom color=#1!26, line width=0.7pt, align=center, inner sep=3pt},
    lin/.style={draw=vidc!65!black, fill=vidc!16, rounded corners=2.5pt, line width=0.7pt,
                minimum width=1.08cm, minimum height=1.0cm, inner sep=1pt, align=center, font=\footnotesize},
    sm/.style={draw=vidc!70!black, fill=vidc!80!black, rounded corners=2.5pt, line width=0.7pt,
               minimum width=1.08cm, minimum height=1.0cm, inner sep=1pt, align=center, font=\footnotesize, text=white},
    aud/.style={draw=audc!80!black, fill=audc!85, rounded corners=2.5pt, line width=0.7pt,
                minimum width=1.08cm, minimum height=0.8cm, inner sep=1pt, align=center, font=\footnotesize, text=white},
    flow/.style={->, draw=black!60, line width=1.0pt, shorten >=1.5pt, shorten <=1.5pt},
    lbl/.style={font=\footnotesize, text=black!70, align=center},
  ]
  \node[box=txtc, minimum width=2.9cm, minimum height=0.95cm] (qwen) at (0.9,4.55) {Qwen3-VL-4B\\[-1pt]{\footnotesize prompt $\rightarrow$ 6-layer features}};
  \begin{scope}[shift={(-0.55,2.05)}]
    \fill[black!80, rounded corners=1.5pt] (0,0) rectangle (1.3,0.72);
    \foreach \k in {0,1,2} { \fill[vidc!40] (0.1+0.4*\k,0.12) rectangle ++(0.33,0.48); }
    \foreach \k in {0,...,7} { \fill[white] (0.05+0.16*\k,0.035) rectangle ++(0.07,0.045); \fill[white] (0.05+0.16*\k,0.64) rectangle ++(0.07,0.045); }
  \end{scope}
  \node[box=vidc, minimum width=1.25cm, minimum height=0.95cm] (vvae) at (1.85,2.41) {VAE\\[-2pt]{\scriptsize $8{\times}32{\times}32$}};
  \draw[flow] (0.78,2.41) -- (vvae.west);
  \begin{scope}[shift={(-0.55,0.08)}]
    \fill[audc!10, rounded corners=1.5pt] (0,0) rectangle (1.3,0.72);
    \draw[audc!80!black, line width=0.6pt, domain=0.06:1.24, samples=110, smooth] plot (\x, {0.36+0.27*sin(deg(\x*20))*exp(-0.9*(\x-0.62)^2)*cos(deg(\x*3))});
  \end{scope}
  \node[box=audc, minimum width=1.25cm, minimum height=0.95cm] (avae) at (1.85,0.44) {VAE\\[-2pt]{\scriptsize 40 tok/s, 64-d}};
  \draw[flow] (0.78,0.44) -- (avae.west);
  \node[lbl, anchor=north] at (0.1,1.95) {video};
  \node[lbl, anchor=north] at (0.1,-0.02) {stereo audio};
  \foreach \k in {0,...,3} {
    \pgfmathsetmacro{\x}{3.55+1.45*\k}
    \pgfmathtruncatemacro{\soft}{mod(\k,4)==3}
    \ifnum\soft=1
      \node[sm] (V\k) at (\x,2.41) {softmax};
    \else
      \node[lin] (V\k) at (\x,2.41) {linear};
    \fi
    \node[aud] (A\k) at (\x,0.44) {softmax};
    \draw[<->, >={Stealth[length=4pt, width=4pt]}, draw=audc!85!black, line width=1.1pt, shorten >=1.5pt, shorten <=1.5pt] (V\k.south) -- (A\k.north);
  }
  \foreach \k [evaluate=\k as \j using int(\k+1)] in {0,...,2} {
    \draw[->, >={Stealth[length=3pt]}, draw=black!45, line width=0.7pt, shorten >=1pt, shorten <=1pt] (V\k.east) -- (V\j.west);
    \draw[->, >={Stealth[length=3pt]}, draw=black!45, line width=0.7pt, shorten >=1pt, shorten <=1pt] (A\k.east) -- (A\j.west);
  }
  \draw[flow] (vvae.east) -- (V0.west);
  \draw[flow] (avae.east) -- (A0.west);
  \foreach \k in {0,1,2} {
    \node[draw=resc!75!black, fill=resc!14, rounded corners=1.5pt, minimum width=0.5cm, minimum height=0.42cm, inner sep=0pt, font=\scriptsize] (b\k) at (1.3+0.58*\k,3.45) {};
  }
  \node[font=\scriptsize] at (b0) {$b_0$}; \node[font=\scriptsize] at (b1) {$b_1$}; \node[font=\scriptsize] at (b2) {$\cdots$};
  \node[lbl, text=resc!85!black, anchor=south, font=\scriptsize] at (b1.north) {block summaries};
  \draw[draw=resc!75, line width=1.1pt] (b2.east) -- (7.9,3.45);
  \foreach \k in {0,...,3} {
    \draw[->, >={Stealth[length=3.5pt]}, draw=resc!75, line width=0.8pt, shorten >=1.5pt] (V\k.north |- 0,3.45) -- (V\k.north);
    \fill[resc!75] (V\k.north |- 0,3.45) circle (1.4pt);
  }
  \node[draw=resc!75!black, fill=resc!14, rounded corners=1.5pt, minimum width=0.5cm, minimum height=0.42cm, inner sep=0pt, font=\scriptsize] (bn) at (9.1,3.45) {$b_n$};
  \draw[->, >={Stealth[length=3.5pt]}, draw=resc!75, line width=0.9pt, shorten <=1pt, shorten >=1.5pt] ([xshift=3pt]V3.north east) to[out=60, in=180] (bn.west);
  \node[lbl, text=resc!85!black, anchor=west, font=\scriptsize] at (bn.east) {block sum};
  \node[lbl, text=resc!85!black, anchor=south] at (6.3,3.5) {\attnres: attention over block summaries};
  \draw[->, draw=txtc!85, line width=1.0pt] (qwen.east) -- (2.95,4.55);
  \draw[decorate, decoration={brace, amplitude=5pt}, draw=txtc!85, line width=0.9pt] (3.0,4.2) -- (8.46,4.2);
  \node[lbl, text=txtc!85!black, anchor=south] at (5.73,4.33) {text cross-attention in every layer};
  \draw[decorate, decoration={brace, amplitude=5pt, mirror}, draw=black!45, line width=0.8pt] (3.0,-0.15) -- (8.46,-0.15)
        node[midway, below=6pt, lbl] {one period of 4 layers, repeated $\times10$ = 40 layers};
  \node[lbl, text=audc!85!black, font=\footnotesize, fill=white, inner sep=1.2pt, minimum width=4.75cm] at ($(V1.south)!0.5!(A2.north)$) {audio--video exchange in every layer};
  \node[box=vidc, minimum width=2.0cm, minimum height=0.95cm] (vout) at (10.45,2.41) {video velocity\\[-2pt]{\scriptsize $\rightarrow$ VAE decode}};
  \node[box=audc, minimum width=2.0cm, minimum height=0.95cm] (aout) at (10.45,0.44) {audio velocity\\[-2pt]{\scriptsize $\rightarrow$ waveform}};
  \draw[flow] (V3.east) -- (vout.west);
  \draw[flow] (A3.east) -- (aout.west);
\end{tikzpicture}}}%
\sbox2{\adjustbox{height=5.0cm}{\definecolor{audc}{HTML}{A31F34}
\definecolor{xc}{HTML}{C77C02}
\definecolor{vidc}{HTML}{76B900}
\begin{tikzpicture}[
    font=\small, >={Stealth[length=4pt, width=3.6pt]},
    b/.style={rounded corners=3pt, align=center, minimum height=0.74cm, text width=2.05cm, inner sep=1.5pt, line width=0.7pt},
    vb/.style={b, fill=vidc!22, draw=vidc!60!black},
    ab/.style={b, fill=audc!16, draw=audc!75!black},
    flow/.style={->, draw=black!60, line width=0.8pt, shorten >=1pt, shorten <=1pt},
    xf/.style={->, draw=xc!90!black, line width=1.1pt, shorten >=1.5pt, shorten <=1.5pt},
    plus/.style={circle, draw=black!65, line width=0.8pt, inner sep=0.5pt, font=\small},
  ]
  \def\VX{0} \def\AX{3.85} \def\MX{1.925} \def\SY{-2.98} \def\PY{-3.78}
  \node[font=\small\bfseries, text=vidc!55!black] at (\VX,0.6) {video};
  \node[font=\small\bfseries, text=audc!85!black] at (\AX,0.6) {audio};
  \node[vb] (vr)  at (\VX,0)     {\attnres\\[-1pt]route};
  \node[vb] (vs)  at (\VX,-1.02) {self-attention\\[-1pt]{\footnotesize linear or softmax}};
  \node[vb] (vt)  at (\VX,-2.04) {cross-attention\\[-1pt]{\footnotesize to text}};
  \node[plus] (vp) at (\VX,\PY) {$+$};
  \node[vb] (vr2) at (\VX,-4.5) {\attnres\\[-1pt]route};
  \node[vb] (vf)  at (\VX,-5.52) {FFN\\[-1pt]SwiGLU};
  \draw[flow] (vr) -- (vs); \draw[flow] (vs) -- (vt); \draw[flow] (vt) -- (vp); \draw[flow] (vp) -- (vr2); \draw[flow] (vr2) -- (vf);
  \node[ab] (as) at (\AX,-0.3) {self-attention\\[-1pt]{\footnotesize softmax}};
  \node[ab] (at) at (\AX,-1.32) {cross-attention\\[-1pt]{\footnotesize to text}};
  \node[plus] (ap) at (\AX,\PY) {$+$};
  \node[ab] (af) at (\AX,-5.52) {FFN\\[-1pt]GELU};
  \draw[flow] ([yshift=0.28cm]as.north) -- (as.north);
  \draw[flow] (as) -- (at); \draw[flow] (at) -- (ap); \draw[flow] (ap) -- (af);
  \fill[rounded corners=3pt, top color=xc!6, bottom color=xc!30] (\MX-1.45,\SY-0.46) rectangle (\MX+1.45,\SY+0.46);
  \draw[rounded corners=3pt, draw=xc!85!black, line width=0.7pt] (\MX-1.45,\SY-0.46) rectangle (\MX+1.45,\SY+0.46);
  \draw[draw=xc!85!black, line width=0.5pt, densely dashed] (\MX,\SY-0.38) -- (\MX,\SY+0.38);
  \node[align=center, font=\footnotesize, inner sep=0pt] (xv) at (\MX-0.72,\SY) {Q: video\\[1pt]K/V: audio};
  \node[align=center, font=\footnotesize, inner sep=0pt] (xa) at (\MX+0.72,\SY) {Q: audio\\[1pt]K/V: video};
  \fill[xc!90!black] (\VX,\SY) circle (1.6pt);
  \fill[xc!90!black] (\AX,\SY) circle (1.6pt);
  \draw[xf] (\VX,\SY) -- (\MX-1.45,\SY);
  \draw[xf] (\AX,\SY) -- (\MX+1.45,\SY);
  \draw[xf] (\MX-0.72,\SY-0.46) |- (vp.east);
  \draw[xf] (\MX+0.72,\SY-0.46) |- (ap.west);
\end{tikzpicture}}}%
\begin{subfigure}[b]{\wd0}
\centering\usebox0
\caption{Full model: one 4-layer period of the hybrid video stream beside the audio stream.}
\end{subfigure}\hspace{0.9em}%
\begin{subfigure}[b]{\wd2}
\centering\usebox2
\caption{Inside one layer, and where the two cross-modal attentions read and write.}
\end{subfigure}
\caption{\textbf{A parallel audio stream exchanges information with every layer of the hybrid video backbone.} The video layers, kernels and 3D RoPE are unchanged. Cross-modal RoPE uses time in seconds, so audio and video tokens of one instant share a phase.}
\label{fig:avblock}
\vspace{-1em}
\end{figure}

\paragraph{A parallel stream with per-layer exchange.}
The audio stream is a 40-layer transformer of width 1408 (2B parameters), one audio layer per video layer (\cref{fig:avblock}).
Since audio tokens are few, every audio layer uses dense softmax self-attention, then cross-attention to its own projection of the text features, a bidirectional audio--video cross-attention, and an FFN.
In the exchange, audio reads video and video reads audio in parallel, from the same snapshot of both streams. Each direction has its own AdaLN modulation, driven by its own modality's noise level, and a zero-initialised residual gate, driven by the other modality's. The update that video reads from audio is added to the video layer's attention update.
At 768p each second contributes 40 audio tokens and about 3{,}000 video tokens to the exchange.

\paragraph{Physical-time alignment.}
Both streams share one clock, measured in seconds (\cref{fig:supp:timeline}). An audio token spans 1/40\,s. The first video latent represents frame zero and each subsequent latent spans eight frames. We assign each token the centre time of its span. Audio self-attention and both cross-modal attentions apply RoPE to these times, so audio and video tokens of the same instant share a phase at any frame rate or duration (details in \cref{sec:supp:av_arch}). On synthetic flashes, clicks and real clips, the VAE round trip keeps the frame offset at 0 and the audio offset at 0\,ms at 16 and 24\,fps.

\subsection{Choosing the Interaction}
\label{sec:arch:study}
We compared four ways of connecting the two streams (\cref{fig:avdesigns}): (a) a 20-layer audio stream whose tokens join the video's softmax attention at the ten softmax layers, (b) a parallel stream with cross-attention in every layer, as in~\cite{hacohen2026ltx2,low2025ovi}, (c) a hybrid audio stream with the video's 3:1 layout that exchanges by cross-attention at linear layers and joint attention at softmax layers, and (d) a mixture of transformers~\cite{liang2024mot} with one attention over both modalities in every layer.
Trained under matched settings, (b) \keyres{has the best text--audio agreement and is the only design that leaves the video attention unchanged}, while running audio inside the video's linear layers (c, d) is not meaningfully cheaper, so we keep (b) (\cref{tab:avarch}). Design (b) keeps the video stream's layer layout, kernels and RoPE, so the joint model inherits the video model's efficiency and its later optimisations, and the audio stream gets its own width and depth.
With matched data, recipe and training steps, its CLAP is $0.073$ higher than the baseline's (better on 16 of 20 prompts), and audio quality is on par across the four designs.
With longer training, it leads MoT on predicted content enjoyment (Audiobox CE 4.54 \vs 4.34) and on the share of clips with zero audio--video offset (0.39 \vs 0.16).

\begin{figure}[t]
\centering
\definecolor{vidc}{HTML}{76B900}
\definecolor{audc}{HTML}{A31F34}
\definecolor{xc}{HTML}{C77C02}
\tikzset{
  avd/.style={rounded corners=2.5pt, line width=0.7pt, minimum width=1.25cm, minimum height=0.6cm, inner sep=1pt,
              align=center, font=\footnotesize},
  vl/.style={avd, draw=vidc!65!black, fill=vidc!18},
  vs/.style={avd, draw=vidc!70!black, fill=vidc!80!black, text=white, font=\footnotesize},
  al/.style={avd, draw=audc!75!black, fill=audc!16},
  as/.style={avd, draw=audc!80!black, fill=audc!85, text=white, font=\footnotesize},
  joint/.style={rounded corners=3pt, line width=1.0pt, draw=xc!85!black, minimum width=1.4cm, minimum height=1.95cm,
                inner sep=1pt, align=center, font=\footnotesize, text=xc!80!black},
  xa/.style={<->, >={Stealth[length=4pt, width=4pt]}, draw=xc!90!black, line width=1.2pt, shorten >=1.5pt, shorten <=1.5pt},
  nx/.style={->, >={Stealth[length=3pt]}, draw=black!45, line width=0.7pt, shorten >=1pt, shorten <=1pt},
  rowl/.style={font=\scriptsize, text=black!60, anchor=east},
}
\providecommand{\avrows}{%
  \useasboundingbox (-0.8,-0.08) rectangle (3.15,1.98);
  \node[rowl, text=vidc!70!black] at (-0.05,1.6) {video};
  \node[rowl, text=audc!80!black] at (-0.05,0.35) {audio};
}
\providecommand{\avdesignA}{%
\begin{tikzpicture}
  \avrows
  \node[vl] (v1) at (0.75,1.6) {linear};
  \node[as] (a1) at (0.75,0.35) {softmax};
  \node[joint, fill=xc!18] (j) at (2.35,0.975) {joint\\AV\\softmax};
  \draw[nx] (v1.east) -- (v1.east -| j.west); \draw[nx] (a1.east) -- (a1.east -| j.west);
\end{tikzpicture}}
\providecommand{\avdesignB}{%
\begin{tikzpicture}
  \avrows
  \node[vl] (v1) at (0.75,1.6) {linear}; \node[vs] (v2) at (2.3,1.6) {softmax};
  \node[as] (a1) at (0.75,0.35) {softmax}; \node[as] (a2) at (2.3,0.35) {softmax};
  \draw[nx] (v1) -- (v2); \draw[nx] (a1) -- (a2);
  \draw[xa] (v1) -- (a1); \draw[xa] (v2) -- (a2);
  \node[font=\scriptsize, text=xc!80!black] at (1.525,0.98) {cross-attn};
\end{tikzpicture}}
\providecommand{\avdesignC}{%
\begin{tikzpicture}
  \avrows
  \node[vl] (v1) at (0.75,1.6) {linear};
  \node[al] (a1) at (0.75,0.35) {linear};
  \draw[xa] (v1) -- (a1);
  \node[joint, fill=xc!18] (j) at (2.35,0.975) {joint\\AV\\softmax};
  \draw[nx] (v1.east) -- (v1.east -| j.west); \draw[nx] (a1.east) -- (a1.east -| j.west);
\end{tikzpicture}}
\providecommand{\avdesignD}{%
\begin{tikzpicture}
  \avrows
  \node[joint, fill=vidc!25, draw=vidc!65!black, text=vidc!40!black] (j1) at (0.8,0.975) {joint\\AV\\linear};
  \node[joint, fill=xc!18] (j2) at (2.35,0.975) {joint\\AV\\softmax};
  \draw[nx] ([yshift=0.62cm]j1.east) -- ([yshift=0.62cm]j2.west); \draw[nx] ([yshift=-0.62cm]j1.east) -- ([yshift=-0.62cm]j2.west);
\end{tikzpicture}}
\captionsetup[subfigure]{justification=centering}
\begin{subfigure}[t]{0.24\linewidth}\centering\avdesignA\caption{Baseline: joint softmax at softmax layers only.}\end{subfigure}\hfill
\begin{subfigure}[t]{0.24\linewidth}\centering\avdesignB\caption{Our cross-attention in~every~layer.}\end{subfigure}\hfill
\begin{subfigure}[t]{0.24\linewidth}\centering\avdesignC\caption{Hybrid: cross-attention, then joint softmax.}\end{subfigure}\hfill
\begin{subfigure}[t]{0.24\linewidth}\centering\avdesignD\caption{MoT: joint attention in~every~layer.}\end{subfigure}
\caption{\textbf{Four ways to connect the audio stream to the hybrid video stream, each shown for one linear and one softmax layer.} Green: video, red: audio, amber: where they exchange information. Only (b) leaves the video attention unchanged.}
\label{fig:avdesigns}
\par\vspace{12pt}
\noindent\begin{minipage}[t]{0.53\linewidth}\vspace{0pt}\centering
{\coltabtype\captionsetup{type=table}\avtabcaption\avtabbody}
\end{minipage}\hfill
\begin{minipage}[t]{0.44\linewidth}\vspace{0pt}\centering
\vspace*{9pt}\adjustbox{max width=\linewidth}{\resizebox{\linewidth}{!}{%
\begin{tikzpicture}[x=6.3cm, y=1cm, font=\scriptsize]
  \definecolor{vidc}{HTML}{76B900}\definecolor{audc}{HTML}{A31F34}
  \fill[vidc!35, draw=vidc!70!black, line width=0.4pt] (-0.5/24,0.55) rectangle (0.5/24,0.95);
  \foreach \k in {1,2,3} {
    \fill[vidc!35, draw=vidc!70!black, line width=0.4pt] ({(8*\k-7.5)/24},0.55) rectangle ({(8*\k+0.5)/24},0.95);
    \node at ({(8*\k-3.5)/24},0.75) {latent \k};
    \fill[black] ({(8*\k-3.5)/24},0.55) circle (0.9pt);
  }
  \fill[black] (0,0.55) circle (0.9pt);
  \node[anchor=east] at (-0.03,0.75) {video};
  \foreach \k in {0,...,39} \fill[audc!35, draw=audc!70!black, line width=0.3pt] ({\k/40},0) rectangle ({(\k+1)/40},0.35);
  \foreach \k in {0,...,39} \fill[black] ({(\k+0.5)/40},0) circle (0.5pt);
  \node[anchor=east] at (-0.03,0.175) {audio};
  \draw[->, black!60] (-0.03,-0.25) -- (1.04,-0.25) node[anchor=west] {s};
  \foreach \t/\l in {0/0, 0.3333/{1/3}, 0.6667/{2/3}, 1/1} { \draw[black!60] (\t,-0.3) -- (\t,-0.2); \node[anchor=north, inner sep=1pt] at (\t,-0.3) {\l}; }
  \foreach \k in {1,2,3} \draw[densely dotted, black!45] ({(8*\k-3.5)/24},0.55) -- ({(8*\k-3.5)/24},-0.2);
\end{tikzpicture}}}
\par\vspace{\avrowH}
\captionsetup{skip=0pt,justification=justified,singlelinecheck=false}
\caption{\textbf{A shared physical-time clock at 24\,fps.} Dots mark token-centre times. Video latent 0 is placed at time 0, and later latents group eight frames. RoPE uses these times even when the two grids do not coincide.}
\label{fig:supp:timeline}
\end{minipage}
\end{figure}

\section{Audio--Video Pre-training}
\label{sec:pretrain}

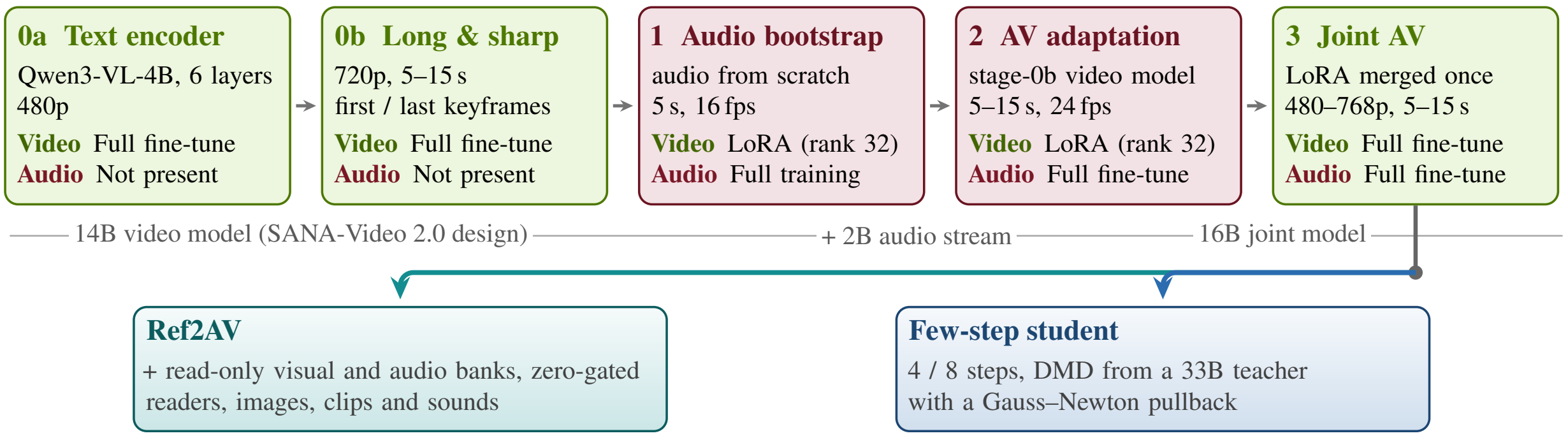
\begin{figure}[t]
\centering
\begin{adjustbox}{max width=\linewidth}
\definecolor{vidc}{HTML}{76B900}
\definecolor{audc}{HTML}{A31F34}
\definecolor{refc}{HTML}{138D90}
\definecolor{fsc}{HTML}{2B6CB0}
\begin{tikzpicture}[
    font=\footnotesize, >={Stealth[length=4pt, width=3.6pt]},
    card/.style={draw=#1!65!black, rounded corners=4pt, line width=0.7pt, fill=#1!13,
                 minimum width=3.1cm},
    head/.style={font=\small\bfseries, text=#1!65!black, anchor=north west, inner sep=0pt},
    fact/.style={font=\footnotesize, text=black!75, anchor=north west, inner sep=0pt, align=left, text width=2.95cm},
    flow/.style={->, draw=black!55, line width=1.0pt, shorten >=1.5pt, shorten <=1.5pt},
  ]
  \foreach \i/\col/\name/\setting/\clips/\videoMode/\audioMode in {
      0/vidc/{0a\; Text encoder}/{Qwen3-VL-4B, 6 layers}/{480p}/{Full fine-tune}/{Not present},
      1/vidc/{0b\; Long \& sharp}/{720p, 5--15\,s}/{first / last keyframes}/{Full fine-tune}/{Not present},
      2/audc/{1\; Audio bootstrap}/{audio from scratch}/{5\,s, 16\,fps}/{LoRA (rank 32)}/{Full training},
      3/audc/{2\; AV adaptation}/{stage-0b video model}/{5--15\,s, 24\,fps}/{LoRA (rank 32)}/{Full fine-tune},
      4/vidc/{3\; Joint AV}/{LoRA merged once}/{480--768p, 5--15\,s}/{Full fine-tune}/{Full fine-tune}} {
    \node[card=\col, anchor=south, align=left, text width=2.82cm,
          inner xsep=4pt, inner ysep=4pt] (S\i) at (3.45*\i,-1.27) {%
      {\small\bfseries\color{\col!65!black}\name\strut}\\[1pt]
      \setting\strut\\\clips\strut\\[2pt]
      \textcolor{vidc!55!black}{\bfseries Video}\enspace \videoMode\strut\\
      \textcolor{audc!75!black}{\bfseries Audio}\enspace \audioMode\strut};
  }
  \foreach \i [evaluate=\i as \j using int(\i+1)] in {0,...,3} \draw[flow] (S\i.east) -- (S\j.west);
  \draw[black!35, line width=0.5pt] (-1.5,-1.6) -- (15.4,-1.6);
  \node[font=\footnotesize, text=black!65, fill=white, inner sep=1.5pt] at (1.67,-1.6) {14B video model (\prev design)};
  \node[font=\footnotesize, text=black!65, fill=white, inner sep=1.5pt] at (8.37,-1.6) {+ 2B audio stream};
  \node[font=\footnotesize, text=black!65, fill=white, inner sep=1.5pt] at (12.35,-1.6) {16B joint model};
  \node[card=refc, top color=refc!4, bottom color=refc!22, minimum width=5.8cm, minimum height=1.35cm] (R) at (2.75,-3.05) {};
  \node[head=refc] at ([xshift=4pt,yshift=-4pt]R.north west) {Ref2AV};
  \node[fact, text width=5.6cm] at ([xshift=4pt,yshift=-17pt]R.north west) {+ read-only visual and audio banks, zero-gated\\readers, images, clips and sounds};
  \node[card=fsc, top color=fsc!4, bottom color=fsc!22, minimum width=5.8cm, minimum height=1.35cm] (F) at (11.05,-3.05) {};
  \node[head=fsc] at ([xshift=4pt,yshift=-4pt]F.north west) {Few-step student};
  \node[fact, text width=5.6cm] at ([xshift=4pt,yshift=-17pt]F.north west) {4 / 8 steps, DMD from a 33B teacher\\with a Gauss--Newton pullback};
  \draw[line width=1.3pt, draw=black!60] (S4.south) -- (13.8,-2.0);
  \fill[black!60] (13.8,-2.0) circle (2.2pt);
  \draw[->, >={Stealth[length=7pt, width=6pt]}, line width=1.5pt, draw=refc, shorten >=2pt, rounded corners=4pt] (13.8,-2.0) -| (R.north);
  \draw[->, >={Stealth[length=7pt, width=6pt]}, line width=1.5pt, draw=fsc, shorten >=2pt, rounded corners=4pt] (13.8,-2.0) -| (F.north);
\end{tikzpicture}
\end{adjustbox}
\caption{\textbf{Training proceeds in stages from the video model to a joint audio--video model, which then seeds Ref2AV and the few-step student.} Video adaptation (0a, 0b), audio bootstrap beside the frozen, LoRA-adapted video (1--2), then joint training (3).}
\label{fig:overview}
\end{figure}

\paragraph{Staged training.}
\takeaway{A randomly initialised audio stream attached to a trained video model sends noise-like gradients into the video stream.} We therefore couple the two streams in stages (\cref{fig:overview}).
\emph{Stage~0} trains the 14B video model alone, following~\cite{chen2026sanavideo2}: it is first adapted to the new text encoder at 480p (0a) and then extended to 720p, 5--15\,s clips and keyframes (0b).
\emph{Stage~1} trains the audio stream from scratch beside the frozen video weights, which get only a rank-32 LoRA~\cite{hu2022lora}.
\emph{Stage~2} transfers the audio stream to the stage-0b video model and retrains the LoRA on 5--15\,s clips.
\emph{Stage~3} merges the LoRA and trains all 16B parameters with per-stream peak learning rates ($2{\times}10^{-5}$ for video, $5{\times}10^{-5}$ for audio) and FSDP~\cite{zhao2023fsdp} (\cref{sec:supp:pretrain}).

\paragraph{Shared noise clock.}
Both streams use rectified flow, $z_\sigma=(1-\sigma)z+\sigma\epsilon$, with independent noise but one shared base time $u$ from a logit-normal density~\cite{esser2024sd3}. Each modality maps it through its own shift, $\sigma_m=\mathrm{shift}(u,s_m)$ with $\mathrm{shift}(u,s)=su/(1+(s-1)u)$, $s_v$ set by the token count and \mbox{$s_a=2$}.
With clean latents $z_m$, noise $\epsilon_m$ and predicted velocities $\hat v_m$, the loss sums the per-modality velocity errors, each averaged over the valid tokens $M_m$ of the clip,
\begin{equation*}
  \mathcal L=\sum_{m\in\{v,a\}}\operatorname{MSE}_{M_m}\!\left(\hat v_m,\,\epsilon_m-z_m\right).
\end{equation*}
Sharing $u$ keeps both modalities at corresponding stages of denoising, as at inference. Since shifts compose (shift $k$ followed by shift $s$ equals shift $ks$), an inference shift $k$ re-spaces the steps but visits only trained $(\sigma_v,\sigma_a)$ pairs.
Training mixes text-to-audio--video with first, last and first-and-last frame conditioning, and drops each modality's text with probability 10\% (5\% jointly), so classifier-free guidance (CFG)~\cite{ho2022cfg} is trained per modality and jointly.

\paragraph{Base model.}
After joint training, the base model is the parent of the students, sampled in \cref{tab:main} with 40 steps and video guidance (80 NFE). A 30-step SDE sampler with one joint unconditional pass (58 NFE) halves the evaluations of the 40-step default with guidance on both streams (120 NFE) and raises HPSv3 (8.71 \vs 8.66), VideoReward (0.30 \vs 0.01), audio quality (PQ 7.11 \vs 6.88) and CLAP (0.167 \vs 0.149) on the final base model. The judge score changes from 8.47 to 8.50 and ASR accuracy falls from 0.872 to 0.851 (\cref{tab:pretrain_eval}). It combines CFG rescale~\cite{lin2024flawed} with interval guidance~\cite{kynkaanniemi2024interval}, selected by rewards and blind review (\cref{sec:supp:sampler}).

\section{Few-Step Distillation from a Stronger Teacher}
\label{sec:fewstep}
To cut the joint model's 40 steps to 4--8, we distill it from a stronger external teacher in another latent space while keeping the student's efficient backbone and VAE, so the teacher's update must be carried back faithfully.

\subsection{Setting}

\paragraph{Distribution matching.}
DMD~\cite{yin2023dmd,yin2024dmd2} trains a few-step generator by minimising the reverse KL between its output distribution and the teacher's at every noise level. With $z_\sigma$ a generated sample $z$ noised to level $\sigma$, and $\hat z_0(\cdot)$ the one-step clean predictions of a frozen teacher (\emph{real}) and an online critic (\emph{fake}), the update is
\begin{equation}
g \;=\; \big(\hat z_0^{\text{fake}}(z_\sigma)-\hat z_0^{\text{real}}(z_\sigma)\big)/c,
\label{eq:dmd}
\end{equation}
with $c$ a per-sample normaliser, and the generator regresses onto $z-g$, towards the teacher. DMD2 adds a GAN loss on the critic's features, and DMD + GAN is our base recipe.

\begin{figure}[t]
\centering
\begin{adjustbox}{max width=\linewidth}
\definecolor{stc}{HTML}{2B6CB0}
\definecolor{adc}{HTML}{E07B00}
\begin{tikzpicture}[
    font=\footnotesize, >={Stealth[length=3.5pt]},
    box/.style={draw=black!60, rounded corners=3pt, fill=black!6, inner sep=3pt, align=center, minimum height=26pt},
    tbox/.style={box, draw=nvgreendark, fill=nvgreen!22},
    sbox/.style={box, draw=stc!80!black, top color=stc!4, bottom color=stc!24},
    num/.style={circle, fill=#1, text=white, font=\scriptsize\bfseries, inner sep=0pt, minimum size=10pt},
    lab/.style={font=\scriptsize, text=black!70, align=center},
    flow/.style={->, draw=black!70, line width=0.6pt},
    fix/.style={->, draw=nvgreendark, line width=1.0pt},
  ]
  \fill[stc!9, rounded corners=5pt] (-1.55,-2.95) rectangle (3.3,1.0);
  \fill[nvgreen!9, rounded corners=5pt] (3.3,-2.95) rectangle (13.45,1.0);
  \fill[stc!9] (2.9,-2.95) rectangle (3.3,1.0);
  \fill[nvgreen!9] (3.3,-2.95) rectangle (3.7,1.0);
  \draw[densely dashed, draw=black!35] (3.3,-2.95) -- (3.3,1.0);
  \node[anchor=south west, font=\scriptsize\bfseries, text=stc!85!black] at (-1.5,-2.92) {student latent space (LTX-2.5 VAE)};
  \node[anchor=south east, font=\scriptsize\bfseries, text=nvgreendark] at (13.4,-2.92) {teacher latent space (\Hthree{} VAE)};
  \node[sbox, text width=2.1cm] (stu) at (0,0) {student $G_\theta$\\4 / 8 steps};
  \node[box, draw=adc, line width=1.2pt, left color=stc!22, right color=nvgreen!30, text width=2.35cm] (T) at (3.3,0)
       {\textbf{latent adapter} $\T$\\{\scriptsize 142M, bridges the two spaces}};
  \node[box, text width=1.1cm] (noise) at (6.15,0) {add noise\\$z_\sigma$};
  \node[tbox, text width=2.6cm, minimum height=18pt] (real) at (8.8,0.42) {real score: frozen \Hthree};
  \node[box, text width=2.6cm, minimum height=18pt] (fake) at (8.8,-0.42) {fake score: \Hthree{} + LoRA};
  \node[box, text width=2.3cm] (g) at (12.0,0) {DMD update $g$\\{\scriptsize teacher's correction}};
  \draw[flow] (stu) -- node[above, lab] {$x$} (T);
  \draw[flow] (T) -- node[above, lab] {$\T(x)$} (noise);
  \draw[flow] (noise.east) -- ++(0.2,0) |- (real.west);
  \draw[flow] (noise.east) -- ++(0.2,0) |- (fake.west);
  \draw[flow] (real.east) -- ++(0.2,0) |- ([yshift=3pt]g.west);
  \draw[flow] (fake.east) -- ++(0.2,0) |- ([yshift=-3pt]g.west);
  \draw[fix] (g.north) -- (12.0,1.3) -| node[pos=0.25, above, text=nvgreendark, font=\footnotesize, inner sep=1.5pt]
      {Gauss--Newton pullback: a student correction whose image under $\J$ approximates $g$\;
       {\scriptsize(5B cosine: backprop 0.22 $\rightarrow$ ours 0.83)}}
      (stu.north);
  \node[num=nvgreendark] at (12.0,1.02) {1};
  \node[sbox, text width=2.5cm] (rv) at (0,-1.8) {\Hthree{} videos, LTX-encoded\\{\scriptsize (smooth, high-motion)}};
  \node[tbox, text width=2.3cm] (D) at (8.8,-1.8) {discriminator $D$\\{\scriptsize on teacher features}};
  \draw[flow] (rv.east) -- node[pos=0.62, below, lab] {real, through $\T$} (D.west);
  \draw[flow, draw=black!55] (T.south) -- (3.3,-1.1) -| node[pos=0.13, above, lab] {fake $\T(x)$} ([xshift=-0.55cm]D.north);
  \draw[fix] (D.south) |- (5.0,-3.18) -| node[pos=0.25, below, lab, text=nvgreendark, inner sep=1.5pt] {GAN loss} (-1.85,0) -- (stu.west);
  \node[num=nvgreendark] at ([xshift=3pt, yshift=3pt]D.north east) {2};
\end{tikzpicture}
\end{adjustbox}
\caption{\textbf{A latent adapter lets a teacher in another latent space supervise the student.} Blue: student latent space. Green: teacher's. Teacher and critic estimate $g$ from the same noised adapter output. \ding{172}~The Gauss--Newton pullback finds a student correction whose image through $\J$ matches $g$. \ding{173}~As in DMD2, a GAN branch uses the teacher's smooth, high-motion videos as reals.}
\label{fig:method}
\end{figure}
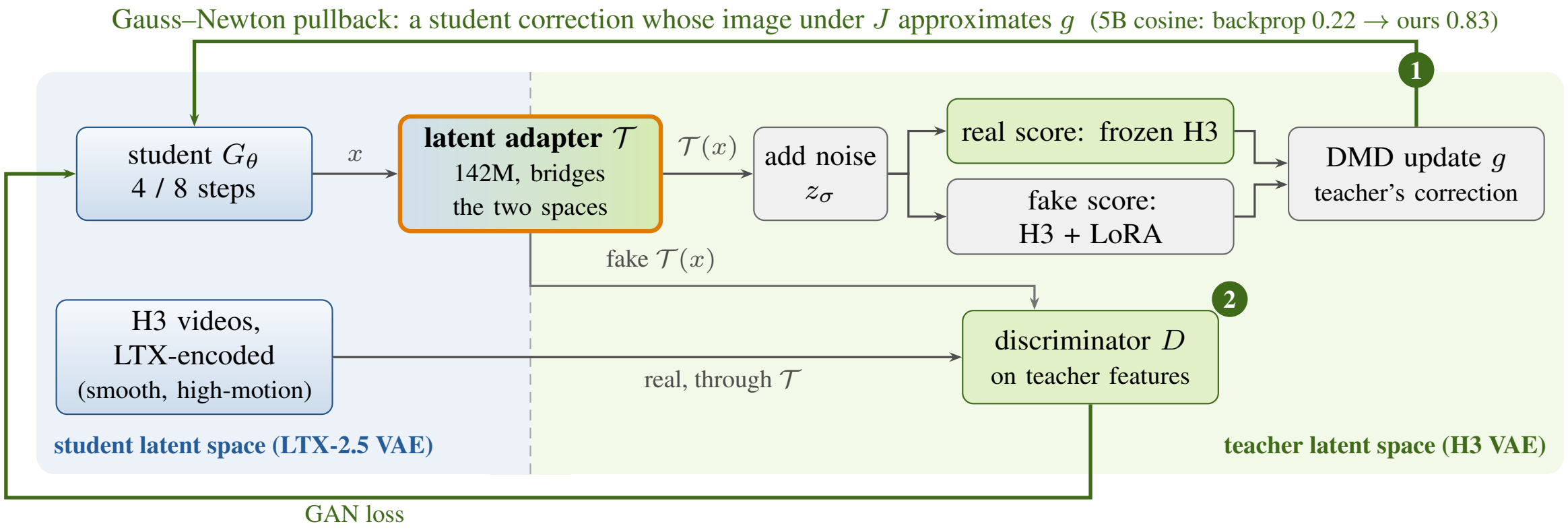

\paragraph{A teacher in another latent space.}
The students start from the 16B \ours model and, for ablations, the 5B \prev. Both generate in the LTX-2.5 latent space, $x\in\mathbb{R}^{n}$, while the teacher, the 33B MiniMax-H3~\cite{minimax2026h3}, has its own VAE, $z\in\mathbb{R}^{m}$.
A latent adapter $\T$ bridges the two, $z=\T(x)$: a 142M-parameter spatiotemporal convolutional network trained on paired encodings of the same videos (\cref{fig:method}). Decoding with the student's VAE and re-encoding with the teacher's would instead cost 5.9\,s per pass and 67\,GiB to backpropagate. The teacher and adapter are used only during training, so the student samples in its own latent space.
We add noise to $z=\T(x)$ and feed the same $z_\sigma$ to the teacher and critic in \cref{eq:dmd}. It remains to turn the correction $g$ in teacher space into a student update.

\subsection{Pullback Distortion and the Gauss--Newton Pullback}
\label{sec:pullback}

\paragraph{Realised fidelity.}
With step size $\eta>0$, the target in teacher space is $\T(x)-\eta g$. We seek a student displacement $\Delta x$ that produces this change. Linearising the adapter gives $\J\Delta x\approx-\eta g$, where $\J=\partial\T/\partial x$. Writing $\Delta x=-\eta d$, we want $\J d\approx g$, and measure directional agreement by $\rho(d)=\cos(\J d,g)$ (1 for the same-latent update $d=g$).
Plain backpropagation chooses $d=\J^\top g$, so its mapped correction is $\J\J^\top g$, which favours the adapter's highest-gain directions.
On 128 rollouts of the 5B student, \takeaway{backpropagation strongly distorts the update direction} ($\rho=0.22$), and the realised move is mostly a global colour shift (83\% of its energy in the lowest spatial frequencies, against 18\% for $g$, \cref{fig:analysis}).
On eight pairs, the exact VAE path gives $\rho=0.02$, versus 0.19 through the adapter.

\paragraph{Gauss--Newton pullback.}
We therefore solve for a student correction whose mapped effect matches $g$, with damping $\lambda>0$ so that the adapter's low-gain directions are not amplified:
\begin{equation}
d_{\text{GN}}=\argmin_d\bigl(\lVert\J d-g\rVert^2+\lambda\lVert d\rVert^2\bigr)=(\J^\top\J+\lambda I)^{-1}\J^\top g.
\end{equation}
We solve it matrix-free with $k$ conjugate-gradient (CG) steps~\cite{hestenes1952cg}, each one Jacobian--vector and one vector--Jacobian product through $\T$.
We match the solution's norm to $\lVert\J^\top g\rVert$ and use the resulting correction $\widetilde d_{\text{GN}}$ in the regression target $x-\eta\widetilde d_{\text{GN}}$, so only the direction changes. The first CG step from zero is parallel to $\J^\top g$, so after norm matching $k=1$ is backpropagation, and we call $k$ steps \GN{$k$}. The derivation, relative damping and rescaling are in \cref{sec:supp:pullback}.
\keyres{GN raises fidelity to 0.62 at $k=8$ and 0.83 at $k=32$} (0.71 and 0.88 for 16B), bringing the low-frequency energy closer to that of $g$. \GN{32} adds about 4\% to the measured 5B training iteration. Low-pass filtering of the VJP direction reaches 0.23, so smoothing alone cannot explain the gain.

\section{Results and Deployment}

\paragraph{Students and implementation.}
The 16B student starts from the 40-step joint model. A single \Hthree evaluation supplies both teacher predictions. Video uses the adapter and \GN{32}, while audio shares \Hthree's VAE and receives its correction directly. The objective combines video and audio DMD with a GAN loss (weight 0.01) on smooth, high-motion teacher videos. We train first at 480p, then at 768p on 5--15\,s clips in both orientations. Mixing 4- and 8-step rollouts gives one checkpoint for both settings (\cref{sec:supp:14b}).
The 5B testbed baseline $\theta^\star$ is our strongest DMD + GAN student with the backpropagated pullback. The pullback ablations continue from its full training state and change nothing else.

\paragraph{Evaluation.}
We use 256 fixed prompts, one clip per model and prompt, scored by HPSv3~\cite{ma2025hpsv3}, VideoReward~\cite{liu2025videoalign}, a multimodal judge and VBench~\cite{huang2023vbench} dimensions. Standard-prompt VBench uses 946 prompts and all 16 dimensions, with 5 samples per prompt for re-tested models in the main comparison and 1 for the 5B pullback ablation (\cref{sec:supp:eval}).
Latency is end to end on one GB200 (text encoding, sampling, decoding) with each model's fastest tested inference.

\subsection{Main Results}

\Cref{tab:main} compares the 16B audio--video student with multi-step and few-step generators.
With audio at $1344{\times}768$, the 8-step student is faster than LTX-2.5 distilled (5.0 \vs 5.3\,s) and scores higher on HPSv3 (10.18 \vs 8.61), VideoReward (2.03 \vs 1.17) and VBench (85.4 \vs 83.4). The 4-step student retains similar scores at 2.9\,s.
At the same resolution, prompts and audio, the few-step \Hthree models (VBench 85.1, HPSv3 9.24--9.25) and the 50-step teacher (HPSv3 9.13, VR 1.69) score below both students at 6--7$\times$ and about 45$\times$ the latency of the 4-step student.
The video-only 4-step baselines generate no audio and take 6--9$\times$ as long as the 4-step student (Wan2.2 with Lightning 27.4\,s at 720p, HunyuanVideo-1.5 17.4\,s at 480p), and they score lower on VideoReward (1.42 and 0.92) and VBench (82.4 and 81.1).
The student also scores above its 40-step parent on both preference models (HPSv3 10.18 \vs 9.03, VideoReward 2.03 \vs 1.41) at 8$\times$ lower latency, and is on par with it on VBench (\cref{tab:vbench}): 85.39 at 8 steps and 85.19 at 4 steps against 85.50, also above the published Wan2.1 few-step students (84.3--85.2 at 480p). The 8-step student has the highest VBench Quality of all re-tested models (86.73), above its parent (86.43), FastH3 (86.59) and LTX-2.5 (84.79).

At 768p, \GN{32} raises video-update fidelity from 0.33 to 0.88 across both orientations. The share of energy in the lowest spatial-frequency band falls from 70\% to 20\%, close to the 19\% in the teacher's correction (\cref{sec:supp:14b}).

\begin{figure}[!b]
\centering
\includegraphics[width=\linewidth]{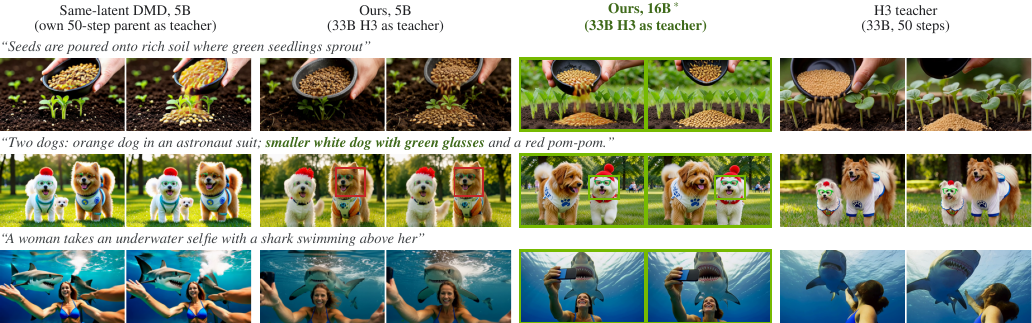}
\caption{\textbf{Appearance transfer and attribute following.} The larger 16B student puts the green glasses on the white dog (green box), the 5B student on the orange dog (red box). Same source captions, 4-step students. $^\ast$: structured rewrite.}
\label{fig:qual}
\end{figure}

In \cref{fig:qual}, the larger 16B student follows the prompt and puts the green glasses on the white dog, while the 5B student wrongly puts them on the orange dog.

\begin{table}[!t]
\centering
\small
\setlength{\tabcolsep}{4.2pt}
\caption{\textbf{Our few-step audio--video model against multi-step and few-step video generators.} Lat.: end-to-end seconds on one GB200 at each model's short-clip setting (\cref{sec:supp:eval}). HPSv3 and VR (VideoReward): 256 prompts. Re-tested VBench: 946 prompts $\times$ 5 seeds. Audio--video models at $1344{\times}768$ with audio, \prev at $1280{\times}704$, Wan2.2 at $1280{\times}720$ and HunyuanVideo-1.5 at $848{\times}480$. $^\S$Published, 121 frames and one seed. Bold: best few-step.}
\label{tab:main}
\begin{adjustbox}{max width=\linewidth}
\begin{tabular}{@{}l c c c r c c c@{}}
\toprule
\textbf{Model} & \textbf{Params} & \textbf{Audio} & \textbf{Steps} & \textbf{Lat.\,(s)} & \textbf{HPSv3}$\uparrow$ & \textbf{VR}$\uparrow$ & \textbf{VBench}$\uparrow$ \\
\midrule
\rowcolor{catgray}\multicolumn{8}{@{}l}{\emph{Multi-step models}} \\
MiniMax-\Hthree~\cite{minimax2026h3} & 33B & \cmark & 50 & 129 & 9.13 & 1.69 & -- \\
\prev~\cite{chen2026sanavideo2} & 5B & -- & 50 & 21.0 & 8.41 & 1.20 & 85.29$^\S$ \\
\rowcolor{nvidiagreen!15}\ours & 16B & \cmark & 40 & 42.2 & 9.03 & 1.41 & 85.50 \\
\midrule
\rowcolor{catgray}\multicolumn{8}{@{}l}{\emph{Few-step models}} \\
\Hthree-Turbo~\cite{lightx2v2026h3turbo} & 33B & \cmark & 5 & 20.0 & 9.24 & 1.69 & 85.07 \\
FastH3~\cite{fastvideo2026fasth3} & 33B & \cmark & 4 & 17.7 & 9.25 & 1.11 & 85.09 \\
Wan2.2-A14B~\cite{wan2025wan} + Lightning~\cite{lightx2v2025wan22lightning} & 27B & -- & 4 & 27.4 & \textbf{10.30} & 1.42 & 82.35 \\
HunyuanVideo-1.5~\cite{wu2025hunyuanvideo15} 4-step~\cite{lightx2v2025hy15distill} & 8.3B & -- & 4 & 17.4 & 8.24 & 0.92 & 81.11 \\
LTX-2.5 distilled~\cite{hacohen2026ltx2} & 22B & \cmark & 8 & 5.3 & 8.61 & 1.17 & 83.37 \\
\rowcolor{nvidiagreen!15}\textbf{\ours} & 16B & \cmark & 8 & 5.0 & 10.18 & 2.03 & \textbf{85.39} \\
\rowcolor{nvidiagreen!15}\textbf{\ours} & 16B & \cmark & 4 & \textbf{2.9} & 10.15 & \textbf{2.07} & 85.19 \\
\bottomrule
\end{tabular}
\end{adjustbox}
\end{table}

\subsection{Audio Quality}
\coltabpar[0.43][0.54]{Audio is mixed to mono, resampled to 48\,kHz and scored on the first 5\,s. Against LTX-2.5 distilled (\cref{tab:audio}), the 8-step student's sound follows the prompt more closely (CLAP 0.180 \vs 0.123), has a higher PaSST Inception Score (3.64 \vs 2.58) and a lower predicted offset (DeSync 0.69 \vs 0.90\,s). The paired 95\% bootstrap intervals for CLAP and DeSync both exclude zero (\cref{sec:supp:audio_ci}). LTX-2.5 scores higher on production quality (6.71 \vs 6.26) and ImageBind similarity (0.182 \vs 0.141). The 8-step student has the lowest predicted offset in the table, including against the 50-step teacher (0.84\,s). The 40-step parent already exceeds the teacher on production quality (6.42 \vs 6.28) and DeSync (0.79 \vs 0.84\,s). The 4-step student still leads LTX-2.5 on CLAP, IS and DeSync (0.178, 3.27 and 0.72\,s).}{\captionsetup{skip=4pt}
\caption{\textbf{Audio of the audio--video models in \cref{tab:main}.} PQ: production quality~\cite{tjandra2025audiobox}. CLAP: text--audio match~\cite{wu2023clap}. IB: ImageBind audio--video similarity~\cite{girdhar2023imagebind}. IS: PaSST Inception Score~\cite{koutini2022passt}. DS: DeSync, audio--video offset in seconds~\cite{iashin2024synchformer}. Bold: best few-step.}
\label{tab:audio}
{\fontsize{8.6}{10.4}\selectfont\setlength{\tabcolsep}{1.3pt}\renewcommand{\arraystretch}{1.45}%
\begin{tabularx}{\linewidth}{@{}l *{6}{>{\centering\arraybackslash}X}@{}}
\toprule
\textbf{Model} & \textbf{Steps} & \textbf{PQ}$\uparrow$ & \textbf{CLAP}$\uparrow$ & \textbf{IB}$\uparrow$ & \textbf{IS}$\uparrow$ & \textbf{DS}$\downarrow$ \\
\midrule
MiniMax-\Hthree~\cite{minimax2026h3} & 50 & 6.28 & 0.207 & 0.193 & 3.84 & 0.84 \\
\rowcolor{nvidiagreen!15}Ours & 40 & 6.42 & 0.194 & 0.159 & 3.32 & 0.79 \\
\midrule
\Hthree-Turbo~\cite{lightx2v2026h3turbo} & 5 & 6.26 & 0.158 & 0.145 & 3.55 & 0.81 \\
FastH3~\cite{fastvideo2026fasth3} & 4 & 6.24 & 0.159 & 0.153 & 3.52 & 0.71 \\
LTX-2.5 distilled~\cite{hacohen2026ltx2} & 8 & \textbf{6.71} & 0.123 & \textbf{0.182} & 2.58 & 0.90 \\
\rowcolor{nvidiagreen!15}\textbf{Ours} & 8 & 6.26 & \textbf{0.180} & 0.141 & \textbf{3.64} & \textbf{0.69} \\
\rowcolor{nvidiagreen!15}\textbf{Ours} & 4 & 6.23 & 0.178 & 0.148 & 3.27 & 0.72 \\
\bottomrule
\end{tabularx}}
}

\begin{figure}[t]
\centering
\noindent\begin{minipage}[c]{0.505\linewidth}\centering
\includegraphics[width=\linewidth]{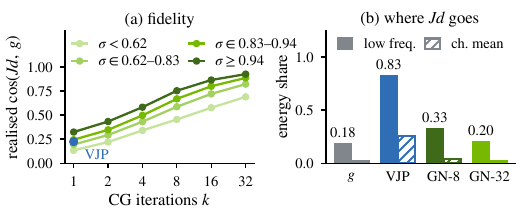}
\end{minipage}\hfill
\begin{minipage}[c]{0.473\linewidth}\centering
\includegraphics[width=\linewidth]{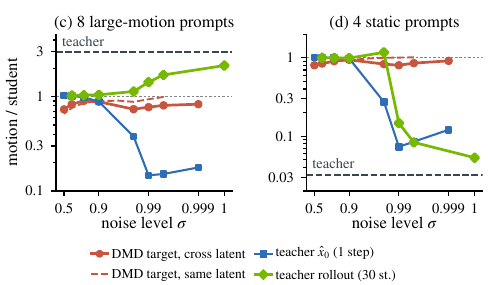}
\end{minipage}
\caption{\textbf{Where the teacher's update loses its direction, and where the motion is (5B student).} (a) Realised fidelity $\cos(\J d,g)$ against CG iterations $k$ ($k=1$: backpropagation). (b) Energy share of $\J d$ in the lowest spatial frequency (solid) and in channel means (hatched). (c, d) Motion relative to the student against $\sigma$. Horizontal lines: student (1) and teacher. (c) On large-motion prompts, the DMD target (red) and the teacher's one-step prediction (blue) move no more than the student, while the 30-step rollout (green) climbs towards the teacher. (d) On static prompts, the rollout removes the student's drift and the DMD target keeps it.}
\label{fig:analysis}\label{fig:motion}
\par\vspace{3pt}
\captionsetup{type=table}
\caption{\textbf{Ablations on the 5B testbed.} (a) Realised fidelity $\rho$ through $\T$, $\T^{+}$ and the exact VAE path (eight pairs, $^\dagger$\GN{6}). VBench: 946 prompts, one sample each, $832{\times}480$, after matched continuations (\cref{sec:supp:eval}). (b) 4-step students. Judge, HPSv3 and VR use 256 prompts. VBench: same 946 prompts, five seeds each, 480p, so the 5B \prev parent differs from \cref{tab:main}.}
\captionsetup[subtable]{font=small,labelfont=bf,justification=raggedright,singlelinecheck=false,skip=3pt,hypcap=true}
\noindent\begin{subtable}[t]{0.475\linewidth}
\centering
\caption{Pullback after matched training continuations}
\label{tab:pullback}
\footnotesize
\setlength{\tabcolsep}{1.8pt}
\begin{tabularx}{\linewidth}{@{}l *{3}{>{\centering\arraybackslash}X} rrr@{}}
\toprule
 & \multicolumn{3}{c}{\textbf{$\rho$ through}} & \multicolumn{3}{c}{\textbf{VBench (1 seed)}} \\
\cmidrule(lr){2-4}\cmidrule(lr){5-7}
\textbf{Method} & \textbf{$\T$} & \textbf{$\T^{+}$} & \textbf{exact} & \textbf{Total}$\uparrow$ & \textbf{Quality}$\uparrow$ & \textbf{Semantic}$\uparrow$ \\
\midrule
VJP & 0.22 & 0.26 & 0.02 & 84.75 & 86.42 & 78.09 \\
low-pass VJP & 0.23 & -- & -- & 84.58 & 86.28 & 77.80 \\
\rowcolor{nvidiagreen!15}\GN{8} & 0.62 & 0.65 & \textbf{0.05}$^\dagger$ & 85.05 & 86.66 & \textbf{78.62} \\
\rowcolor{nvidiagreen!15}\GN{32} & \textbf{0.83} & \textbf{0.85} & -- & \textbf{85.07} & \textbf{86.81} & 78.12 \\
\bottomrule
\end{tabularx}
\end{subtable}
\hfill
\begin{subtable}[t]{0.502\linewidth}
\centering
\caption{Teacher, GAN, pullback and motion conditioning}
\label{tab:fiveb}
\footnotesize
\setlength{\tabcolsep}{2.1pt}
\begin{tabularx}{\linewidth}{@{}>{\raggedright\arraybackslash}X l r r r r@{}}
\toprule
\textbf{5B model} & \textbf{Teacher} & \textbf{Judge}$\uparrow$ & \textbf{HPSv3}$\uparrow$ & \textbf{VR}$\uparrow$ & \textbf{VBench}$\uparrow$ \\
\midrule
parent, 50 steps & -- & 7.11 & 8.41 & 1.20 & 84.84 \\
\midrule
same-latent DMD & parent & 7.54 & 10.19 & 1.73 & 84.49 \\
cross-latent DMD & \Hthree & 7.17 & \textbf{10.58} & \textbf{1.99} & 83.18 \\
\rowcolor{nvidiagreen!15}+ GAN + GN pullback & \Hthree & \textbf{7.85} & 10.29 & 1.48 & 84.97 \\
\rowcolor{nvidiagreen!15}\quad + motion score & \Hthree & 7.83 & 10.29 & 1.46 & \textbf{85.63} \\
\bottomrule
\end{tabularx}
\end{subtable}

\end{figure}

\subsection{Ablations}

\paragraph{Pullback.}
\takeaway{GN improves realised fidelity on the 5B testbed} (\cref{fig:analysis}). After matched continuations, \GN{8} and \GN{32} also score higher on VBench Total (85.05 and 85.07 \vs 84.75 for VJP, \cref{tab:pullback}). \GN{32} leads on Quality (86.81) and \GN{8} on Semantic (78.62). Both GN arms end above their common start $\theta^\star$ (84.97, one seed), while VJP ends below it.
Low-pass VJP, as smooth as \GN{8} but no more faithful, scores 84.58, and a more accurate adapter $\T^{+}$ lifts VJP fidelity only from 0.22 to 0.26. Solving for the realised move gives larger fidelity gains than smoothing the gradient or a better adapter. The exact decode--re-encode path does not help either ($\rho=0.019$, and 0.048 after six GN iterations, \cref{sec:supp:exact}). On 256 other prompts, \GN{32} raises VBench Quality from 85.72 to 86.02 (95\% interval $[0.01,0.59]$, \cref{tab:supp:pullback_vbench}).

\paragraph{Stronger teacher.}
\takeaway{With the GAN branch and the GN pullback, distilling from \Hthree raises the judge score, HPSv3 and VBench over distilling from the student's own parent} (\cref{tab:fiveb}), and the students take on the teacher's colour, lighting and layout (\cref{fig:qual}). The judge score rises from 7.54 (same-latent DMD) to 7.85, above the 50-step parent (7.11), at 1.1\,s per $1280{\times}704$ video.
The plain cross-latent student has the lowest VBench Total (83.18). Adding a per-prompt motion score to the GAN and GN recipe raises Total (five seeds) from 84.97 to 85.63, above the 5B parent's 84.84.

\paragraph{Motion in the distillation target.}
Decoding each term of \cref{eq:dmd} shows that the teacher's and critic's one-step predictions are blurred conditional means whose shared motion cancels in their difference (\cref{fig:motion}c,d and \cref{sec:motion}), so \takeaway{the one-step DMD targets provide little of the missing motion on these probes}. The motion is not lost in the teacher itself: in the example of \cref{fig:supp:chain_strip}, its 30-step rollout from the same noise moves 2.8$\times$ as much as the student, while the one-step DMD target moves 0.85$\times$ as much. The student can also represent this motion: direct regression onto 30-step teacher rollouts raises its motion from 0.27$\times$ to 0.83$\times$ the teacher's (\cref{sec:motion}).

\subsection{Efficiency and Deployment}
\label{sec:deployment}
\begin{figure}[!t]
\noindent\begin{minipage}[t]{0.49\linewidth}\vspace{0pt}\centering
\includegraphics[width=\linewidth]{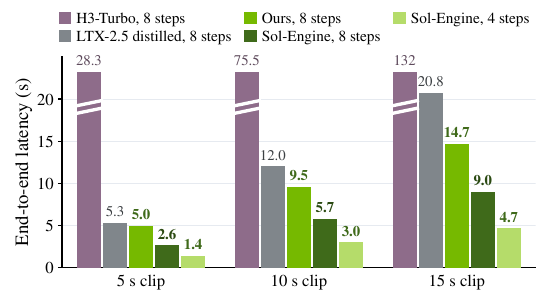}
\captionsetup{skip=3pt,justification=justified,singlelinecheck=false}
\caption{\textbf{The student is faster than 8-step \Hthree-Turbo and LTX-2.5 distilled at every clip length, and the gap widens with duration.} End to end on one GB200 at $1344{\times}768$ with audio. \Hthree-Turbo: LightX2V LoRA (5 steps in \cref{tab:main}), bars truncated. Durations use the nearest supported frame counts (\cref{sec:supp:deployment}). Sol-Engine uses MXFP8 and TAEHV at the indicated step count.}
\label{fig:deployment}
\end{minipage}\hfill
\begin{minipage}[t]{0.49\linewidth}\vspace{0pt}\centering
\includegraphics[width=\linewidth]{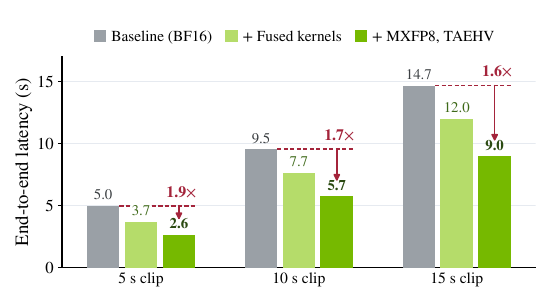}
\captionsetup{skip=3pt,justification=justified,singlelinecheck=false}
\caption{\textbf{Sol-Engine deployment~\cite{solengine}} of the 8-step student. Latency on one GB200 at $1344{\times}768$ with audio. Cumulative settings from the student baseline of \cref{fig:deployment}. TAEHV: distilled VAE decoder. Red: speedup over the baseline. Each bar is the median of twelve requests on one node.}
\label{fig:engine}
\end{minipage}
\end{figure}

\Cref{fig:deployment} reports end-to-end generation time on one GB200, with each model in its fastest tested implementation. The student uses FlashAttention-4 for its softmax layers, padding-free cross-attention with cached text keys and values, merged projections and compiled pointwise work, while LTX-2.5 runs its compiled native pipeline (\cref{sec:supp:deployment}). For our head dimension of 256, FlashAttention-4 is 3.6--4.0$\times$ faster per call than the FlashAttention-2 fallback, and together these changes cut 5, 10 and 15\,s generation from 6.7, 13.4 and 21.5\,s to 5.0, 9.5 and 14.7\,s. Compiling LTX-2.5 likewise cuts its eager 6.8, 14.6 and 24.3\,s to 5.3, 12.0 and 20.8\,s.
The 8-step student is 8$\times$ faster than its 40-step parent and faster than 8-step LTX-2.5 distilled at every length, by 1.06$\times$ at 5\,s and 1.41$\times$ at 15\,s. From 5 to 15\,s the student's latency grows 2.9$\times$, compared with 3.9$\times$ for LTX-2.5.
It is 5.7$\times$, 7.9$\times$ and 9.0$\times$ faster than 8-step \Hthree-Turbo at 5, 10 and 15\,s.

\paragraph{Sol-Engine.}\label{sec:engine}
Keeping the denoiser weights and step count, we deploy the student with Sol-Engine~\cite{solengine,li2026solh3} (\cref{fig:engine,sec:supp:deployment}).
In the bfloat16 (BF16) baseline, matrix multiplications (GEMMs) already run at the GPU power limit and take 43--48\% of denoising time, normalisation, modulation and residual updates about 40\%, and attention 8--18\%. Most of the remaining cost is therefore glue code and kernel launches.
Sol-Engine fuses these pointwise operations, folds activations and biases into GEMM epilogues and captures FlashAttention-4 cross-attention in a CUDA graph for each clip length. This cuts a 15\,s clip from the 14.7\,s baseline to 12.0\,s in BF16, with the same numerical error against the native model as the baseline (relative velocity error 0.022--0.025). The 8-bit microscaling format MXFP8~\cite{rouhani2023mx} and the distilled TAEHV VAE decoder~\cite{boerbohan2025taehv} bring it to 9.0\,s, and to 4.7\,s at 4 steps. Every Sol-Engine setting generates a 15\,s clip with sound in less time than it plays, and the 4-step configuration is 3.2$\times$ faster than real time.

\section{Reference-Conditioned Audio--Video Generation}
\label{sec:ref2av}

Text alone cannot pin down a specific person, product, place or voice.
\ours therefore extends the joint model to \emph{Ref2AV}: generate video and its sound from a prompt together with several reference assets, such as photos of the characters, an object or a location, a source clip or a voice, with every asset bound to the subject it defines.

\begin{figure}[t]
\centering
\begin{subfigure}[b]{0.684\linewidth}
\centering
\begin{adjustbox}{width=\linewidth}
\input{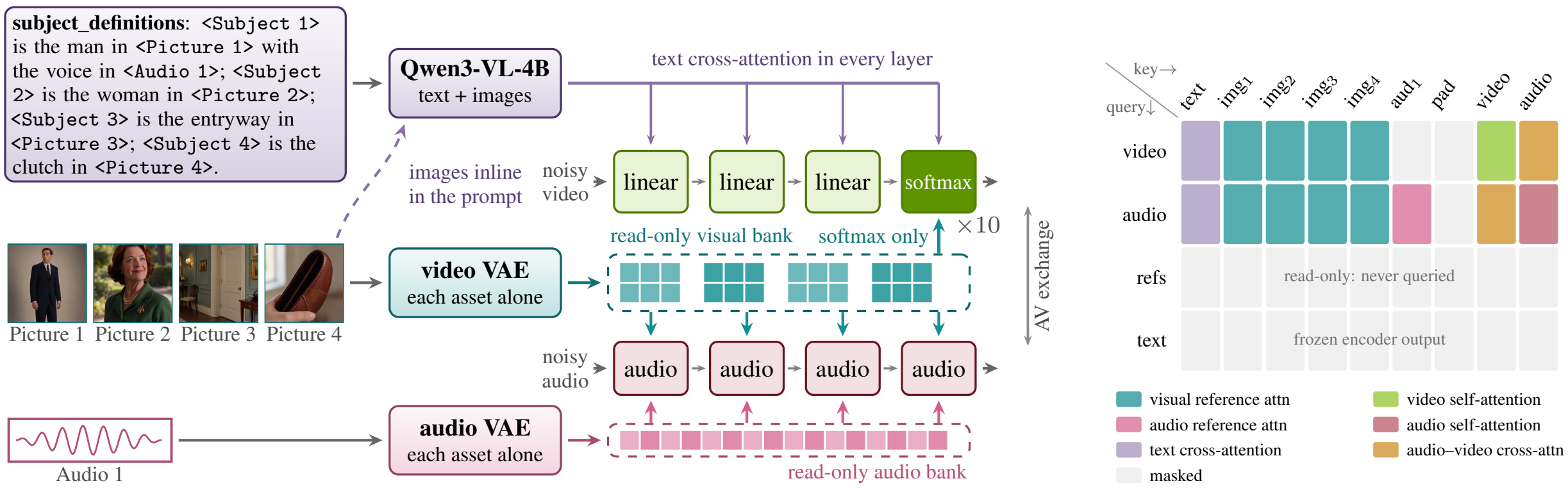}
\end{adjustbox}
\caption{Reference paths into the video and audio streams, in computation order.}
\end{subfigure}\hfill
\begin{subfigure}[b]{0.316\linewidth}
\centering
\begin{adjustbox}{width=\linewidth}
\definecolor{refc}{HTML}{138D90}
\definecolor{txtc}{HTML}{6A4C93}
\definecolor{vidc}{HTML}{76B900}
\definecolor{audc}{HTML}{A31F34}
\definecolor{xc}{HTML}{C77C02}
\definecolor{arfc}{HTML}{D45D8C}
\begin{tikzpicture}[font=\footnotesize]
  \def\cw{0.6}\def\ch{0.9}
  \foreach \j/\lab in {0/text, 1/img$_1$, 2/img$_2$, 3/img$_3$, 4/img$_4$, 5/aud$_1$, 6/pad, 7/video, 8/audio}
    \node[rotate=50, anchor=south west, inner sep=1pt] at (\j*\cw+0.12,4*\ch+0.05) {\lab};
  \foreach \i/\lab in {0/video, 1/audio, 2/refs, 3/text}
    \node[anchor=east] at (-0.06,3.5*\ch-\i*\ch) {\lab};
  \draw[black!40, line width=0.5pt] (-1.05,4*\ch+0.8) -- (-0.02,4*\ch+0.02);
  \node[font=\scriptsize, text=black!60, anchor=south west, inner sep=1pt] at (-1.08,4*\ch+0.0) {query$\downarrow$};
  \node[font=\scriptsize, text=black!60, anchor=north east, inner sep=1pt] at (0.02,4*\ch+0.85) {key$\rightarrow$};
  \def\T{t}\def\R{r}\def\V{v}\def\A{a}\def\X{x}\def\Q{q}
  \foreach \i/\row in {0/{t,r,r,r,r,n,n,v,x}, 1/{t,r,r,r,r,q,n,x,a}, 2/{n,n,n,n,n,n,n,n,n}, 3/{n,n,n,n,n,n,n,n,n}} {
    \foreach \c [count=\j from 0] in \row {
      \edef\cc{\c}
      \def\fillc{black!6}
      \ifx\cc\T \def\fillc{txtc!45}\fi
      \ifx\cc\R \def\fillc{refc!72}\fi
      \ifx\cc\Q \def\fillc{arfc!70}\fi
      \ifx\cc\V \def\fillc{vidc!60}\fi
      \ifx\cc\A \def\fillc{audc!55}\fi
      \ifx\cc\X \def\fillc{xc!65}\fi
      \fill[\fillc, rounded corners=1pt] (\j*\cw+0.025, 3*\ch-\i*\ch+0.025) rectangle ++(\cw-0.05, \ch-0.05);
    }
  }
  \node[font=\scriptsize, text=black!55] at (4.5*\cw,1.5*\ch) {read-only: never queried};
  \node[font=\scriptsize, text=black!55] at (4.5*\cw,0.5*\ch) {frozen encoder output};
  \foreach \x/\y/\c/\t in {-0.9/-0.38/refc!72/{visual reference attn}, -0.9/-0.74/arfc!70/{audio reference attn}, -0.9/-1.1/txtc!45/{text cross-attention}, -0.9/-1.46/black!6/{masked},
                           2.75/-0.38/vidc!60/{video self-attention}, 2.75/-0.74/audc!55/{audio self-attention}, 2.75/-1.1/xc!65/{audio--video cross-attn}} {
    \fill[\c, rounded corners=1pt] (\x,\y-0.11) rectangle ++(0.36,0.22);
    \node[anchor=west, font=\scriptsize, inner sep=2pt] at (\x+0.4,\y) {\t};
  }
\end{tikzpicture}
\end{adjustbox}
\caption{Attention map of one video--audio layer pair.}
\end{subfigure}
\caption{\textbf{Ref2AV reads references through a semantic path and read-only banks.} Qwen3-VL reads the images inline with the prompt. The video softmax layers read the visual bank and every audio layer reads both banks, all through zero-gated attention.}
\label{fig:ref2av}
\end{figure}

\paragraph{Task and prompt.}
A Ref2AV prompt follows the reference format of the H3 prompt-writing skill~\cite{minimax2026h3skill} and has six sections: subject definitions, summary, retention analysis, a time-ordered detailed description, the overall soundscape and the non-diegetic music.
Assets are cited with typed tags (\texttt{<Picture N>}, \texttt{<Video N>}, \texttt{<Audio N>}) and bound to \texttt{<Subject N>} entities, possibly many-to-many (one subject shown in two photos, or one photo defining a person and the outfit they wear), and speakers are tied to subjects for dialogue.
A reference image determines \emph{what} appears, and its subject may then be shown in any pose, viewpoint or moment.

\paragraph{Reference paths.}
Semantic binding and appearance detail need different representations (\cref{fig:ref2av}a).
On the \emph{semantic path}, the reference images are given to Qwen3-VL inline with the prompt, so the multi-layer features that already condition both streams know which subject each tag denotes.
On the \emph{appearance path}, each image is encoded on its own as a one-frame latent by the video VAE at its own resolution and aspect ratio (a short edge of 1{,}024 or 2{,}048 pixels in training and 2{,}048 at inference), and each reference clip is encoded on its own in the same way. Their patches form a visual bank of image and clip references.
Bank tokens carry their own spatial coordinates and an asset index in a separate reference coordinate space, and clip tokens also carry their own clock, so the generated video keeps its 3D RoPE and the audio--video clock unchanged, and a reference is never mistaken for a frame of the target.
Each reference sound is encoded on its own by the audio VAE into an audio bank with its own clock.

\begin{figure}[t]
\centering
\begin{subfigure}{\linewidth}
\centering
\includegraphics[width=\linewidth]{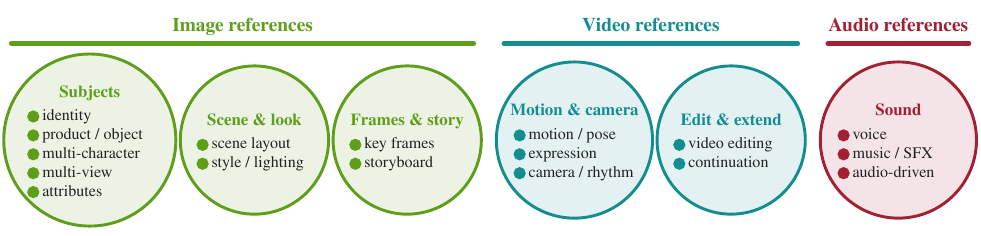}
\caption{Seventeen reference-conditioning tasks in six groups across image, video and audio references.}
\label{fig:ref2av_tax}
\end{subfigure}\par\medskip
\begin{subfigure}{\linewidth}
\centering
\includegraphics[width=\linewidth]{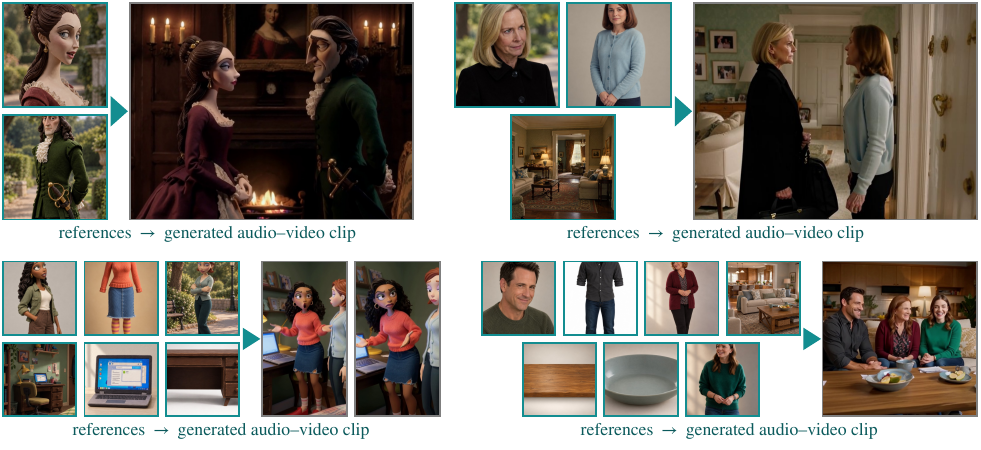}
\caption{Reference images composed into one coherent shot on held-out prompts.}
\label{fig:ref2av_qual}
\end{subfigure}
\caption{\textbf{Ref2AV defines seventeen reference-conditioning tasks (a) and composes reference images into one coherent shot on held-out prompts (b).} In (b), left of each arrow: references in prompt order. Right: frames of the generated clip with sound.}
\end{figure}

\paragraph{Read-only reference attention.}
Both banks are computed once per sample and reused, unchanged, by every layer, denoising step and guidance pass.
The visual bank is read by the ten video softmax layers, where video queries attend to it through an extra attention with its own normalisation and a zero-initialised gate (\cref{fig:ref2av}b). The 30 linear layers and the original softmax attention do not read the references.
Each of the 40 audio layers has two such readers. One reads the audio bank after the audio self-attention, so a reference voice or sound shapes the generated audio, and one reads the visual bank after the text attention, so the sound can follow what the references show. Both act before the audio--video exchange.
References never receive attention, FFN, \attnres or audio--video updates, so the reference features stay fixed throughout denoising, and zero-initialised gates preserve the joint model's output at the start of training.

\paragraph{Capabilities.}
We organise reference-conditioned generation into 17 tasks in six groups over image, clip and sound references (\cref{fig:ref2av_tax}): subjects (identity, products, several interacting characters, multi-view and attribute binding such as a person and their outfit), scene and look, frames and storyboards, motion and camera, editing and continuation of a source video, and sound (a voice, music or effects, and an existing audio track to which the picture must be synchronised).
Training starts from the stage-3 joint model and fine-tunes all denoiser weights with the text encoder and both VAEs frozen.
On held-out prompts (\cref{fig:ref2av_qual}), characters, outfits, objects and rooms from reference images are composed into one shot, each subject keeps its identity and clothing throughout, and the generated sound follows the prompt.

\section{Related Work}

\paragraph{Efficient video diffusion transformers.}
Video generation has moved from cascaded and latent U-Net diffusion~\cite{ho2022imagenvideo,singer2022makeavideo,blattmann2023svd,chen2024videocrafter2,bartal2024lumiere} to large diffusion transformers~\cite{openai2024sora,yang2024cogvideox,kong2024hunyuanvideo,zheng2024opensora,peng2025opensora2,lin2024opensoraplan,genmo2024mochi,hacohen2024ltxvideo,nvidia2025cosmos,chen2025goku,ma2025stepvideo,seawead2025seaweed,sandai2025magi1,chen2025skyreelsv2,gao2025seedance,wiedemer2025veo3}.
Recent large video and audio--video generators~\cite{polyak2024moviegen,wan2025wan,wu2025hunyuanvideo15,longcat2025video,hacohen2026ltx2,seedance2026seedance20advancingvideo,minimax2026h3} target increasingly long or high-resolution clips. Where a layer applies full softmax attention over all spatio-temporal tokens, its attention cost grows quadratically with the number of tokens.
Linear attention reduces this cost in high-resolution image synthesis~\cite{xie2024sana}, minute-scale and block-causal video~\cite{wang2024lingen,chen2025sanavideo}, and world modelling~\cite{zhu2026sanawm}. \prev~\cite{chen2026sanavideo2} matches full-softmax quality by combining gated linear layers~\cite{yang2025gdn,yang2025kimi,xi2026videodeltanet}, periodic softmax anchors as in hybrid language models~\cite{qwen3next}, and attention residuals across depth~\cite{kimiteam2026attnres}.
At inference, sparse attention~\cite{li2026solattn}, full-stack engines~\cite{solengine,li2026solh3}, one-step refiners~\cite{liu2026solrefiner} and step reduction~\cite{zhang2025accvideo,zou2026hiar} cut the cost further.
\ours keeps this backbone and scales it to 15\,s 768p clips, a second modality and large-scale joint training.

\paragraph{Joint audio--video generation.}
Early joint models denoise two coupled U-Nets~\cite{ruan2023mmdiffusion} or align pretrained single-modality generators through a shared latent aligner~\cite{xing2024seeinghearing}.
Recent large models fall into two families.
Dual-stream models keep a network per modality and exchange information through cross-modal attention, either between twin backbones~\cite{low2025ovi}, between asymmetric video and audio streams~\cite{hacohen2026ltx2}, or between pretrained experts stitched together~\cite{wang2025universe1}, and synchronisation priors can be added explicitly~\cite{liu2025javisdit}.
Single-stream models pack video and audio tokens into one sequence and share attention, optionally with modality-specific weights~\cite{liang2024mot,minimax2026h3}.
Our dual-stream design keeps the video stream unchanged and gives the best text--audio agreement in our controlled study (\cref{sec:arch:study}).
To our knowledge, ours is the first joint audio--video model on a hybrid linear-attention backbone.

\paragraph{Reference-conditioned video generation.}
Subject-driven generation started from image adapters that inject an image embedding through extra cross-attention~\cite{ye2023ipadapter}.
For video, identity-preserving models condition on a face~\cite{yuan2025consisid}, and multi-subject models concatenate reference latents with the video tokens~\cite{liu2025phantom,fei2025skyreelsa2}, add them through context adapters~\cite{jiang2025vace}, or fuse them with a multimodal language model~\cite{hu2025hunyuancustom}.
Ref2AV uses separately encoded, read-only visual and audio banks. Video softmax layers read the visual bank, while every audio layer reads both banks through zero-gated attention. A vision--language model binds each asset to a subject and speaker, so their sound is generated as well.

\paragraph{Few-step distillation.}
Diffusion and flow-matching models~\cite{lipman2022flowmatching,liu2022rectifiedflow} are distilled progressively~\cite{salimans2022progressive}, by score distillation~\cite{luo2023diffinstruct,zhou2024sid}, and, for video, with distribution matching~\cite{yin2023dmd,yin2024dmd2,yin2024causvid,huang2025selfforcing,yang2025longlive,chen2026longlive2,gu2026anyflow,lu2025rewardforcing,jiang2025dmdr,luo2026afd}, consistency~\cite{song2023cm,lu2024scm,chen2025sanasprint,zheng2025rcm} or adversarial objectives~\cite{goodfellow2014gan,sauer2023add,sauer2024ladd,lu2025adm}.
Reward fine-tuning and preference optimisation~\cite{clark2023draft,xu2023imagereward,wallace2023diffusiondpo,wu2025densedpo,fu2026graftood} are complementary.
DMD students can lose motion or diversity, visible in dynamics benchmarks~\cite{liao2024devil,zheng2025vbench2}, motivating changes to schedules, divergences and critics~\cite{fan2025phaseddmd,wu2026sgmd,xu2026dymd,wang2026pdmd,chen2026dfd,wu2026dpdmd,zhang2026codmd,you2026avd,xue2025mogan,huang2026dynaforcing}. Few-step MiniMax-H3 variants retain the teacher's family and latent space~\cite{fastvideo2026fasth3,lightx2v2026h3turbo}.
Latent adapters can align a representation to a semantic teacher~\cite{nutzel2026bilipschitz} or connect heterogeneous generators through a shared interface~\cite{dai2026instantfusion}. Cross-space distillation uses bridges to score the student~\cite{nguyen2026crossspace,zhan2026duomatching}, distils heterogeneous teachers on policy~\cite{fu2026anyopd,fu2026polyopd,xue2026fpopd}, or rescales the bridged guidance~\cite{jiang2026mild}. The standard backpropagation route through a bridge is a plain vector--Jacobian product.
The same product carries a diffusion score through a generator or decoder in score distillation~\cite{poole2022dreamfusion,wang2022sjc,wang2023prolificdreamer,richterpowell2025audiosds} and latent inverse problems~\cite{graikos2024newton,bai2025instantvir,wu2026hdps}, where the decoder Jacobian distorts the update~\cite{pan2023pgc,skorokhodov2026pixsds}. DDRep~\cite{savani2024ddrep} is the closest precedent: it pulls a diffusion score back through a renderer with $(J^\top J)^{-1}J^\top$ to sample a single representation, without damping and without training a generator.
We measure how much of the teacher's update this product realises and correct it with a damped Gauss--Newton pullback~\cite{martens2014ggn}. Related work also distinguishes the accuracy of a learned map from that of its derivatives~\cite{oleary2022dino}. The pullback is related to target propagation and to Gauss--Newton credit assignment in neural networks~\cite{bengio2020dtp,meulemans2020tp,meulemans2021dfc,roulet2021tp,castiglione2026cgtp,dumont2026gndrifting}.

\section{Conclusion}

\ours turns a hybrid linear-attention backbone into a joint audio--video generator that runs in seconds. A time-aligned parallel audio stream adds sound, staged pre-training trains the audio stream before the video weights are unfrozen, and few-step distillation from a stronger teacher in another latent space works once its update is pulled back faithfully. In our motion probes, one-step DMD targets add little of the motion that longer teacher rollouts recover. The 4- and 8-step students both outperform their 40-step parent and the 50-step teacher on HPSv3. The 4-step student takes about $1/45$ of the teacher's time per short clip. On one GB200 the 8-step student generates a 15\,s $1344{\times}768$ clip with audio in 14.7\,s, and in 9.0\,s with Sol-Engine. Its measured latency advantage over LTX-2.5 grows with duration: from 5 to 15\,s the student's time grows 2.9$\times$ against 3.9$\times$ for LTX-2.5, and the fastest 4-step Sol-Engine setting generates a 15\,s clip in 4.7\,s. Fine-tuned with read-only reference banks, it also generates video and sound from multiple reference images, clips and sounds bound to subjects in the prompt, such as a character's face, outfit and voice.

\paragraph{Limitations and future work.}
Future work will further improve complex limb motion, legible on-screen text and overall realism, and extend generation beyond 15\,s clips.
Ref2AV can be taken further with a quantitative benchmark over the 17 reference tasks of \cref{fig:ref2av_tax} and few-step distillation of the reference model to run in seconds.

\section*{Acknowledgements}
We thank NVIDIA colleagues Yuyang Zhao for training infrastructure support and Tian Ye for discussions on DMD. We also thank Sensen Gao for discussions on refiner optimisation.

\clearpage
\AddToHook{cmd/thebibliography/after}{\parfillskip=.84\linewidth minus .84\linewidth\relax}
\begingroup
\addtolength{\bibsep}{-1pt}

\endgroup

\clearpage
\appendix
\section{Audio--Video Model and Pre-training Details}
\label{sec:supp:pretrain}

\subsection{Architecture and Alignment}
\label{sec:supp:av_arch}

\begin{table}[ht]
\centering
\small
\setlength{\tabcolsep}{5pt}
\caption{\textbf{Configurations of the 5B \prev model, the 14B video seed and the joint 16B model.} Video-only columns show the starting models before adapting the denoiser to Qwen3-VL and adding audio. The 5B model is the testbed (\cref{sec:supp:fewstep_extra}).}
\label{tab:config}
\begin{adjustbox}{max width=\linewidth}
\begin{tabular}{@{}l c c c@{}}
\toprule
 & \textbf{\prev 5B} & \textbf{14B video seed} & \textbf{\ours 16B AV} \\
\midrule
Video depth / width & 32 / 2560 & 40 / 4096 & 40 / 4096 \\
Linear / softmax layers & 24 / 8 & 30 / 10 & 30 / 10 \\
Linear / softmax heads & $20{\times}128$ / $10{\times}256$ & $32{\times}128$ / $16{\times}256$ & $32{\times}128$ / $16{\times}256$ \\
\attnres & \multicolumn{3}{c}{block size 8, depth-shared routing queries, no timestep conditioning} \\
Text encoder & Gemma-2-2B, 300 tok. & Gemma-2-2B, 300 tok. & Qwen3-VL-4B (6-layer features), 1024 tok. \\
Clips & 480--720p, 81--193 fr. & 480p, 81/121 fr. & 480--768p, 121/241/361 fr. (5--15\,s) \\
Conditioning & text & text & text + first/last frame \\
Audio stream & -- & -- & 40 layers, width 1408, waveform VAE \\
Parameters & 4.5B & 14.25B & 14B video + 2B audio \\
\bottomrule
\end{tabular}
\end{adjustbox}
\end{table}

\paragraph{Audio stream.}
The audio stream uses the frozen waveform-VAE latents of \cref{sec:arch}, at 40 tokens per second. Each of its 40 layers applies softmax self-attention, cross-attention to audio text features, exchange with the video layer of the same depth, and a GELU FFN (\cref{tab:config}). Both streams condition on six concatenated Qwen3-VL hidden-layer features through separate projections. Self-attention uses 22 heads of width 64 and RoPE in seconds. Every residual branch has its own AdaLN~\cite{peebles2023dit} scale, shift and gate.
In the cross-modal exchange, each direction uses its own modality's noise level for affine modulation and the other modality's for its residual gate. Both directions read the same snapshot of the two streams, taken before either stream is updated.

\paragraph{Codec timing check.}
To verify that the two codecs preserve timing, we encode and decode clips with flashes and clicks. The round trip keeps the video frame offset at 0 and local audio timing within about 2\,ms. The synchrony of generated clips is evaluated separately with Synchformer~\cite{iashin2024synchformer} (\cref{sec:supp:eval}).

\subsection{Pre-training Details}

\paragraph{Task mix and text dropout.}
The task mix combines text-to-audio--video with first-frame, last-frame and first-and-last-frame conditioning. Text dropout for classifier-free guidance~\cite{ho2022cfg} is drawn once per sample: video text only (5\%), audio text only (5\%), both (5\%) or neither (85\%).

\paragraph{Optimisation.}
Stages 0--2 use a constant rate of $10^{-4}$. In stage 3, AdamW~\cite{loshchilov2019adamw} uses peak rates of $2{\times}10^{-5}$ for video and $5{\times}10^{-5}$ for audio, both halved on a shared cosine schedule. An FP32 EMA (decay 0.999) covers both streams.

\paragraph{Training system.}
We use BF16 with FSDP~\cite{zhao2023fsdp}. The text encoder and VAEs run online during training, so no latent cache is needed. One duration--resolution bucket per step gives ranks one shape without padding.

\subsection{Sampler Search}
\label{sec:supp:sampler}

Diffusion solvers have advanced considerably: ODE solvers include DDIM~\cite{song2021ddim}, DPM-Solver~\cite{lu2022dpmsolver}, DPM-Solver++~\cite{lu2022dpmsolverpp} and UniPC~\cite{zhao2023unipc}, and SDE solvers include the stochastic sampler of EDM~\cite{karras2022edm} and SA-Solver~\cite{xue2023sasolver}.
The default 40-step sampler with guidance on both streams needs 120 network evaluations per clip. To reduce this cost without losing quality, we searched solver, steps, shift, CFG scale and variant, and perturbation guidance using reward models, VBench dimensions, audio metrics and a multimodal judge. Finalists underwent blind side-by-side review on held-out prompts because rewards distinguished Euler from SDE more reliably than close SDE variants.
The search used an intermediate base checkpoint. \Cref{tab:supp:sampler_results} reports the finalists with their raw scores on held-out prompts. The selected recipe uses 30-step SDE sampling, a joint unconditional pass, CFG rescale~\cite{lin2024flawed} and interval guidance~\cite{kynkaanniemi2024interval}.

\begin{table}[!htbp]
\centering
\small
\setlength{\tabcolsep}{3.5pt}
\caption{\textbf{Sampler finalists on an intermediate base checkpoint} (36 held-out prompts, 720p, 5--15\,s). Joint: one joint unconditional pass. Int.: interval guidance. Time: denoising seconds per clip on one B200. Judge Score: multimodal judge overall score. Production Quality: Audiobox Aesthetics PQ. Gray: real clips. Bold: best generated. Highlighted: adopted.}
\label{tab:supp:sampler_results}
\begin{adjustbox}{max width=\linewidth}
\begin{tabular}{@{}l c r r c c c c@{}}
\toprule
\textbf{Sampler} & \textbf{Steps} & \textbf{NFE} & \textbf{Time (s)} & \textbf{Judge Score} & \textbf{HPSv3} & \textbf{VideoReward} & \textbf{Production Quality} \\
\midrule
\textcolor{black!60}{Real clips} & & & & \textcolor{black!60}{8.67} & \textcolor{black!60}{9.04} & \textcolor{black!60}{$-0.08$} & \textcolor{black!60}{7.20} \\
Euler (default) & 40 & 120 & 305 & 8.42 & 8.65 & $-0.03$ & 6.85 \\
SDE & 40 & 120 & 304 & 8.49 & 8.78 & \textbf{0.11} & \textbf{7.11} \\
SDE, joint & 50 & 100 & 254 & 8.54 & 8.76 & 0.10 & 7.09 \\
SDE, joint & 30 & 60 & 151 & \textbf{8.55} & 8.80 & 0.08 & 7.09 \\
SDE, rescale+int. & 30 & 85 & 216 & 8.50 & \textbf{8.83} & 0.07 & 6.97 \\
\rowcolor{nvidiagreen!15}SDE, all (ours) & 30 & \textbf{58} & \textbf{149} & 8.53 & 8.77 & 0.08 & 7.06 \\
\bottomrule
\end{tabular}
\end{adjustbox}
\end{table}

\paragraph{Adopted sampler on the final base model.}
\Cref{tab:pretrain_eval} repeats the comparison on the final base model. The selected recipe \keyres{halves the network evaluations (58 instead of 120) and improves HPSv3, VideoReward, production quality and CLAP}. The parent in \cref{tab:main} uses 40 steps with video guidance only (80 NFE). Sampler times here are denoising-only on one B200, unlike the end-to-end GB200 timings that are reported in \cref{sec:supp:deployment}.

\begin{table}[!htbp]
\centering
\small
\setlength{\tabcolsep}{5pt}
\caption{\textbf{Default and tuned samplers on the same base checkpoint} (36 held-out prompts, 720p, 5--15\,s). Gray: real clips, for reference. Judge Score: multimodal judge score. Production Quality: Audiobox Aesthetics PQ. ASR Accuracy: Whisper speech-recognition accuracy. All columns are higher-is-better. Bold: best generated scores.}
\label{tab:pretrain_eval}
\begin{adjustbox}{max width=\linewidth}
\begin{tabular}{@{}l c c c c c c@{}}
\toprule
& \textbf{Judge Score} & \textbf{HPSv3} & \textbf{VideoReward} & \textbf{Production Quality} & \textbf{CLAP Score} & \textbf{ASR Accuracy} \\
\midrule
\textcolor{black!60}{Real clips} & \textcolor{black!60}{8.67} & \textcolor{black!60}{9.04} & \textcolor{black!60}{$-$0.08} & \textcolor{black!60}{7.20} & \textcolor{black!60}{0.136} & \textcolor{black!60}{0.974} \\
Default, 120 NFE & 8.47 & 8.66 & 0.01 & 6.88 & 0.149 & \textbf{0.872} \\
\rowcolor{nvidiagreen!15}Tuned, 58 NFE & \textbf{8.50} & \textbf{8.71} & \textbf{0.30} & \textbf{7.11} & \textbf{0.167} & 0.851 \\
\bottomrule
\end{tabular}
\end{adjustbox}
\end{table}

\FloatBarrier

\section{Few-Step Distillation: Analyses}
\label{sec:supp:fewstep_extra}

\subsection{Where the Students' Motion Comes From}
\label{sec:motion}

We measure \emph{effective motion} as coherent optical flow after removing jitter. The \emph{allocation slope} is the regression slope of a model's per-prompt log motion on the teacher's log motion. A slope of 1 means proportional allocation across prompts, and the intercept sets the overall amount. We use both to trace where the 5B student's motion comes from.

\paragraph{One-step DMD target.}
\Cref{fig:motion}c,d plot the motion of each signal relative to the student against the noise level, on 8 large-motion and 4 static prompts. On the large-motion prompts, the teacher's own videos move about 3$\times$ as much as the student. The teacher's and critic's one-step clean predictions are conditional means~\cite{efron2011tweedie}, with aligned motion components (cosine 0.55--0.89). Much of this shared motion cancels in their difference, the correction $g$ of \cref{eq:dmd}. The resulting target in teacher space, $z-g$, with $z=\T(x)$ in cross-latent DMD, retains only 0.73--0.90$\times$ the student's coherent motion at every noise level we tested and also adds frame-to-frame jitter to the target.

\paragraph{Multi-step rollout probe.}
When the teacher denoises the noised student sample over 30 steps from $\sigma_0\ge0.99$, its output moves 1.4--2.1$\times$ as much as the student on dynamic prompts and removes most drift on static prompts (\cref{fig:motion}d). Direct regression onto these rollouts raises the student's motion from 0.27$\times$ to 0.83$\times$ the teacher's. \takeaway{The student can represent the missing motion, but the one-step DMD target does not supply it as a learning signal.}

\paragraph{Implication for distillation.}
The 14B video stream of the 16B student shows the same pattern, with target motion at 0.90--1.00$\times$ the student's for $\sigma\ge0.98$. Rollout regression serves as a capacity probe. The final recipe uses one-step DMD, starts from a parent that already moves well and uses smooth high-motion teacher videos as GAN reals.

\begin{figure*}[ht]
\centering
\includegraphics[width=\linewidth]{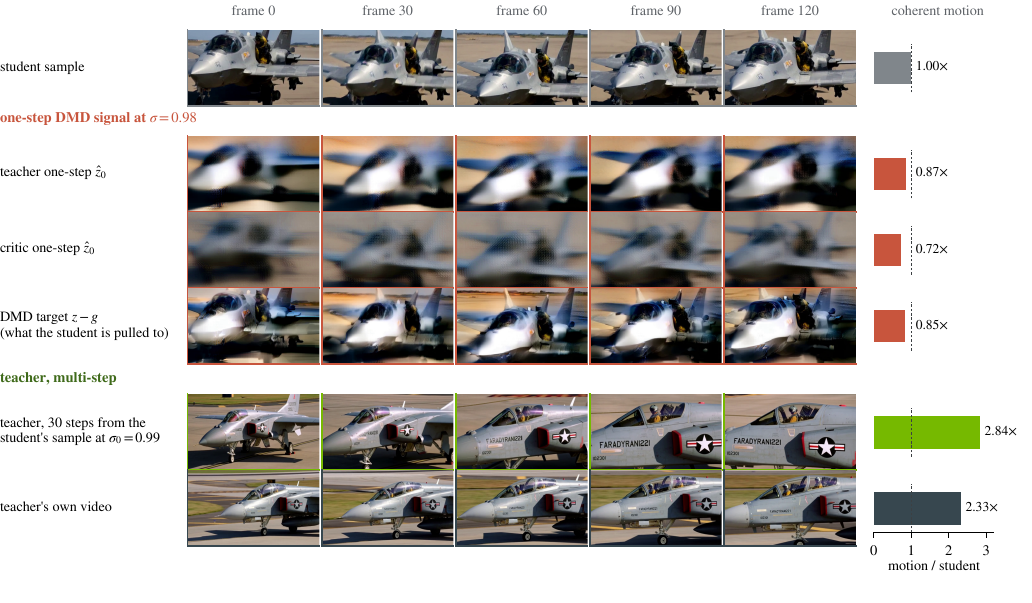}
\caption{\textbf{The teacher's multi-step rollout recovers motion missing from the one-step DMD target} (a fighter jet taxiing). Red: at $\sigma=0.98$ the one-step predictions are blurred, and the DMD target moves no more than the student (0.85$\times$). Green: the teacher's 30-step rollout from the same noise moves 2.8$\times$ as much, like the teacher's own video (2.3$\times$).}
\label{fig:supp:chain_strip}
\end{figure*}

\section{Evaluation Protocol}
\label{sec:supp:eval}

\paragraph{Prompts and baselines.}
The HPSv3, VideoReward and audio scores in \cref{tab:main,tab:audio} use the same 256 prompts, one generated clip per prompt and the prompt index as seed. Audio--video models receive the structured prompts of~\cite{minimax2026h3skill}.
MiniMax-H3 uses 50 steps. H3-Turbo uses LightX2V's 4-step v0.1 LoRA~\cite{lightx2v2026h3turbo}, whose released schedule requires 5 network evaluations, reported as 5 in \cref{tab:main,tab:audio,tab:vbench}. FastH3 uses FastVideo's 4-step Preview-v1 dense LoRA~\cite{fastvideo2026fasth3}. All three run MiniMax-H3's official diffusers pipeline at $1344{\times}768$ with audio. \Cref{fig:deployment} instead times an 8-step LightX2V Turbo configuration. LTX-2.5 distilled uses its official 8-step pipeline at the same resolution with audio, without the spatial upsampler or second stage.
Video-only baselines are Wan2.2-A14B with Lightning at $1280{\times}720$ and HunyuanVideo-1.5 4-step at $848{\times}480$, both with 81 frames. The short-clip timings use 121 frames for our model and LTX-2.5, and 124 for \Hthree. Our 40-step parent uses video guidance (80 NFE), while the 4- and 8-step students are sampled without classifier-free guidance.

\paragraph{Audio metrics.}\label{sec:supp:audio_ci}
For scoring, audio is mixed to mono and resampled to 48\,kHz, and the first 5\,s are evaluated. This scoring rate is separate from the model's 32\,kHz stereo waveform representation.
PQ is Audiobox Aesthetics production quality~\cite{tjandra2025audiobox}. CLAP~\cite{wu2023clap} compares audio with the prompt's sound description, while IB measures ImageBind~\cite{girdhar2023imagebind} audio--video similarity. IS is the PaSST~\cite{koutini2022passt} Inception Score, and DeSync is the audio--video offset in seconds predicted by Synchformer~\cite{iashin2024synchformer}.
For the 8-step student minus LTX-2.5, paired bootstrap 95\% intervals give CLAP $+0.057$ $[+0.043,+0.073]$ and DeSync $-0.21$ $[-0.32,-0.10]$\,s, both excluding zero.

\paragraph{Standard-prompt VBench.}
The main VBench~\cite{huang2023vbench} comparison uses 946 extended prompts, 5 samples per prompt and all 16 dimensions. The re-tested audio--video models, including ours, generate from H3-formatted rewrites of these prompts~\cite{minimax2026h3skill}, while VBench scores each clip against the original benchmark prompt. \Cref{tab:vbench} reports official Total, Quality and Semantic scores. Published Wan2.1-based methods keep their original settings. The 5B pullback ablation uses the same prompts with one sample each.

\newcommand{\vbenchrows}{
\toprule
\textbf{Model} & \textbf{Steps} & \textbf{Total}$\uparrow$ & \textbf{Quality}$\uparrow$ & \textbf{Semantic}$\uparrow$ \\
\midrule
Self-Forcing~\cite{huang2025selfforcing} (1.3B) & 4 & 84.31 & 85.07 & 81.28 \\
rCM~\cite{zheng2025rcm} (14B) & 4 & 84.92 & 85.43 & 82.88 \\
TMD~\cite{nie2026tmd} (14B) & 2.75 & 84.92 & 85.54 & 82.42 \\
PTD~\cite{feng2026ptd} (14B) & 4 & 84.57 & 85.44 & 81.10 \\
DMAD~\cite{yu2026dmad} (14B) & 4 & 85.15 & 85.96 & 81.90 \\
\midrule
\Hthree-Turbo~\cite{lightx2v2026h3turbo} (33B) & 5 & 85.07 & 86.01 & 81.33 \\
FastH3~\cite{fastvideo2026fasth3} (33B) & 4 & 85.09 & 86.59 & 79.13 \\
Wan2.2-A14B~\cite{wan2025wan} + Lightning~\cite{lightx2v2025wan22lightning} (27B) & 4 & 82.35 & 83.87 & 76.27 \\
HunyuanVideo-1.5~\cite{wu2025hunyuanvideo15} 4-step~\cite{lightx2v2025hy15distill} (8.3B) & 4 & 81.11 & 83.97 & 69.66 \\
LTX-2.5 distilled~\cite{hacohen2026ltx2} (22B) & 8 & 83.37 & 84.79 & 77.69 \\
\rowcolor{nvidiagreen!15}ours, 16B & 40 & \textbf{85.50} & 86.43 & \textbf{81.79} \\
\rowcolor{nvidiagreen!15}\textbf{ours}, 16B & 8 & 85.39 & \textbf{86.73} & 80.03 \\
\rowcolor{nvidiagreen!15}\textbf{ours}, 16B & 4 & 85.19 & 86.40 & 80.37 \\
\bottomrule
}

\newcommand{\vbenchcompactbody}{{\footnotesize\setlength{\tabcolsep}{2.2pt}\begin{tabularx}{\linewidth}{@{}>{\raggedright\arraybackslash}X c c c c@{}}
\vbenchrows
\end{tabularx}}}
\newcommand{\vbenchcaption}{\textbf{VBench scores,} higher is better. Top: as published, on Wan2.1 at 480p. Bottom: re-tested with 5 seeds, best in bold, ours, LTX and \Hthree at $1344{\times}768$, Wan2.2 at $1280{\times}720$ and HunyuanVideo-1.5 at $848{\times}480$.}

\newcommand{\pbvcaption}{\textbf{VBench quality of the 5B 4-step students} of \cref{tab:pullback} on 256 anchor prompts. LP: low-pass VJP. Bold: best per row. Highlighted: GN pullbacks.}
\newcommand{\pbvbody}{{\footnotesize\renewcommand{\arraystretch}{\pbvstretch}\setlength{\tabcolsep}{2pt}%
\begin{tabularx}{\linewidth}{@{}>{\raggedright\arraybackslash}X rr >{\columncolor{nvidiagreen!15}}r >{\columncolor{nvidiagreen!15}}r@{}}
\toprule
\textbf{Dimension} & \textbf{VJP} & \textbf{LP} & \textbf{\GN{8}} & \textbf{\GN{32}} \\
\midrule
\rowcolor{catgray}\multicolumn{5}{@{}l}{\emph{Appearance}} \\
Aesthetic quality & 65.62 & 65.80 & \textbf{66.51} & 66.40 \\
Imaging quality & 70.65 & 70.61 & 71.75 & \textbf{71.95} \\
\midrule
\rowcolor{catgray}\multicolumn{5}{@{}l}{\emph{Consistency over time}} \\
Subject consistency & 94.43 & 94.54 & \textbf{94.90} & 94.73 \\
Background consistency & 95.92 & 96.03 & \textbf{96.24} & 96.23 \\
\midrule
\rowcolor{catgray}\multicolumn{5}{@{}l}{\emph{Motion}} \\
Temporal flickering & 97.02 & \textbf{97.07} & \textbf{97.07} & 97.01 \\
Motion smoothness & 99.12 & 99.12 & 99.19 & \textbf{99.21} \\
Dynamic degree & \textbf{86.33} & 84.77 & 80.08 & 83.98 \\
\midrule
VBench Quality & 85.72 & 85.68 & 85.76 & \textbf{86.02} \\
\bottomrule
\end{tabularx}}}
\providecommand{\pbvstretch}{1.08}

\begin{table}[!htbp]
\captionsetup{type=table}
\noindent\begin{minipage}[t]{0.56\linewidth}\vspace{0pt}
\caption{\vbenchcaption}\label{tab:vbench}
\end{minipage}\hfill
\begin{minipage}[t]{0.41\linewidth}\vspace{0pt}
\caption{\pbvcaption}\label{tab:supp:pullback_vbench}
\end{minipage}\par\vspace{2pt}
\noindent\begin{minipage}[t]{0.56\linewidth}\vspace{0pt}
\vbenchcompactbody
\end{minipage}\hfill
\begin{minipage}[t]{0.41\linewidth}\vspace{0pt}
\pbvbody
\end{minipage}
\end{table}

\paragraph{Standard-prompt pullback ablation.}
The four 5B arms in \cref{tab:pullback} continue from the same training state $\theta^\star$ under matched settings. Each arm generates one video per prompt on VBench's 946 extended prompts at $832{\times}480$, with 4 sampling steps, CFG 1 and identical seeds. All 16 dimensions use the official prompt subsets, normalisation and weights, and the common start $\theta^\star$ scores 84.97 Total, 86.58 Quality and 78.52 Semantic on this benchmark.

\paragraph{Anchor-set VBench scores.}
A separate set of 256 anchor prompts is rendered once per arm, with the prompt index as seed, and scored on the seven VBench~\cite{huang2023vbench} quality dimensions. \Cref{tab:supp:pullback_vbench} lists the raw scores (0--100, higher is better) in three groups, after matched continuations, and combines them into VBench Quality with the official weights. Both GN arms raise appearance and temporal consistency over VJP and low-pass VJP while temporal flickering and motion smoothness stay level with VJP, and \GN{32} reaches the highest overall Quality, 86.02. A paired bootstrap over prompts gives \GN{32} a Quality gain of $+0.299$ $[+0.012,+0.587]$ over VJP, with an interval that excludes zero.

\section{Latent Transfer and Pullback Details}
\label{sec:supp:pullback}

\paragraph{The adapter.}
The adapter $\T$ maps a student latent $x$ (128 channels, 16 latent frames) into the teacher's latent space (24 channels, 37 latent frames). In the 5B configuration, it projects to width 640, applies factorised spatio-temporal residual blocks, interpolates in time to the teacher's frame centres and upsamples $2{\times}$ in space. It is trained on paired encodings of the same videos by both VAEs and takes 45\,ms per clip, small next to the 33B teacher and critic passes. $\T^{+}$ in \cref{tab:pullback} is a more accurate adapter version, and the 16B adapter is described in \cref{sec:supp:14b}.

\paragraph{Notation.}
The DMD correction $g$ of \cref{eq:dmd} lives in teacher space, while the student can only change $x$. The adapter Jacobian $\J=\partial\T/\partial x$ at the current sample links the two spaces. A small student move $d$ changes the teacher latent by $\J d$, a forward Jacobian--vector product (JVP). The transpose $\J^\top$ maps a vector in teacher space back to student space, a vector--Jacobian product (VJP), which is what backpropagation through $\T$ computes. A good update $d$ satisfies $\J d\approx g$, and the realised fidelity $\rho(d)=\cos(\J d,g)$ measures how well its image points along $g$ (1: same direction).

\paragraph{Step 1: backpropagation distorts the direction.}
Backpropagating $g$ through $\T$ gives the update $b=\J^\top g$. Pushing $b$ forward through $\J$ gives its realised move $\J\J^\top g$, which is in general not parallel to $g$. The singular value decomposition of $\J$ shows why. Let $s_i>0$ be its nonzero singular values, with singular vectors $u_i$ in teacher space and $v_i$ in student space, and let $\gamma_i=u_i^\top g$ be the component of $g$ along $u_i$. The realised move of backpropagation is then
\begin{equation*}
\J\J^\top g=\sum_i s_i^2\,\gamma_i\,u_i .
\end{equation*}
Each component of $g$ is scaled by $s_i^2$, so the few high-gain directions of the adapter dominate the realised move. On 128 rollout samples, this gives $\rho=0.22$ (\cref{sec:pullback}).

\paragraph{Step 2: solve for the update whose forward image matches $g$.}
Instead of using $b$ directly, we look for the student update whose forward image is closest to $g$:
\begin{equation*}
\begin{gathered}
d_{\text{GN}}=\argmin_d\bigl(\lVert\J d-g\rVert^2+\lambda\lVert d\rVert^2\bigr),\\
\text{equivalently}\quad(\J^\top\J+\lambda I)\,d_{\text{GN}}=b .
\end{gathered}
\end{equation*}
The damping $\lambda>0$ keeps the solve from inverting tiny singular values. In the same basis, the realised move becomes
\begin{equation*}
\J d_{\text{GN}}=\sum_i\frac{s_i^2}{s_i^2+\lambda}\,\gamma_i\,u_i .
\end{equation*}
Directions with $s_i^2\gg\lambda$ retain nearly their full component of $g$, while directions with $s_i^2\ll\lambda$ are attenuated. As $\lambda\to0$, $\J d_{\text{GN}}$ becomes the projection of $g$ onto that space.

\paragraph{Step 3: matrix-free solve.}
$\J$ is never formed. Starting from zero, $k$ conjugate-gradient (CG) iterations solve the system of Step~2, and each applies $v\mapsto\J^\top(\J v)+\lambda v$ with one JVP and one VJP through $\T$. If $b=0$, the correction is zero. Otherwise we set $\lambda=0.01\lVert b\rVert^2/\lVert g\rVert^2$, so damping scales with $\J^\top\J$. Rescaling $g$ leaves $\lambda$ unchanged and rescales the solution by the same factor. All products run through the 142M-parameter adapter, never through the 33B teacher.

\paragraph{Step 4: norm matching and regression target.}
To change only the direction of the update, we rescale the CG iterate $d^{(k)}$ to the length of the backpropagated update, $\widetilde d_{\text{GN}}=(\lVert b\rVert/\lVert d^{(k)}\rVert)\,d^{(k)}$. The student then regresses onto the stopped target $x-\eta\widetilde d_{\text{GN}}$ with step size $\eta>0$. The first CG iterate is parallel to $b$, so $k=1$ recovers backpropagation after norm matching. Additional CG iterations approximate the damped solution of Step~2.

\paragraph{Why the update is safe.}
The matrix $\J^\top\J+\lambda I$ is positive definite for $\lambda>0$, so $\langle b,d_{\text{GN}}\rangle>0$ whenever $b\neq0$. The GN update thus never opposes the backpropagated one, and it vanishes only when $b$ does. It is also a combination of the $v_i$, so it never moves the student along the null space of $\J$, where a change of $x$ has no effect on the teacher latent.

\paragraph{Exact decode--re-encode comparison.}\label{sec:supp:exact}
A natural alternative to the learned adapter is the exact path that decodes with the student's VAE and re-encodes with the teacher's. On an eight-pair probe, separate from the 128 rollouts above, backpropagation through the exact VAE path gives $\rho=0.019$, compared with 0.187 through the learned adapter on the same pairs. Six GN iterations through the exact path reach 0.048. Its top singular value is about 100$\times$ larger, with distortion concentrated in isolated hot spots. \takeaway{The exact VAE path distorts the correction more than the learned adapter}, so all our students use the learned adapter.

\section{16B Student Details}
\label{sec:supp:14b}
The 16B student pairs a 14B video stream with a 2B audio stream. Training starts with 5\,s clips at 480p and continues at 768p in both orientations on 5, 10 and 15\,s clips. Both 4- and 8-step results use the same checkpoint.

\paragraph{Adapter for several durations and orientations.}
The 5B adapter is trained on 5\,s landscape clips. For the 16B student, we continue it on 5, 10 and 15\,s clips in both orientations. One adapter covers all evaluated buckets with held-out reconstruction cosine 0.983--0.984. Update fidelity through the adapter is measured separately below.

\paragraph{Objective and updates.}
The student minimises $\mathcal{L}=\mathcal{L}^{v}_{\text{DMD}}+\mathcal{L}^{a}_{\text{DMD}}+0.01\,\mathcal{L}_{\text{GAN}}$. A joint \Hthree teacher evaluation supplies both video and audio teacher predictions at the paired noise levels of \cref{sec:pretrain}. The online critic (\Hthree with a rank-64 LoRA) supplies the fake predictions in \cref{eq:dmd} and receives five updates per generator update. The GAN head reads frozen-teacher features, with smooth, high-motion teacher videos as reals.

\paragraph{Pullback fidelity at 16B.}
On the student's rollout probes, VJP gives $\rho=0.33$, compared with 0.71 for \GN{8} and 0.88 for \GN{32} (5B: 0.22, 0.62 and 0.83). The gain holds at every tested noise level and in both orientations.
The VJP correction puts 70\% of its energy in the lowest spatial-frequency bin, compared with 19\% for $g$. \GN{32} reduces this share to 20\%, consistent with the 5B spectral diagnosis.

\section{Deployment Details}
\label{sec:supp:deployment}

\paragraph{Timing protocol.}
We time text encoding, denoising and decoding on one GB200 (warm, batch one). For 5, 10 and 15\,s at 24\,fps, the student and LTX-2.5 use 121/241/361 frames, and \Hthree-Turbo uses 124/243/345. Baselines use medians of six requests on three prompts. The 8-step engine sweep alternates settings on one node, with twelve requests each.

\paragraph{Baseline implementations.}
LTX-2.5 uses its official 8-step sampler, compiled native transformer and convolutional decoder without the upsampler or second stage. Decoding takes 0.8--2.7\,s. On the student, PyTorch falls back to FlashAttention-2 at head dimension 256. Switching to FlashAttention-4 alone saves 0.5, 2.3 and 4.8\,s at 5, 10 and 15\,s. \Cref{sec:deployment} reports the combined gains from attention and the other optimisations~\cite{li2026solh3}.

\paragraph{Sol-Engine comparison.}
At baseline, power-limited GEMMs take 43--48\% of denoising time, pointwise operations about 40\%, and attention 8--18\%. Sol-Engine fuses normalisation, modulation, gates and residuals with unchanged rounding, folds activations into GEMM epilogues, captures FlashAttention-4 cross-attention in a CUDA graph per clip length and decodes without tiling. On identical inputs, both BF16 versions have relative velocity error 0.022--0.025 against the native model. The fastest setting adds MXFP8 GEMMs~\cite{rouhani2023mx} and the TAEHV decoder~\cite{boerbohan2025taehv} (\cref{fig:engine}).

\section{Additional Qualitative Results}
\Cref{fig:supp:t2av,fig:supp:multishot} show 5--15\,s samples. \Cref{fig:supp:cmp5,fig:supp:cmp_long} compare with \Hthree and LTX-2.5 distilled on matched prompts and seeds. Selected clips can be played with sound on the project page.

\newcommand{\qstudent}{16B}
\begingroup
\newcounter{nvqualfigure}
\renewenvironment{figure*}[1][]{%
  \stepcounter{nvqualfigure}%
  \ifnum\value{nvqualfigure}=1
    \begin{figure}[H]
    \renewcommand{\qgallerywidth}{0.87\linewidth}
  \else
    \begin{figure}[p]
  \fi
}{%
  \end{figure}%
}
\providecommand{\qstudent}{16B}
\providecommand{\qgallerywidth}{\textwidth}
\newcommand{\qsep}[1]{{\setlength{\fboxsep}{1.5pt}\colorbox{nvidiagreen!22}{\parbox{\dimexpr\qgallerywidth-2\fboxsep\relax}{\centering\small\textbf{#1}}}}\\[0.5pt]}

\begin{figure*}[p]
\centering
\captionsetup{skip=4pt}
\qsep{8-step student}
\includegraphics[width=\qgallerywidth]{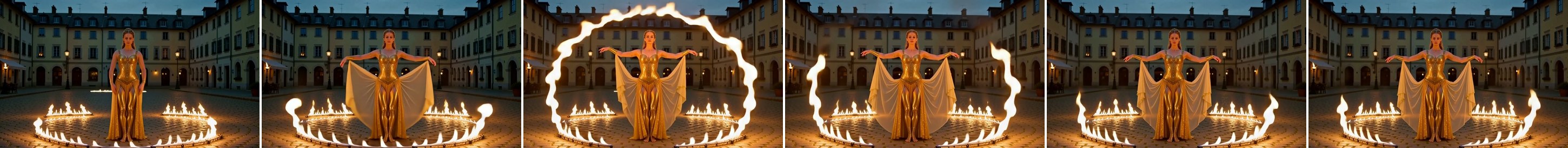}\\[-1pt]
  \parbox{\qgallerywidth}{\scriptsize\itshape A dancer in a golden bodysuit spins in a cobblestone square at dusk as a ring of fire rises around her.}\\[0.25pt]
\includegraphics[width=\qgallerywidth]{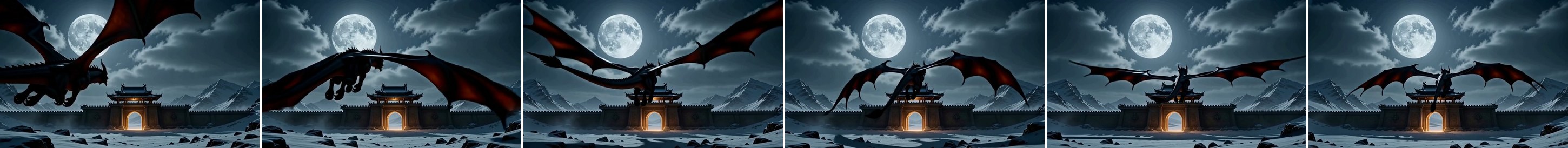}\\[-1pt]
  \parbox{\qgallerywidth}{\scriptsize\itshape A black dragon soars over a moonlit, snowy landscape and lands on the roof of a palace gate.}\\[0.25pt]
\includegraphics[width=\qgallerywidth]{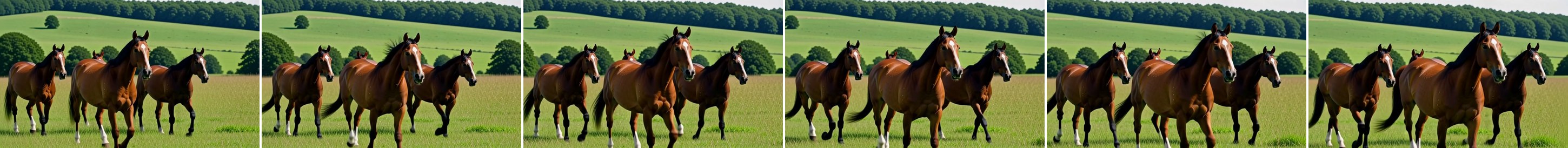}\\[-1pt]
  \parbox{\qgallerywidth}{\scriptsize\itshape A tracking shot follows a group of horses trotting across a lush green field.}\\[0.25pt]
\includegraphics[width=\qgallerywidth]{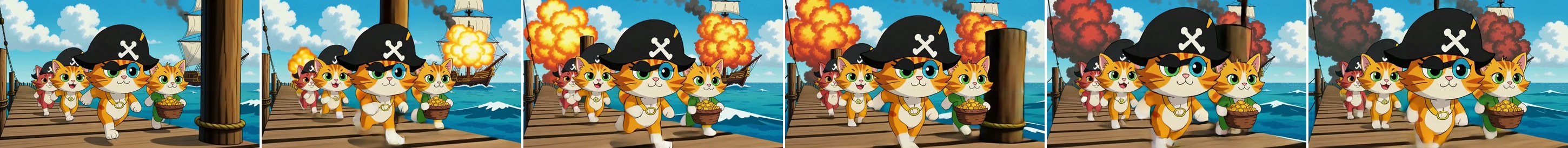}\\[-1pt]
  \parbox{\qgallerywidth}{\scriptsize\itshape 2D animation: pirate kittens run along a wooden pier as cannon fire explodes behind them.}\\[0.25pt]
\includegraphics[width=\qgallerywidth]{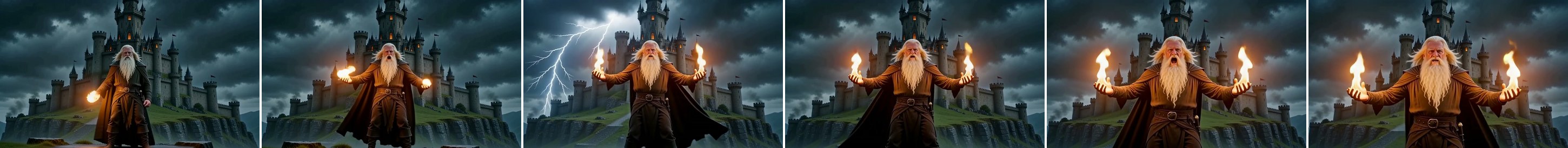}\\[-1pt]
  \parbox{\qgallerywidth}{\scriptsize\itshape A wizard raises flaming hands before a storm-lit castle as lightning strikes behind him.}\\[0.25pt]
\vspace{0pt}
\qsep{4-step student}
\includegraphics[width=\qgallerywidth]{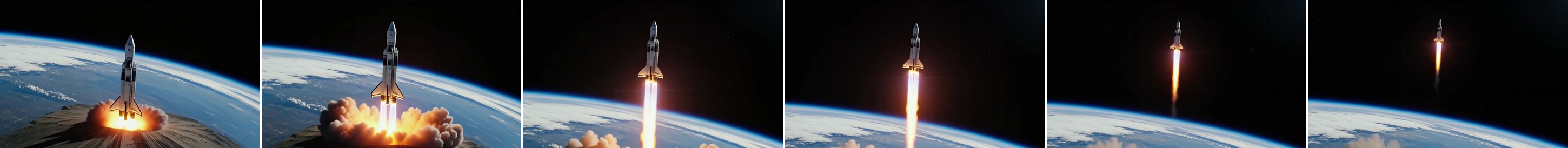}\\[-1pt]
  \parbox{\qgallerywidth}{\scriptsize\itshape A spacecraft lifts off, breaks through the atmosphere and climbs away from the curve of the Earth.}\\[0.25pt]
\includegraphics[width=\qgallerywidth]{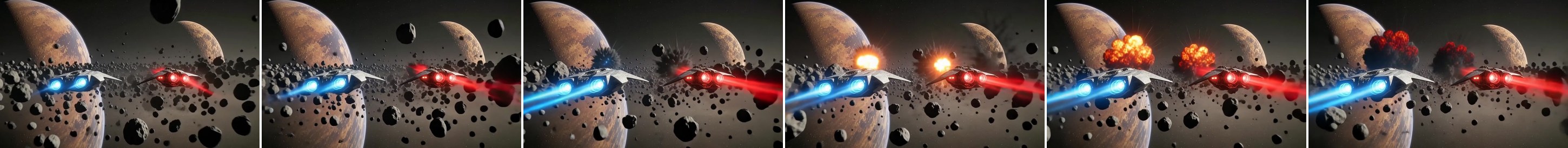}\\[-1pt]
  \parbox{\qgallerywidth}{\scriptsize\itshape 3D CG: two spacecraft chase each other through an asteroid field, firing lasers that blow up the rocks.}\\[0.25pt]
\includegraphics[width=\qgallerywidth]{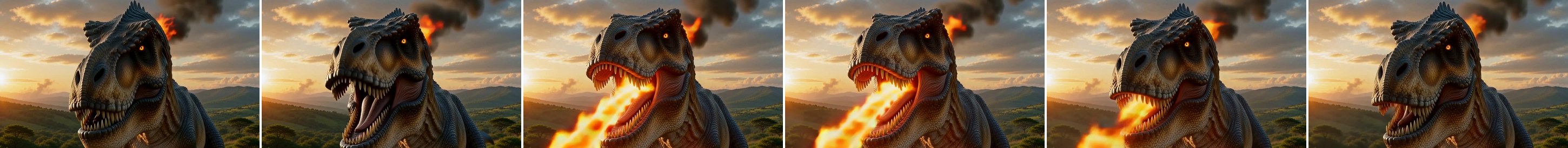}\\[-1pt]
  \parbox{\qgallerywidth}{\scriptsize\itshape A Tyrannosaurus rex roars against a sunset sky as flames erupt from its jaws.}\\[0.25pt]
\includegraphics[width=\qgallerywidth]{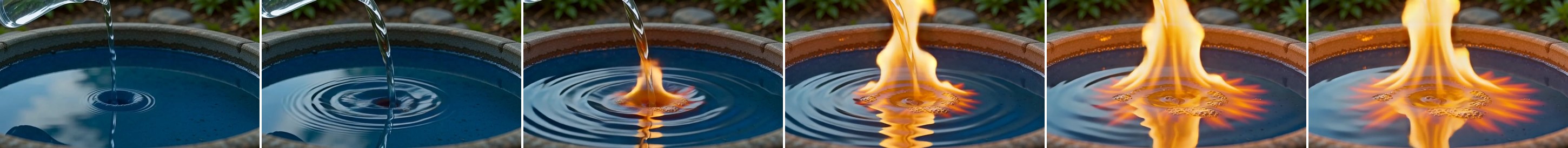}\\[-1pt]
  \parbox{\qgallerywidth}{\scriptsize\itshape Clear liquid is poured into a pool of blue water, and a burst of flame erupts from the ripples.}\\[0.25pt]
\includegraphics[width=\qgallerywidth]{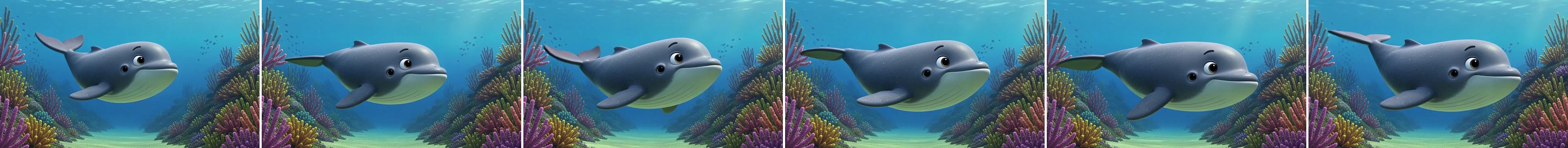}\\[-1pt]
  \parbox{\qgallerywidth}{\scriptsize\itshape 3D animation: a baby whale swims through a colourful coral reef, flapping its tail.}\\[0.25pt]
\caption{\textbf{Text-to-audio--video samples of the \qstudent{} student at 8 steps (top) and 4 steps (bottom).} 5\,s, $1344{\times}768$, with audio.}
\label{fig:supp:t2av}
\end{figure*}

\begin{figure*}[p]
\centering
\qsep{15\,s, 3 shots}
\includegraphics[width=\qgallerywidth]{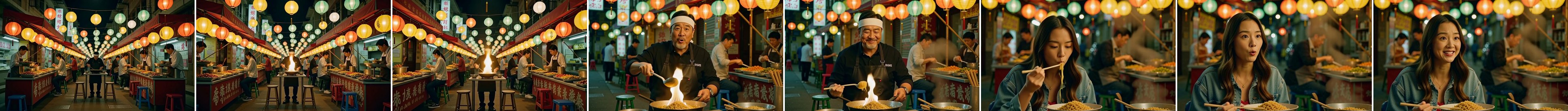}\\[-1pt]
  \parbox{\qgallerywidth}{\scriptsize\itshape Dialogue: a night market street, a cook tossing noodles in a flaming wok who calls out an order, and a customer tasting them.}\\[0.25pt]
\includegraphics[width=\qgallerywidth]{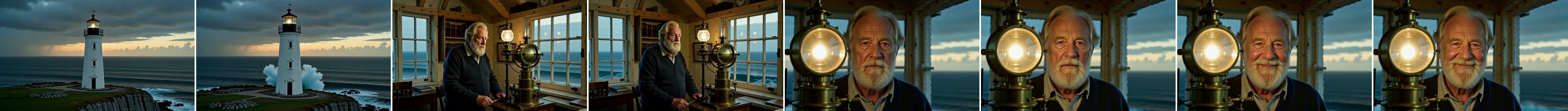}\\[-1pt]
  \parbox{\qgallerywidth}{\scriptsize\itshape Dialogue: waves crash below a lighthouse at dusk, the old keeper turns a brass crank in the lamp room, and the lamp flares and sweeps the sea.}\\[0.25pt]
\includegraphics[width=\qgallerywidth]{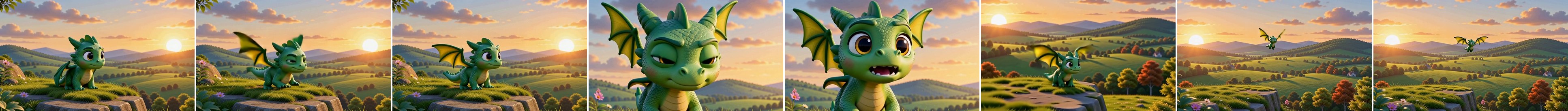}\\[-1pt]
  \parbox{\qgallerywidth}{\scriptsize\itshape Dialogue, 3D animation: a baby dragon hops on a cliff at sunrise, says ``I can do this'', leaps off and soars over the valley.}\\[0.25pt]
\vspace{2pt}
\qsep{10\,s, 3 shots}
\includegraphics[width=\qgallerywidth]{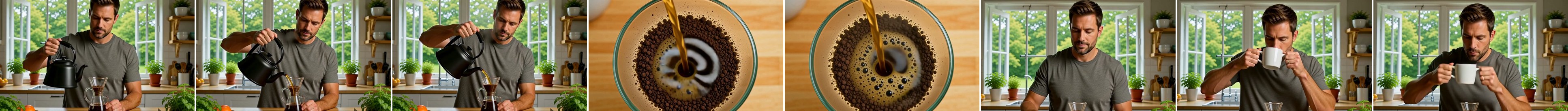}\\[-1pt]
  \parbox{\qgallerywidth}{\scriptsize\itshape A man pours hot water over coffee in a sunlit kitchen, the grounds bloom in close-up, and he takes the first sip.}\\[0.25pt]
\vspace{2pt}
\qsep{10\,s, 2 shots}
\includegraphics[width=\qgallerywidth]{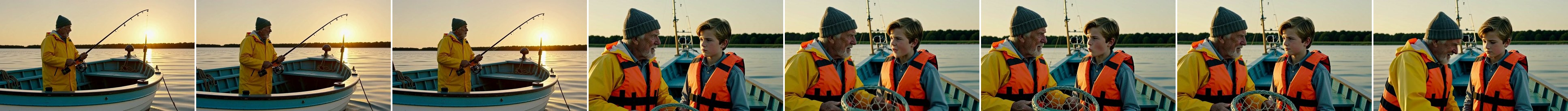}\\[-1pt]
  \parbox{\qgallerywidth}{\scriptsize\itshape Dialogue: an old fisherman fights a bending rod at sunrise, and his grandson grabs the landing net.}\\[0.25pt]
\includegraphics[width=\qgallerywidth]{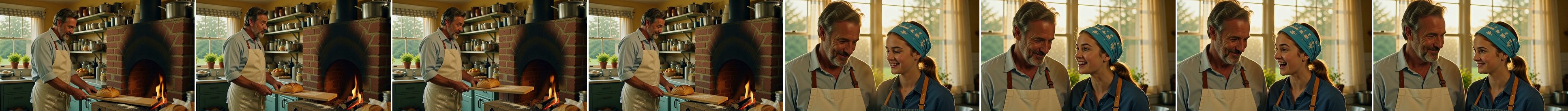}\\[-1pt]
  \parbox{\qgallerywidth}{\scriptsize\itshape Dialogue: a baker slides a loaf out of a brick oven at dawn, and his apprentice breathes in the steam and smiles.}\\[0.25pt]
\includegraphics[width=\qgallerywidth]{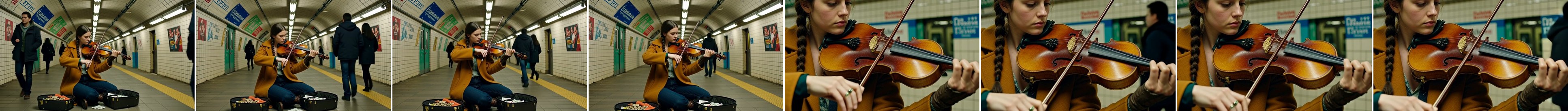}\\[-1pt]
  \parbox{\qgallerywidth}{\scriptsize\itshape Music: a street violinist plays a fast folk tune in a subway passage, then a close-up of bow and fingers.}\\[0.25pt]
\includegraphics[width=\qgallerywidth]{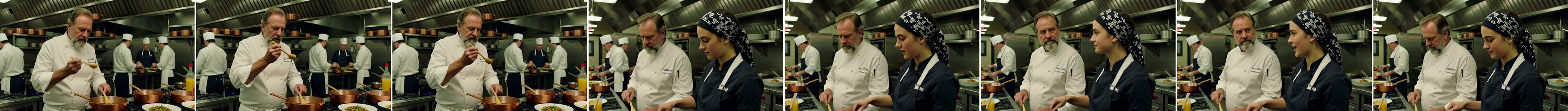}\\[-1pt]
  \parbox{\qgallerywidth}{\scriptsize\itshape Dialogue: a head chef tastes a sauce and asks for more salt, and a young sous chef answers as she stirs the pan.}\\[0.25pt]
\includegraphics[width=\qgallerywidth]{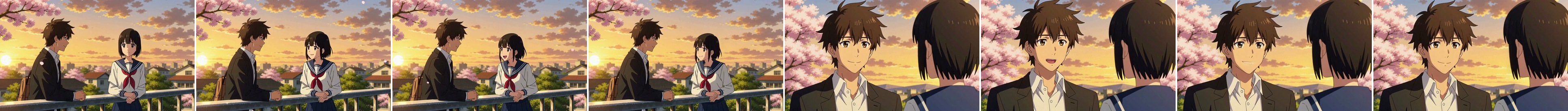}\\[-1pt]
  \parbox{\qgallerywidth}{\scriptsize\itshape Dialogue, 2D anime: two students promise to stay friends on a school rooftop at sunset among drifting cherry blossoms.}\\[0.25pt]
\includegraphics[width=\qgallerywidth]{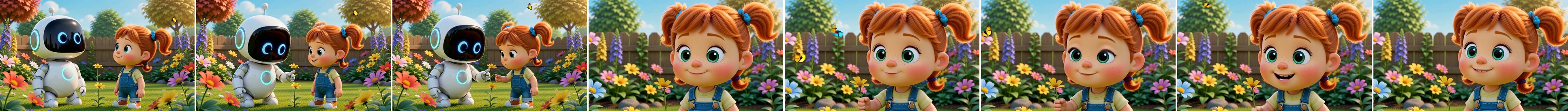}\\[-1pt]
  \parbox{\qgallerywidth}{\scriptsize\itshape Dialogue, 3D animation: a small robot asks a girl about a flower in a sunny garden, and she explains it is alive.}\\[0.25pt]
\caption{\textbf{Multi-shot 10 and 15\,s samples of the 8-step \qstudent{} student.} $1344{\times}768$ with audio, eight evenly spaced frames per clip. Shots, camera moves, speech and music all come from one prompt per clip, without any editing.}
\label{fig:supp:multishot}
\end{figure*}

\newcommand{\qlab}[1]{\makebox[0.075\textwidth][l]{\scriptsize #1}}

\begin{figure*}[p]
\centering
\qlab{Ours}\raisebox{-.5\height}{\includegraphics[width=0.925\textwidth]{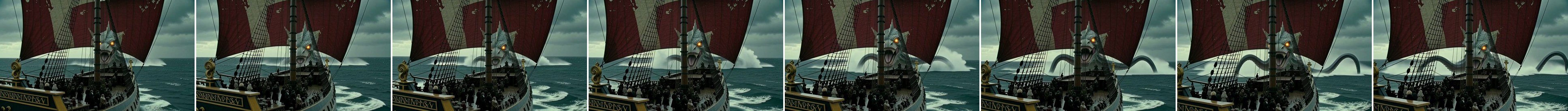}}\\[1pt]
  \qlab{H3}\raisebox{-.5\height}{\includegraphics[width=0.925\textwidth]{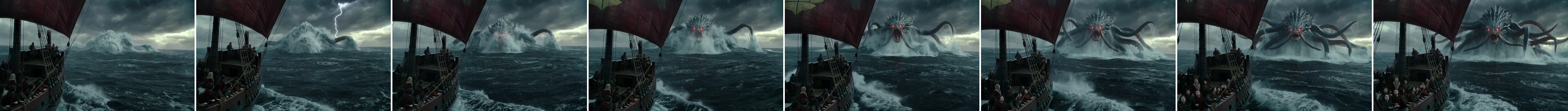}}\\[1pt]
  \qlab{LTX-2.5}\raisebox{-.5\height}{\includegraphics[width=0.925\textwidth]{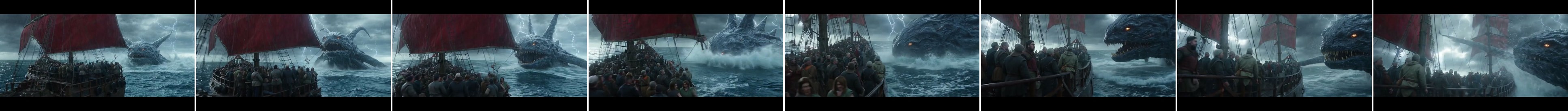}}\\[1pt]
  \hspace*{0.075\textwidth}\parbox[t]{0.925\textwidth}{\scriptsize\itshape A ship with a red sail carries people through a storm as a sea monster with glowing eyes and tentacles rises from the waves.}\\[4pt]
\qlab{Ours}\raisebox{-.5\height}{\includegraphics[width=0.925\textwidth]{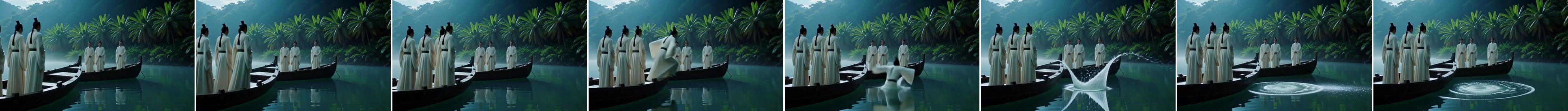}}\\[1pt]
  \qlab{H3}\raisebox{-.5\height}{\includegraphics[width=0.925\textwidth]{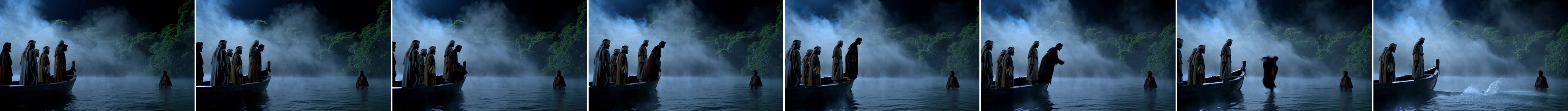}}\\[1pt]
  \qlab{LTX-2.5}\raisebox{-.5\height}{\includegraphics[width=0.925\textwidth]{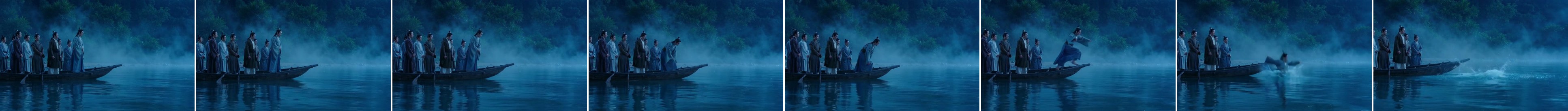}}\\[1pt]
  \hspace*{0.075\textwidth}\parbox[t]{0.925\textwidth}{\scriptsize\itshape On a misty river at night, a man in traditional robes steps off a wooden boat and leaps gracefully into the water.}\\[4pt]
\qlab{Ours}\raisebox{-.5\height}{\includegraphics[width=0.925\textwidth]{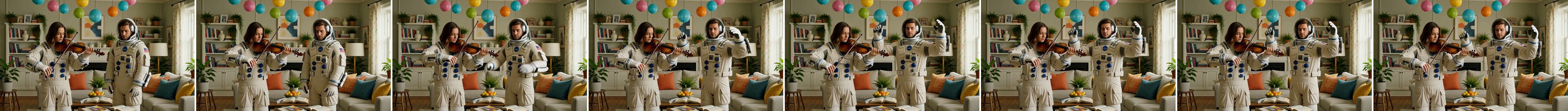}}\\[1pt]
  \qlab{H3}\raisebox{-.5\height}{\includegraphics[width=0.925\textwidth]{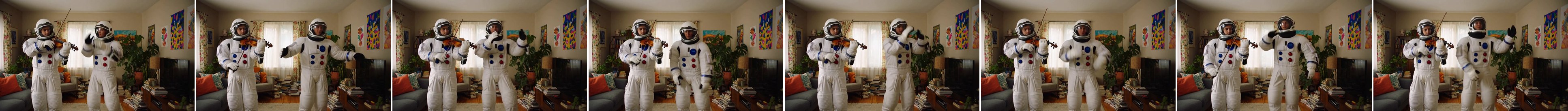}}\\[1pt]
  \qlab{LTX-2.5}\raisebox{-.5\height}{\includegraphics[width=0.925\textwidth]{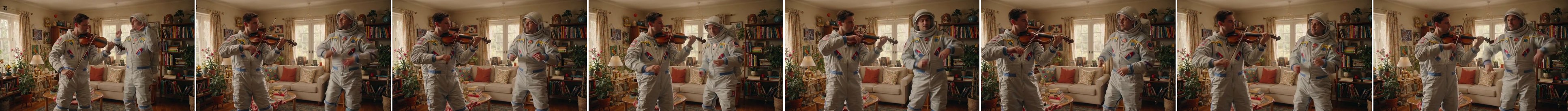}}\\[1pt]
  \hspace*{0.075\textwidth}\parbox[t]{0.925\textwidth}{\scriptsize\itshape Two people in astronaut suits in a cozy living room, one playing the violin and the other dancing energetically.}\\[4pt]
\qlab{Ours}\raisebox{-.5\height}{\includegraphics[width=0.925\textwidth]{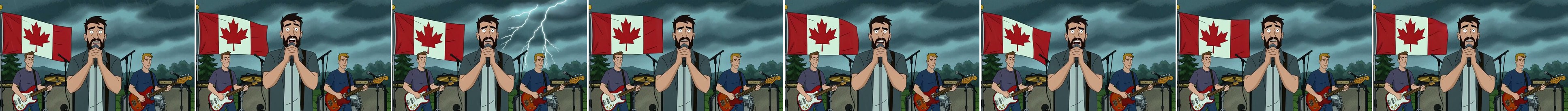}}\\[1pt]
  \qlab{H3}\raisebox{-.5\height}{\includegraphics[width=0.925\textwidth]{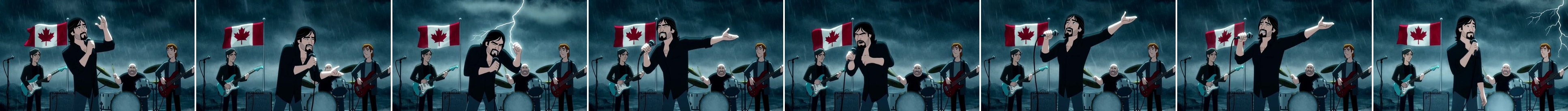}}\\[1pt]
  \qlab{LTX-2.5}\raisebox{-.5\height}{\includegraphics[width=0.925\textwidth]{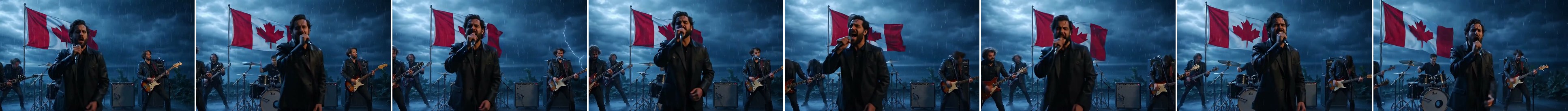}}\\[1pt]
  \hspace*{0.075\textwidth}\parbox[t]{0.925\textwidth}{\scriptsize\itshape 2D animation: a band performs against a stormy sky, the bearded singer belting into the microphone as a Canadian flag waves.}\\[4pt]
\qlab{Ours}\raisebox{-.5\height}{\includegraphics[width=0.925\textwidth]{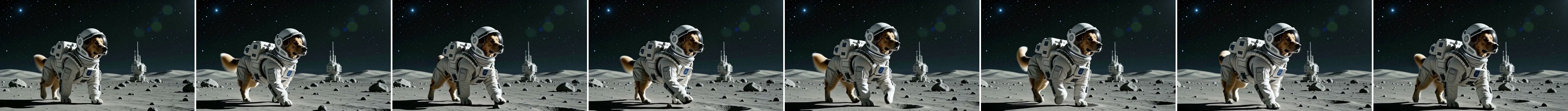}}\\[1pt]
  \qlab{H3}\raisebox{-.5\height}{\includegraphics[width=0.925\textwidth]{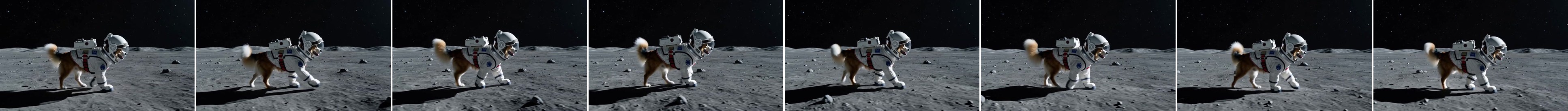}}\\[1pt]
  \qlab{LTX-2.5}\raisebox{-.5\height}{\includegraphics[width=0.925\textwidth]{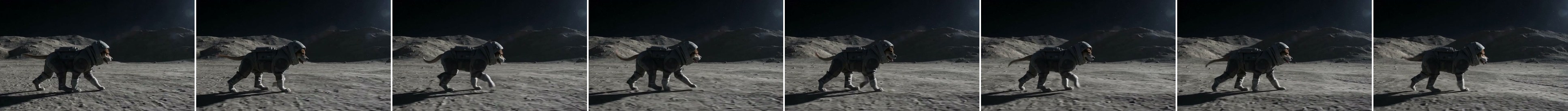}}\\[1pt]
  \hspace*{0.075\textwidth}\parbox[t]{0.925\textwidth}{\scriptsize\itshape A dog in a full astronaut suit and helmet walks across the lunar surface under a starry sky.}\\[4pt]
\caption{\textbf{5\,s text-to-audio--video comparison against MiniMax-H3 and LTX-2.5 distilled.} Ours and LTX-2.5: 8 steps. H3: 50 steps. Same prompt and seed, $1344{\times}768$ with audio, eight evenly spaced frames per clip. Prompts are shortened.}
\label{fig:supp:cmp5}
\end{figure*}

\begin{figure*}[p]
\centering
\qsep{15\,s}
\qlab{Ours}\raisebox{-.5\height}{\includegraphics[width=0.925\textwidth]{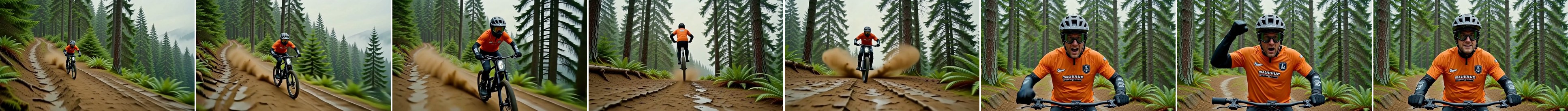}}\\[1pt]
  \qlab{H3}\raisebox{-.5\height}{\includegraphics[width=0.925\textwidth]{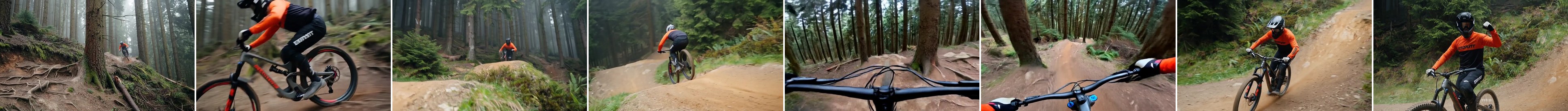}}\\[1pt]
  \qlab{LTX-2.5}\raisebox{-.5\height}{\includegraphics[width=0.925\textwidth]{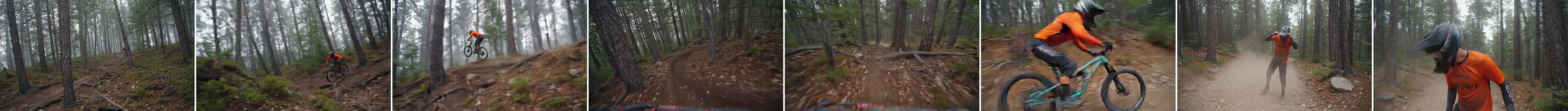}}\\[1pt]
  \hspace*{0.075\textwidth}\parbox[t]{0.925\textwidth}{\scriptsize\itshape 4 shots: a mountain biker descends a forest trail with a pan, a jump, a handlebar view and a skid stop.}\\[4pt]
\qlab{Ours}\raisebox{-.5\height}{\includegraphics[width=0.925\textwidth]{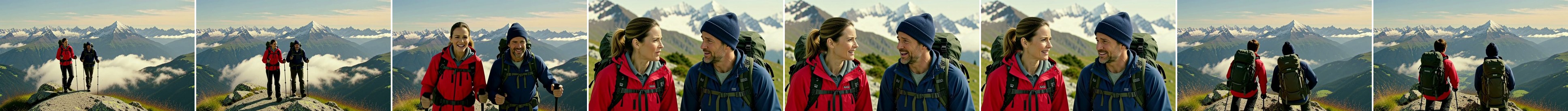}}\\[1pt]
  \qlab{H3}\raisebox{-.5\height}{\includegraphics[width=0.925\textwidth]{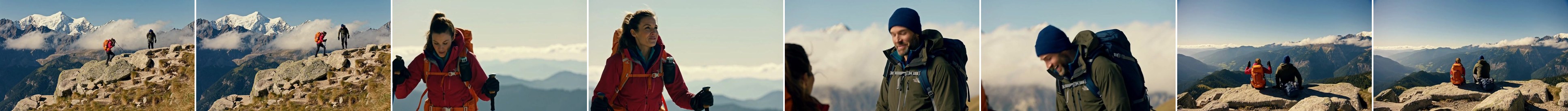}}\\[1pt]
  \qlab{LTX-2.5}\raisebox{-.5\height}{\includegraphics[width=0.925\textwidth]{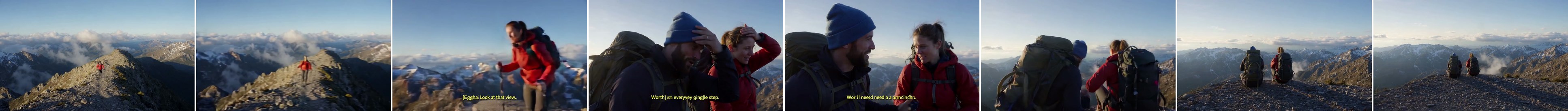}}\\[1pt]
  \hspace*{0.075\textwidth}\parbox[t]{0.925\textwidth}{\scriptsize\itshape 4 shots with dialogue: two hikers reach a mountain summit, exchange a few words and sit down to take in the view.}\\[4pt]
\qlab{Ours}\raisebox{-.5\height}{\includegraphics[width=0.925\textwidth]{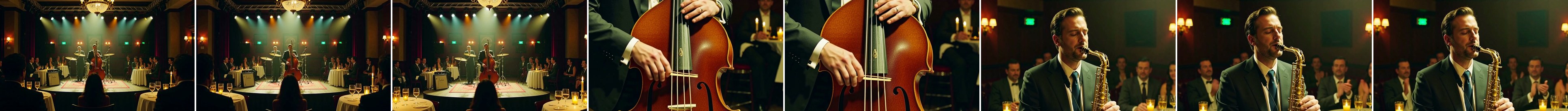}}\\[1pt]
  \qlab{H3}\raisebox{-.5\height}{\includegraphics[width=0.925\textwidth]{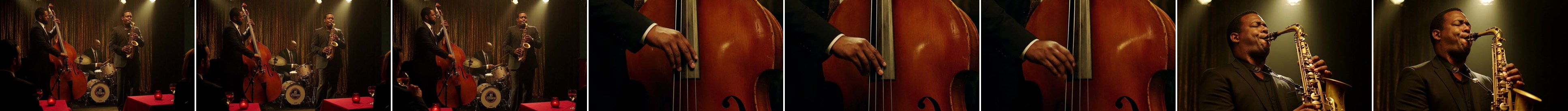}}\\[1pt]
  \qlab{LTX-2.5}\raisebox{-.5\height}{\includegraphics[width=0.925\textwidth]{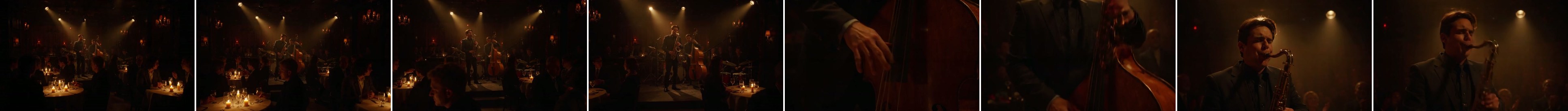}}\\[1pt]
  \hspace*{0.075\textwidth}\parbox[t]{0.925\textwidth}{\scriptsize\itshape 3 shots with music: a jazz trio in a dim club, an upright-bass close-up, and a saxophone solo to applause.}\\[4pt]
\qsep{10\,s}
\qlab{Ours}\raisebox{-.5\height}{\includegraphics[width=0.925\textwidth]{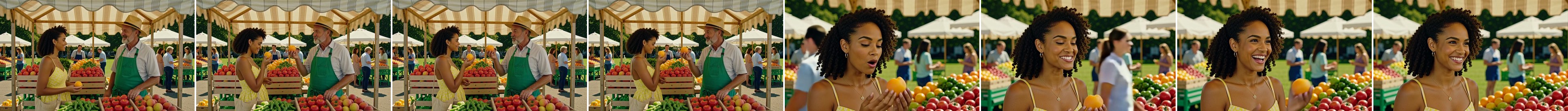}}\\[1pt]
  \qlab{H3}\raisebox{-.5\height}{\includegraphics[width=0.925\textwidth]{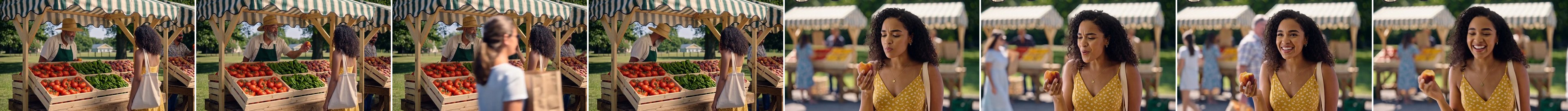}}\\[1pt]
  \qlab{LTX-2.5}\raisebox{-.5\height}{\includegraphics[width=0.925\textwidth]{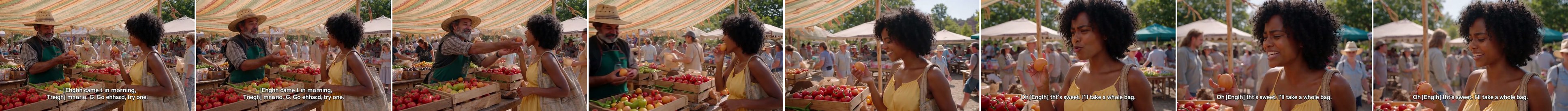}}\\[1pt]
  \hspace*{0.075\textwidth}\parbox[t]{0.925\textwidth}{\scriptsize\itshape 2 shots with dialogue: a market vendor offers a peach, and the customer tastes it and decides to buy a whole bag.}\\[4pt]
\qlab{Ours}\raisebox{-.5\height}{\includegraphics[width=0.925\textwidth]{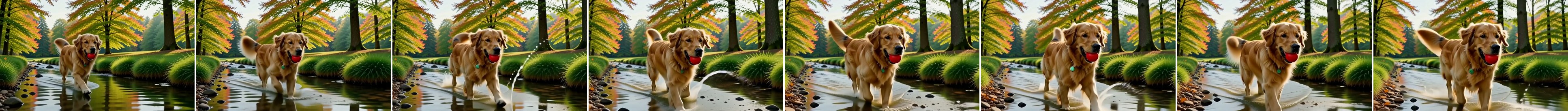}}\\[1pt]
  \qlab{H3}\raisebox{-.5\height}{\includegraphics[width=0.925\textwidth]{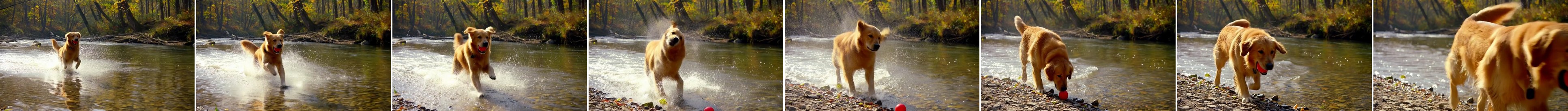}}\\[1pt]
  \qlab{LTX-2.5}\raisebox{-.5\height}{\includegraphics[width=0.925\textwidth]{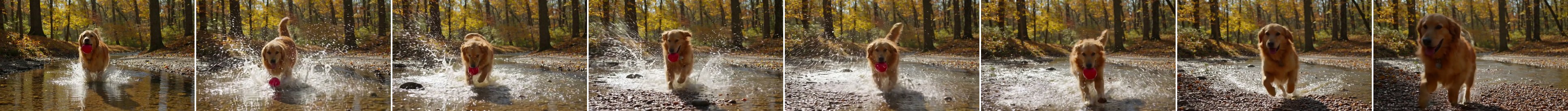}}\\[1pt]
  \hspace*{0.075\textwidth}\parbox[t]{0.925\textwidth}{\scriptsize\itshape A golden retriever splashes through a forest stream with a red ball in its mouth and trots toward the camera.}\\[4pt]
\caption{\textbf{Multi-shot 15\,s (top) and 10\,s (bottom) comparison against MiniMax-H3 and LTX-2.5 distilled.} Ours and LTX-2.5: 8 steps. H3: 50 steps. Same prompt, seed, $1344{\times}768$ with audio, eight frames per clip.}
\label{fig:supp:cmp_long}
\end{figure*}

\endgroup
\clearpage

\clearpage


\begin{thebibliography}{158}
\providecommand{\natexlab}[1]{#1}
\providecommand{\url}[1]{\texttt{#1}}
\expandafter\ifx\csname urlstyle\endcsname\relax
  \providecommand{\doi}[1]{doi: #1}\else
  \providecommand{\doi}{doi: \begingroup \urlstyle{rm}\Url}\fi

\bibitem[Bai et~al.(2025)Bai, Cai, Chen, Chen, Chen, Cheng, et~al.]{bai2025qwen3vl}
Shuai Bai, Yuxuan Cai, Ruizhe Chen, Keqin Chen, Xionghui Chen, Zesen Cheng, et~al.
\newblock {Qwen3-VL Technical Report}.
\newblock \emph{arXiv preprint arXiv:2511.21631}, 2025.

\bibitem[Bai et~al.(2026)Bai, Xu, Ren, Hao, Sun, Chen, and Sun]{bai2025instantvir}
Weimin Bai, Suzhe Xu, Yiwei Ren, Jinhua Hao, Ming Sun, Wenzheng Chen, and He~Sun.
\newblock {InstantViR: Real-Time Video Inverse Problem Solver with Distilled Diffusion Prior}.
\newblock In \emph{Proceedings of the IEEE/CVF Conference on Computer Vision and Pattern Recognition (CVPR)}, pages 16583--16592, 2026.

\bibitem[Bar-Tal et~al.(2024)Bar-Tal, Chefer, Tov, Herrmann, Paiss, Zada, Ephrat, Hur, Liu, Raj, Li, Rubinstein, Michaeli, Wang, Sun, Dekel, and Mosseri]{bartal2024lumiere}
Omer Bar-Tal, Hila Chefer, Omer Tov, Charles Herrmann, Roni Paiss, Shiran Zada, Ariel Ephrat, Junhwa Hur, Guanghui Liu, Amit Raj, Yuanzhen Li, Michael Rubinstein, Tomer Michaeli, Oliver Wang, Deqing Sun, Tali Dekel, and Inbar Mosseri.
\newblock {Lumiere: A Space-Time Diffusion Model for Video Generation}.
\newblock In \emph{SIGGRAPH Asia 2024 Conference Papers}, pages 1--11, 2024.
\newblock \doi{10.1145/3680528.3687614}.

\bibitem[Bengio(2020)]{bengio2020dtp}
Yoshua Bengio.
\newblock {Deriving Differential Target Propagation from Iterating Approximate Inverses}.
\newblock \emph{arXiv preprint arXiv:2007.15139}, 2020.

\bibitem[Blattmann et~al.(2023)Blattmann, Dockhorn, Kulal, Mendelevitch, Kilian, Lorenz, Levi, English, Voleti, Letts, Jampani, and Rombach]{blattmann2023svd}
Andreas Blattmann, Tim Dockhorn, Sumith Kulal, Daniel Mendelevitch, Maciej Kilian, Dominik Lorenz, Yam Levi, Zion English, Vikram Voleti, Adam Letts, Varun Jampani, and Robin Rombach.
\newblock {Stable Video Diffusion: Scaling Latent Video Diffusion Models to Large Datasets}.
\newblock \emph{arXiv preprint arXiv:2311.15127}, 2023.

\bibitem[Bohan(2025)]{boerbohan2025taehv}
Ollin~Boer Bohan.
\newblock {TAEHV: Tiny AutoEncoder for Hunyuan Video (and other video models)}.
\newblock \url{https://github.com/madebyollin/taehv}, 2025.

\bibitem[Castiglione et~al.(2026)Castiglione, Buckley, and Simpson]{castiglione2026cgtp}
Giuseppe Castiglione, Christopher Buckley, and Ivor Simpson.
\newblock {Trust Your Targets: Is Target Propagation Block-Diagonal Gauss-Newton?}
\newblock \emph{arXiv preprint arXiv:2610.26283}, 2026.

\bibitem[Chen et~al.(2025{\natexlab{a}})Chen, Lin, Yang, Lin, Zhu, Fan, et~al.]{chen2025skyreelsv2}
Guibin Chen, Dixuan Lin, Jiangping Yang, Chunze Lin, Junchen Zhu, Mingyuan Fan, et~al.
\newblock {SkyReels-V2: Infinite-length Film Generative Model}.
\newblock \emph{arXiv preprint arXiv:2504.13074}, 2025{\natexlab{a}}.

\bibitem[Chen et~al.(2024)Chen, Zhang, Cun, Xia, Wang, Weng, and Shan]{chen2024videocrafter2}
Haoxin Chen, Yong Zhang, Xiaodong Cun, Menghan Xia, Xintao Wang, Chao Weng, and Ying Shan.
\newblock {VideoCrafter2: Overcoming Data Limitations for High-Quality Video Diffusion Models}.
\newblock In \emph{Proceedings of the IEEE/CVF Conference on Computer Vision and Pattern Recognition (CVPR)}, pages 7310--7320, 2024.

\bibitem[Chen et~al.(2025{\natexlab{b}})Chen, Xue, Zhao, Yu, Paul, Chen, et~al.]{chen2025sanasprint}
Junsong Chen, Shuchen Xue, Yuyang Zhao, Jincheng Yu, Sayak Paul, Junyu Chen, et~al.
\newblock {SANA-Sprint: One-Step Diffusion with Continuous-Time Consistency Distillation}.
\newblock \emph{arXiv preprint arXiv:2503.09641}, 2025{\natexlab{b}}.

\bibitem[Chen et~al.(2026{\natexlab{a}})Chen, Yu, Li, Xue, Liu, Xin, Zhao, Ye, Wu, Wang, Zhou, Luo, Han, and Xie]{chen2026sanavideo2}
Junsong Chen, Jincheng Yu, Yitong Li, Shuchen Xue, Haozhe Liu, Jingyu Xin, Yuyang Zhao, Tian Ye, Zhangjie Wu, Zian Wang, Daquan Zhou, Ping Luo, Song Han, and Enze Xie.
\newblock {SANA-Video 2.0: Hybrid Linear Attention with Attention Residuals for Efficient Video Generation}.
\newblock \emph{arXiv preprint arXiv:2607.21553}, 2026{\natexlab{a}}.

\bibitem[Chen et~al.(2026{\natexlab{b}})Chen, Zhao, Yu, Chu, Chen, Yang, Wang, Pan, Zhou, Ling, Liu, Yi, Zhang, Li, Chen, Cai, Fidler, Luo, Han, and Xie]{chen2025sanavideo}
Junsong Chen, Yuyang Zhao, Jincheng Yu, Ruihang Chu, Junyu Chen, Shuai Yang, Xianbang Wang, Yicheng Pan, Daquan Zhou, Huan Ling, Haozhe Liu, Hongwei Yi, Hao Zhang, Muyang Li, Yukang Chen, Han Cai, Sanja Fidler, Ping Luo, Song Han, and Enze Xie.
\newblock {SANA-Video: Efficient Video Generation with Block Linear Diffusion Transformer}.
\newblock In \emph{International Conference on Learning Representations (ICLR)}, 2026{\natexlab{b}}.

\bibitem[Chen et~al.(2025{\natexlab{c}})Chen, Ge, Zhang, Zhang, Zhu, Yang, et~al.]{chen2025goku}
Shoufa Chen, Chongjian Ge, Yuqi Zhang, Yida Zhang, Fengda Zhu, Hao Yang, et~al.
\newblock {Goku: Flow Based Video Generative Foundation Models}.
\newblock In \emph{Proceedings of the IEEE/CVF Conference on Computer Vision and Pattern Recognition (CVPR)}, pages 23516--23527, 2025{\natexlab{c}}.

\bibitem[Chen et~al.(2026{\natexlab{c}})Chen, Liu, Jia, Wang, Ling, Qu, and Gao]{chen2026dfd}
Siyi Chen, Shaowei Liu, Yixuan Jia, Zian Wang, Huan Ling, Qing Qu, and Jun Gao.
\newblock {Data-Forcing Distillation: Restoring Diversity and Fidelity in Few-Step Video Generation}.
\newblock \emph{arXiv preprint arXiv:2606.18478}, 2026{\natexlab{c}}.

\bibitem[Chen et~al.(2026{\natexlab{d}})Chen, Wang, Huang, Yang, Zhang, Xiao, Chu, Mao, Hu, Liu, Zhao, Mao, Chen, Xie, Qi, and Han]{chen2026longlive2}
Yukang Chen, Luozhou Wang, Wei Huang, Shuai Yang, Bohan Zhang, Yicheng Xiao, Ruihang Chu, Weian Mao, Qixin Hu, Shaoteng Liu, Yuyang Zhao, Huizi Mao, Ying-Cong Chen, Enze Xie, Xiaojuan Qi, and Song Han.
\newblock {LongLive-2.0: An NVFP4 Parallel Infrastructure for Long Video Generation}.
\newblock \emph{arXiv preprint arXiv:2605.18739}, 2026{\natexlab{d}}.

\bibitem[Clark et~al.(2024)Clark, Vicol, Swersky, and Fleet]{clark2023draft}
Kevin Clark, Paul Vicol, Kevin Swersky, and David~J Fleet.
\newblock {Directly Fine-Tuning Diffusion Models on Differentiable Rewards}.
\newblock In \emph{International Conference on Learning Representations (ICLR)}, 2024.

\bibitem[Dai et~al.(2026)Dai, Bu, Zhou, Liu, Zhang, Yang, Liu, and Zang]{dai2026instantfusion}
Xuanlang Dai, Jiazi Bu, Yujie Zhou, Yuhong Liu, Beichen Zhang, Penghui Yang, Ziyu Liu, and Yuhang Zang.
\newblock {InstantFusion: A Shared Latent Interface for Heterogeneous Image Generators}.
\newblock \emph{arXiv preprint arXiv:2610.13944}, 2026.

\bibitem[Dumont et~al.(2026)Dumont, Lacombe, and Vialard]{dumont2026gndrifting}
Théo Dumont, Théo Lacombe, and François-Xavier Vialard.
\newblock {Gauss-Newton drifting extends natural gradient descent}.
\newblock \emph{arXiv preprint arXiv:2609.17167}, 2026.

\bibitem[Efron(2011)]{efron2011tweedie}
Bradley Efron.
\newblock {Tweedie's Formula and Selection Bias}.
\newblock \emph{Journal of the American Statistical Association}, 106\penalty0 (496):\penalty0 1602--1614, 2011.
\newblock \doi{10.1198/jasa.2011.tm11181}.

\bibitem[Esser et~al.(2024)Esser, Kulal, Blattmann, Entezari, M{\"u}ller, Saini, Levi, Lorenz, Sauer, Boesel, Podell, Dockhorn, English, and Rombach]{esser2024sd3}
Patrick Esser, Sumith Kulal, Andreas Blattmann, Rahim Entezari, Jonas M{\"u}ller, Harry Saini, Yam Levi, Dominik Lorenz, Axel Sauer, Frederic Boesel, Dustin Podell, Tim Dockhorn, Zion English, and Robin Rombach.
\newblock {Scaling Rectified Flow Transformers for High-Resolution Image Synthesis}.
\newblock In \emph{Proceedings of the 41st International Conference on Machine Learning (ICML)}, volume 235 of \emph{Proceedings of Machine Learning Research}, pages 12606--12633, 2024.

\bibitem[Fan et~al.(2026)Fan, Qiu, Wu, Wang, Lin, Ren, Lin, Gong, and Yang]{fan2025phaseddmd}
Xiangyu Fan, Zesong Qiu, Zhuguanyu Wu, Fanzhou Wang, Zhiqian Lin, Tianxiang Ren, Dahua Lin, Ruihao Gong, and Lei Yang.
\newblock {Phased DMD: Few-step Distribution Matching Distillation via Score Matching within Subintervals}.
\newblock In \emph{Proceedings of the IEEE/CVF Conference on Computer Vision and Pattern Recognition (CVPR)}, pages 41667--41676, 2026.

\bibitem[{FastVideo Team}(2026)]{fastvideo2026fasth3}
{FastVideo Team}.
\newblock {FastVideo-FastH3-4-step-Preview-v1-LoRA}.
\newblock \url{https://huggingface.co/FastVideo/FastVideo-FastH3-4-step-Preview-v1-LoRA}, 2026.
\newblock Model card.

\bibitem[Fei et~al.(2025)Fei, Li, Qiu, Wang, Dou, Wang, Xu, Fan, Chen, Li, and Zhou]{fei2025skyreelsa2}
Zhengcong Fei, Debang Li, Di~Qiu, Jiahua Wang, Yikun Dou, Rui Wang, Jingtao Xu, Mingyuan Fan, Guibin Chen, Yang Li, and Yahui Zhou.
\newblock {SkyReels-A2: Compose Anything in Video Diffusion Transformers}.
\newblock \emph{arXiv preprint arXiv:2504.02436}, 2025.

\bibitem[Feng et~al.(2026)Feng, Karkus, Igl, Berner, Chen, Tan, Alahi, Ivanovic, and Pavone]{feng2026ptd}
Lan Feng, Peter Karkus, Maximilian Igl, Julius Berner, Yuxiao Chen, Shuhan Tan, Alexandre Alahi, Boris Ivanovic, and Marco Pavone.
\newblock {Parametric Trajectory Distillation for Few-Step Video Generation}.
\newblock \emph{arXiv preprint arXiv:2610.11498}, 2026.

\bibitem[Fu et~al.(2026{\natexlab{a}})Fu, Fu, He, Wang, Huang, Ma, Zhong, Huang, He, and Xu]{fu2026anyopd}
Siming Fu, Zheming Fu, Ruizhe He, Hualiang Wang, Jie Huang, Xiaoxiao Ma, Mingchen Zhong, Weihu Huang, Xiaoxuan He, and Haojun Xu.
\newblock {Any-OPD: Heterogeneous On-Policy Distillation for Flow-Matching Models via Representation-Space Bridging}.
\newblock \emph{arXiv preprint arXiv:2608.03316}, 2026{\natexlab{a}}.

\bibitem[Fu et~al.(2026{\natexlab{b}})Fu, Xu, He, Fu, Wang, Huang, Ma, Zhong, Huang, He, Huang, and Liu]{fu2026polyopd}
Siming Fu, Haojun Xu, Ruizhe He, Zheming Fu, Hualiang Wang, Jie Huang, Xiaoxiao Ma, Mingchen Zhong, Weihu Huang, Xiaoxuan He, Linjiang Huang, and Si~Liu.
\newblock {Poly-OPD: Heterogeneous Multi-Teacher On-Policy Distillation for Capability-Selectable Flow Models}.
\newblock \emph{arXiv preprint arXiv:2608.04349}, 2026{\natexlab{b}}.

\bibitem[Fu et~al.(2026{\natexlab{c}})Fu, He, Ning, Ma, Yang, Mu, Huang, Liu, Wang, and Fu]{fu2026graftood}
Zheming Fu, Ruizhe He, Wenbin Ning, Xiaoxiao Ma, Xuan Yang, Lianrui Mu, Haoyang Huang, Chang Liu, Lei Wang, and Siming Fu.
\newblock {GraftOOD: Grafting Out-of-Distribution Exemplars into Advantage-Weighted Rectification for Video Reward Optimization}.
\newblock \emph{arXiv preprint arXiv:2610.13470}, 2026{\natexlab{c}}.

\bibitem[Gao et~al.(2025)Gao, Guo, Hoang, Huang, Jiang, Kong, et~al.]{gao2025seedance}
Yu~Gao, Haoyuan Guo, Tuyen Hoang, Weilin Huang, Lu~Jiang, Fangyuan Kong, et~al.
\newblock {Seedance 1.0: Exploring the Boundaries of Video Generation Models}.
\newblock \emph{arXiv preprint arXiv:2506.09113}, 2025.

\bibitem[{Genmo Team}(2024)]{genmo2024mochi}
{Genmo Team}.
\newblock {Mochi 1}.
\newblock \url{https://github.com/genmoai/mochi}, 2024.

\bibitem[Girdhar et~al.(2023)Girdhar, El-Nouby, Liu, Singh, Alwala, Joulin, and Misra]{girdhar2023imagebind}
Rohit Girdhar, Alaaeldin El-Nouby, Zhuang Liu, Mannat Singh, Kalyan~Vasudev Alwala, Armand Joulin, and Ishan Misra.
\newblock {ImageBind: One Embedding Space To Bind Them All}.
\newblock In \emph{Proceedings of the IEEE/CVF Conference on Computer Vision and Pattern Recognition (CVPR)}, pages 15180--15190, 2023.

\bibitem[Goodfellow et~al.(2014)Goodfellow, Pouget-Abadie, Mirza, Xu, Warde-Farley, Ozair, Courville, and Bengio]{goodfellow2014gan}
Ian~J. Goodfellow, Jean Pouget-Abadie, Mehdi Mirza, Bing Xu, David Warde-Farley, Sherjil Ozair, Aaron Courville, and Yoshua Bengio.
\newblock {Generative Adversarial Nets}.
\newblock In \emph{Advances in Neural Information Processing Systems (NeurIPS)}, volume~27, 2014.

\bibitem[Graikos et~al.(2025)Graikos, Jojic, and Samaras]{graikos2024newton}
Alexandros Graikos, Nebojsa Jojic, and Dimitris Samaras.
\newblock {Fast constrained sampling in pre-trained diffusion models}.
\newblock In \emph{Advances in Neural Information Processing Systems (NeurIPS)}, 2025.

\bibitem[Gu et~al.(2026)Gu, Fang, Jiang, Mao, Han, Cai, and Shou]{gu2026anyflow}
Yuchao Gu, Guian Fang, Yuxin Jiang, Weijia Mao, Song Han, Han Cai, and Mike~Zheng Shou.
\newblock {AnyFlow: Any-Step Video Diffusion Model with On-Policy Flow Map Distillation}.
\newblock \emph{arXiv preprint arXiv:2605.13724}, 2026.

\bibitem[HaCohen et~al.(2024)HaCohen, Chiprut, Brazowski, Shalem, Moshe, Richardson, Levin, Shiran, Zabari, Gordon, Panet, Weissbuch, Kulikov, Bitterman, Melumian, and Bibi]{hacohen2024ltxvideo}
Yoav HaCohen, Nisan Chiprut, Benny Brazowski, Daniel Shalem, Dudu Moshe, Eitan Richardson, Eran Levin, Guy Shiran, Nir Zabari, Ori Gordon, Poriya Panet, Sapir Weissbuch, Victor Kulikov, Yaki Bitterman, Zeev Melumian, and Ofir Bibi.
\newblock {LTX-Video: Realtime Video Latent Diffusion}.
\newblock \emph{arXiv preprint arXiv:2501.00103}, 2024.

\bibitem[HaCohen et~al.(2026)HaCohen, Brazowski, Chiprut, Bitterman, Kvochko, Berkowitz, et~al.]{hacohen2026ltx2}
Yoav HaCohen, Benny Brazowski, Nisan Chiprut, Yaki Bitterman, Andrew Kvochko, Avishai Berkowitz, et~al.
\newblock {LTX-2: Efficient Joint Audio-Visual Foundation Model}.
\newblock \emph{arXiv preprint arXiv:2601.03233}, 2026.

\bibitem[Hestenes and Stiefel(1952)]{hestenes1952cg}
Magnus~R. Hestenes and Eduard Stiefel.
\newblock {Methods of Conjugate Gradients for Solving Linear Systems}.
\newblock \emph{Journal of Research of the National Bureau of Standards}, 49\penalty0 (6):\penalty0 409--436, 1952.

\bibitem[Ho and Salimans(2022)]{ho2022cfg}
Jonathan Ho and Tim Salimans.
\newblock {Classifier-Free Diffusion Guidance}.
\newblock \emph{arXiv preprint arXiv:2207.12598}, 2022.

\bibitem[Ho et~al.(2022)Ho, Chan, Saharia, Whang, Gao, Gritsenko, Kingma, Poole, Norouzi, Fleet, and Salimans]{ho2022imagenvideo}
Jonathan Ho, William Chan, Chitwan Saharia, Jay Whang, Ruiqi Gao, Alexey Gritsenko, Diederik~P. Kingma, Ben Poole, Mohammad Norouzi, David~J. Fleet, and Tim Salimans.
\newblock {Imagen Video: High Definition Video Generation with Diffusion Models}.
\newblock \emph{arXiv preprint arXiv:2210.02303}, 2022.

\bibitem[Hu et~al.(2022)Hu, Shen, Wallis, Allen-Zhu, Li, Wang, Wang, and Chen]{hu2022lora}
Edward~J. Hu, Yelong Shen, Phillip Wallis, Zeyuan Allen-Zhu, Yuanzhi Li, Shean Wang, Lu~Wang, and Weizhu Chen.
\newblock {LoRA: Low-Rank Adaptation of Large Language Models}.
\newblock In \emph{International Conference on Learning Representations (ICLR)}, 2022.

\bibitem[Hu et~al.(2025)Hu, Yu, Zhou, Liang, Zhou, Lin, and Lu]{hu2025hunyuancustom}
Teng Hu, Zhentao Yu, Zhengguang Zhou, Sen Liang, Yuan Zhou, Qin Lin, and Qinglin Lu.
\newblock {HunyuanCustom: A Multimodal-Driven Architecture for Customized Video Generation}.
\newblock \emph{arXiv preprint arXiv:2505.04512}, 2025.

\bibitem[Huang et~al.(2025)Huang, Li, He, Zhou, and Shechtman]{huang2025selfforcing}
Xun Huang, Zhengqi Li, Guande He, Mingyuan Zhou, and Eli Shechtman.
\newblock {Self Forcing: Bridging the Train-Test Gap in Autoregressive Video Diffusion}.
\newblock In \emph{Advances in Neural Information Processing Systems (NeurIPS)}, 2025.

\bibitem[Huang et~al.(2026)Huang, Zhao, Yao, Zhang, Huang, Cui, Wu, and Chen]{huang2026dynaforcing}
Yubo Huang, Sirui Zhao, Xinchen Yao, Zhengye Zhang, Jinyang Huang, Fengqi Cui, Shiwei Wu, and Enhong Chen.
\newblock {DynaForcing: Overcoming Dynamic Collapse in Self-Forcing Distillation for Streaming Avatar Generation}.
\newblock In \emph{Proceedings of the 34th ACM International Conference on Multimedia (MM)}, pages 15379--15387, 2026.
\newblock \doi{10.1145/3767308.3835407}.

\bibitem[Huang et~al.(2024)Huang, He, Yu, Zhang, Si, Jiang, Zhang, Wu, Jin, Chanpaisit, Wang, Chen, Wang, Lin, Qiao, and Liu]{huang2023vbench}
Ziqi Huang, Yinan He, Jiashuo Yu, Fan Zhang, Chenyang Si, Yuming Jiang, Yuanhan Zhang, Tianxing Wu, Qingyang Jin, Nattapol Chanpaisit, Yaohui Wang, Xinyuan Chen, Limin Wang, Dahua Lin, Yu~Qiao, and Ziwei Liu.
\newblock {VBench: Comprehensive Benchmark Suite for Video Generative Models}.
\newblock In \emph{Proceedings of the IEEE/CVF Conference on Computer Vision and Pattern Recognition (CVPR)}, pages 21807--21818, 2024.

\bibitem[Iashin et~al.(2024)Iashin, Xie, Rahtu, and Zisserman]{iashin2024synchformer}
Vladimir Iashin, Weidi Xie, Esa Rahtu, and Andrew Zisserman.
\newblock {Synchformer: Efficient Synchronization from Sparse Cues}.
\newblock In \emph{IEEE International Conference on Acoustics, Speech and Signal Processing (ICASSP)}, pages 5325--5329, 2024.
\newblock \doi{10.1109/ICASSP48485.2024.10448489}.

\bibitem[Jiang et~al.(2026{\natexlab{a}})Jiang, Luo, Yu, Han, Su, and Zou]{jiang2026mild}
Bingqing Jiang, Li~Luo, Zichao Yu, Yujin Han, Zhaolong Su, and Difan Zou.
\newblock {From Static to Dynamic: On-Policy Distillation from Image to Video Diffusion Models}.
\newblock \emph{arXiv preprint arXiv:2609.34371}, 2026{\natexlab{a}}.

\bibitem[Jiang et~al.(2026{\natexlab{b}})Jiang, Liu, Wang, Wu, Li, Li, Jin, Lu, Li, Wang, Hoi, Gao, and Yang]{jiang2025dmdr}
Dengyang Jiang, Dongyang Liu, Zanyi Wang, Qilong Wu, Liuzhuozheng Li, Hengzhuang Li, Xin Jin, Changsheng Lu, Zhen Li, Mengmeng Wang, Steven Hoi, Peng Gao, and Harry Yang.
\newblock {Distribution Matching Distillation Meets Reinforcement Learning}.
\newblock In \emph{Computer Vision -- {ECCV} 2026}, Lecture Notes in Computer Science, pages 281--299, 2026{\natexlab{b}}.
\newblock \doi{10.1007/978-3-032-37550-6_16}.

\bibitem[Jiang et~al.(2025)Jiang, Han, Mao, Zhang, Pan, and Liu]{jiang2025vace}
Zeyinzi Jiang, Zhen Han, Chaojie Mao, Jingfeng Zhang, Yulin Pan, and Yu~Liu.
\newblock {VACE: All-in-One Video Creation and Editing}.
\newblock In \emph{Proceedings of the IEEE/CVF International Conference on Computer Vision (ICCV)}, pages 17191--17202, 2025.

\bibitem[Karras et~al.(2022)Karras, Aittala, Aila, and Laine]{karras2022edm}
Tero Karras, Miika Aittala, Timo Aila, and Samuli Laine.
\newblock {Elucidating the Design Space of Diffusion-Based Generative Models}.
\newblock In \emph{Advances in Neural Information Processing Systems (NeurIPS)}, volume~35, pages 26565--26577, 2022.

\bibitem[{Kimi Team} et~al.(2025){Kimi Team}, Zhang, Lin, Yao, Hu, Meng, et~al.]{yang2025kimi}
{Kimi Team}, Yu~Zhang, Zongyu Lin, Xingcheng Yao, Jiaxi Hu, Fanqing Meng, et~al.
\newblock {Kimi Linear: An Expressive, Efficient Attention Architecture}.
\newblock \emph{arXiv preprint arXiv:2510.26692}, 2025.

\bibitem[{Kimi Team} et~al.(2026){Kimi Team}, Chen, Zhang, Su, Xu, Pan, et~al.]{kimiteam2026attnres}
{Kimi Team}, Guangyu Chen, Yu~Zhang, Jianlin Su, Weixin Xu, Siyuan Pan, et~al.
\newblock {Attention Residuals}.
\newblock \emph{arXiv preprint arXiv:2603.15031}, 2026.

\bibitem[Kong et~al.(2024)Kong, Tian, Zhang, Min, Dai, Zhou, et~al.]{kong2024hunyuanvideo}
Weijie Kong, Qi~Tian, Zijian Zhang, Rox Min, Zuozhuo Dai, Jin Zhou, et~al.
\newblock {HunyuanVideo: A Systematic Framework For Large Video Generative Models}.
\newblock \emph{arXiv preprint arXiv:2412.03603}, 2024.

\bibitem[Koutini et~al.(2022)Koutini, Schl{\"u}ter, Eghbal-zadeh, and Widmer]{koutini2022passt}
Khaled Koutini, Jan Schl{\"u}ter, Hamid Eghbal-zadeh, and Gerhard Widmer.
\newblock {Efficient Training of Audio Transformers with Patchout}.
\newblock In \emph{Interspeech}, pages 2753--2757, 2022.

\bibitem[Kynk{\"a}{\"a}nniemi et~al.(2024)Kynk{\"a}{\"a}nniemi, Aittala, Karras, Laine, Aila, and Lehtinen]{kynkaanniemi2024interval}
Tuomas Kynk{\"a}{\"a}nniemi, Miika Aittala, Tero Karras, Samuli Laine, Timo Aila, and Jaakko Lehtinen.
\newblock {Applying Guidance in a Limited Interval Improves Sample and Distribution Quality in Diffusion Models}.
\newblock In \emph{Advances in Neural Information Processing Systems (NeurIPS)}, volume~37, pages 122458--122483, 2024.

\bibitem[Li et~al.(2026{\natexlab{a}})Li, Li, Chen, Ye, Liu, Yu, Wang, Zhang, Xie, Xie, and Han]{li2026solattn}
Haopeng Li, Yitong Li, Junsong Chen, Tian Ye, Haozhe Liu, Jincheng Yu, Duomin Wang, Ruihua Zhang, Zeke Xie, Enze Xie, and Song Han.
\newblock {Sol-Attn: Accelerating Video Generation Inference via On-the-Fly Attention Sparsification}.
\newblock \emph{arXiv preprint arXiv:2607.24027}, 2026{\natexlab{a}}.

\bibitem[Li et~al.(2026{\natexlab{b}})Li, Chen, Li, Liu, Yu, Zhu, Luo, Han, and Xie]{solengine}
Yitong Li, Junsong Chen, Haopeng Li, Haozhe Liu, Jincheng Yu, Ligeng Zhu, Ping Luo, Song Han, and Enze Xie.
\newblock {Sol Video Inference Engine: Agent-Native Full-Stack Acceleration Framework for Efficient Video Generation}.
\newblock \emph{arXiv preprint arXiv:2606.23743}, 2026{\natexlab{b}}.

\bibitem[Li et~al.(2026{\natexlab{c}})Li, Yu, Chen, Li, Xue, Liu, Luo, Han, and Xie]{li2026solh3}
Yitong Li, Jincheng Yu, Junsong Chen, Haopeng Li, Shuchen Xue, Haozhe Liu, Ping Luo, Song Han, and Enze Xie.
\newblock {Sol-H3: Recursive Self-Improvement for MiniMax-H3 Inference Acceleration on Sol-Engine across Cloud and Edge}.
\newblock \emph{arXiv preprint arXiv:2609.35110}, 2026{\natexlab{c}}.

\bibitem[Liang et~al.(2025)Liang, Yu, Luo, Iyer, Dong, Zhou, Ghosh, Lewis, tau Yih, Zettlemoyer, and Lin]{liang2024mot}
Weixin Liang, Lili Yu, Liang Luo, Srinivasan Iyer, Ning Dong, Chunting Zhou, Gargi Ghosh, Mike Lewis, Wen tau Yih, Luke Zettlemoyer, and Xi~Victoria Lin.
\newblock {Mixture-of-Transformers: A Sparse and Scalable Architecture for Multi-Modal Foundation Models}.
\newblock \emph{Transactions on Machine Learning Research}, 2025.

\bibitem[Liao et~al.(2024)Liao, Lu, Zhang, Wan, Wang, Zhao, Zuo, Ye, and Wang]{liao2024devil}
Mingxiang Liao, Hannan Lu, Xinyu Zhang, Fang Wan, Tianyu Wang, Yuzhong Zhao, Wangmeng Zuo, Qixiang Ye, and Jingdong Wang.
\newblock {Evaluation of Text-to-Video Generation Models: A Dynamics Perspective}.
\newblock In \emph{Advances in Neural Information Processing Systems (NeurIPS)}, volume~37, pages 109790--109816, 2024.

\bibitem[{LightX2V Team}(2025{\natexlab{a}})]{lightx2v2025hy15distill}
{LightX2V Team}.
\newblock {Hy1.5-Distill-Models}.
\newblock \url{https://huggingface.co/lightx2v/Hy1.5-Distill-Models}, 2025{\natexlab{a}}.
\newblock Model card.

\bibitem[{LightX2V Team}(2025{\natexlab{b}})]{lightx2v2025wan22lightning}
{LightX2V Team}.
\newblock {Wan2.2-Lightning}.
\newblock \url{https://huggingface.co/lightx2v/Wan2.2-Lightning}, 2025{\natexlab{b}}.
\newblock Model card.

\bibitem[{LightX2V Team}(2026)]{lightx2v2026h3turbo}
{LightX2V Team}.
\newblock {MiniMax-H3 Turbo}.
\newblock \url{https://huggingface.co/lightx2v/Minimax-h3-Turbo}, 2026.
\newblock Model card.

\bibitem[Lin et~al.(2024{\natexlab{a}})Lin, Ge, Cheng, Li, Zhu, Wang, et~al.]{lin2024opensoraplan}
Bin Lin, Yunyang Ge, Xinhua Cheng, Zongjian Li, Bin Zhu, Shaodong Wang, et~al.
\newblock {Open-Sora Plan: Open-Source Large Video Generation Model}.
\newblock \emph{arXiv preprint arXiv:2412.00131}, 2024{\natexlab{a}}.

\bibitem[Lin et~al.(2024{\natexlab{b}})Lin, Liu, Li, and Yang]{lin2024flawed}
Shanchuan Lin, Bingchen Liu, Jiashi Li, and Xiao Yang.
\newblock {Common Diffusion Noise Schedules and Sample Steps are Flawed}.
\newblock In \emph{Proceedings of the IEEE/CVF Winter Conference on Applications of Computer Vision (WACV)}, pages 5404--5411, 2024{\natexlab{b}}.

\bibitem[Lipman et~al.(2023)Lipman, Chen, Ben-Hamu, Nickel, and Le]{lipman2022flowmatching}
Yaron Lipman, Ricky T.~Q. Chen, Heli Ben-Hamu, Maximilian Nickel, and Matt Le.
\newblock {Flow Matching for Generative Modeling}.
\newblock In \emph{International Conference on Learning Representations (ICLR)}, 2023.

\bibitem[Liu et~al.(2026{\natexlab{a}})Liu, Ye, Xue, Gao, Li, Chen, Li, Yu, Wang, Zhang, Zhu, Bian, Han, and Xie]{liu2026solrefiner}
Haozhe Liu, Tian Ye, Shuchen Xue, Sensen Gao, Yitong Li, Junsong Chen, Haopeng Li, Jincheng Yu, Duomin Wang, Ruihua Zhang, Lei Zhu, Jia-Wang Bian, Song Han, and Enze Xie.
\newblock {SoL-Refiner: Speed-of-Light One-Step Refinement for High-Resolution Video}.
\newblock \emph{arXiv preprint arXiv:2609.37969}, 2026{\natexlab{a}}.

\bibitem[Liu et~al.(2025{\natexlab{a}})Liu, Liu, Liang, Yuan, Liu, Zheng, Wu, Wang, Xia, Wang, Liu, Yang, Wan, Zhang, Gai, Yang, and Ouyang]{liu2025videoalign}
Jie Liu, Gongye Liu, Jiajun Liang, Ziyang Yuan, Xiaokun Liu, Mingwu Zheng, Xiele Wu, Qiulin Wang, Menghan Xia, Xintao Wang, Xiaohong Liu, Fei Yang, Pengfei Wan, Di~Zhang, Kun Gai, Yujiu Yang, and Wanli Ouyang.
\newblock {Improving Video Generation with Human Feedback}.
\newblock In \emph{Advances in Neural Information Processing Systems (NeurIPS)}, 2025{\natexlab{a}}.

\bibitem[Liu et~al.(2026{\natexlab{b}})Liu, Li, Chen, Wu, Zheng, Ji, Zhou, Luo, Liu, Fei, and Chua]{liu2025javisdit}
Kai Liu, Wei Li, Lai Chen, Shengqiong Wu, Yanhao Zheng, Jiayi Ji, Fan Zhou, Jiebo Luo, Ziwei Liu, Hao Fei, and Tat-Seng Chua.
\newblock {JavisDiT: Joint Audio-Video Diffusion Transformer with Hierarchical Spatio-Temporal Prior Synchronization}.
\newblock In \emph{International Conference on Learning Representations (ICLR)}, 2026{\natexlab{b}}.

\bibitem[Liu et~al.(2025{\natexlab{b}})Liu, Ma, Li, Chen, Liu, Li, Zhou, He, and Wu]{liu2025phantom}
Lijie Liu, Tianxiang Ma, Bingchuan Li, Zhuowei Chen, Jiawei Liu, Gen Li, Siyu Zhou, Qian He, and Xinglong Wu.
\newblock {Phantom: Subject-Consistent Video Generation via Cross-Modal Alignment}.
\newblock In \emph{Proceedings of the IEEE/CVF International Conference on Computer Vision (ICCV)}, pages 14951--14961, 2025{\natexlab{b}}.

\bibitem[Liu et~al.(2023)Liu, Gong, and Liu]{liu2022rectifiedflow}
Xingchao Liu, Chengyue Gong, and Qiang Liu.
\newblock {Flow Straight and Fast: Learning to Generate and Transfer Data with Rectified Flow}.
\newblock In \emph{International Conference on Learning Representations (ICLR)}, 2023.

\bibitem[Loshchilov and Hutter(2019)]{loshchilov2019adamw}
Ilya Loshchilov and Frank Hutter.
\newblock {Decoupled Weight Decay Regularization}.
\newblock In \emph{International Conference on Learning Representations (ICLR)}, 2019.

\bibitem[Low et~al.(2025)Low, Wang, and Katyal]{low2025ovi}
Chetwin Low, Weimin Wang, and Calder Katyal.
\newblock {Ovi: Twin Backbone Cross-Modal Fusion for Audio-Video Generation}.
\newblock \emph{arXiv preprint arXiv:2510.01284}, 2025.

\bibitem[Lu and Song(2025)]{lu2024scm}
Cheng Lu and Yang Song.
\newblock {Simplifying, Stabilizing and Scaling Continuous-Time Consistency Models}.
\newblock In \emph{International Conference on Learning Representations (ICLR)}, 2025.

\bibitem[Lu et~al.(2022)Lu, Zhou, Bao, Chen, Li, and Zhu]{lu2022dpmsolver}
Cheng Lu, Yuhao Zhou, Fan Bao, Jianfei Chen, Chongxuan Li, and Jun Zhu.
\newblock {DPM-Solver: A Fast ODE Solver for Diffusion Probabilistic Model Sampling in Around 10 Steps}.
\newblock In \emph{Advances in Neural Information Processing Systems (NeurIPS)}, volume~35, pages 5775--5787, 2022.

\bibitem[Lu et~al.(2025{\natexlab{a}})Lu, Zhou, Bao, Chen, Li, and Zhu]{lu2022dpmsolverpp}
Cheng Lu, Yuhao Zhou, Fan Bao, Jianfei Chen, Chongxuan Li, and Jun Zhu.
\newblock {DPM-Solver++: Fast Solver for Guided Sampling of Diffusion Probabilistic Models}.
\newblock \emph{Machine Intelligence Research}, 22\penalty0 (4):\penalty0 730--751, 2025{\natexlab{a}}.
\newblock \doi{10.1007/s11633-025-1562-4}.

\bibitem[Lu et~al.(2025{\natexlab{b}})Lu, Ren, Xia, Lin, Wang, Xiao, Ma, Xie, and Lai]{lu2025adm}
Yanzuo Lu, Yuxi Ren, Xin Xia, Shanchuan Lin, Xing Wang, Xuefeng Xiao, Andy~J. Ma, Xiaohua Xie, and Jian-Huang Lai.
\newblock {Adversarial Distribution Matching for Diffusion Distillation Towards Efficient Image and Video Synthesis}.
\newblock In \emph{Proceedings of the IEEE/CVF International Conference on Computer Vision (ICCV)}, pages 16818--16829, 2025{\natexlab{b}}.

\bibitem[Lu et~al.(2026)Lu, Zeng, Li, Ouyang, Wang, Cheng, Zhu, Cao, Zhang, Zhu, Shen, and Zhang]{lu2025rewardforcing}
Yunhong Lu, Yanhong Zeng, Haobo Li, Hao Ouyang, Qiuyu Wang, Ka~Leong Cheng, Jiapeng Zhu, Hengyuan Cao, Zhipeng Zhang, Xing Zhu, Yujun Shen, and Min Zhang.
\newblock {Reward Forcing: Efficient Streaming Video Generation with Rewarded Distribution Matching Distillation}.
\newblock In \emph{Proceedings of the IEEE/CVF Conference on Computer Vision and Pattern Recognition (CVPR)}, pages 34385--34397, 2026.

\bibitem[Luo et~al.(2023)Luo, Hu, Zhang, Sun, Li, and Zhang]{luo2023diffinstruct}
Weijian Luo, Tianyang Hu, Shifeng Zhang, Jiacheng Sun, Zhenguo Li, and Zhihua Zhang.
\newblock {Diff-Instruct: A Universal Approach for Transferring Knowledge From Pre-trained Diffusion Models}.
\newblock In \emph{Advances in Neural Information Processing Systems (NeurIPS)}, volume~36, pages 76525--76546, 2023.

\bibitem[Luo et~al.(2026)Luo, Qian, Tang, Zhu, Liu, Wang, and You]{luo2026afd}
Yang Luo, Shengju Qian, Xiaohang Tang, Zirui Zhu, Yong Liu, Xin Wang, and Yang You.
\newblock {On-Policy Adversarial Flow Distillation for Autoregressive Video Generation}.
\newblock \emph{arXiv preprint arXiv:2605.26105}, 2026.

\bibitem[Ma et~al.(2025{\natexlab{a}})Ma, Huang, Yan, Chen, Duan, Yin, et~al.]{ma2025stepvideo}
Guoqing Ma, Haoyang Huang, Kun Yan, Liangyu Chen, Nan Duan, Shengming Yin, et~al.
\newblock {Step-Video-T2V Technical Report: The Practice, Challenges, and Future of Video Foundation Model}.
\newblock \emph{arXiv preprint arXiv:2502.10248}, 2025{\natexlab{a}}.

\bibitem[Ma et~al.(2025{\natexlab{b}})Ma, Wu, Sun, and Li]{ma2025hpsv3}
Yuhang Ma, Xiaoshi Wu, Keqiang Sun, and Hongsheng Li.
\newblock {HPSv3: Towards Wide-Spectrum Human Preference Score}.
\newblock In \emph{Proceedings of the IEEE/CVF International Conference on Computer Vision (ICCV)}, pages 15086--15095, 2025{\natexlab{b}}.

\bibitem[Martens(2020)]{martens2014ggn}
James Martens.
\newblock {New Insights and Perspectives on the Natural Gradient Method}.
\newblock \emph{Journal of Machine Learning Research}, 21\penalty0 (146):\penalty0 1--76, 2020.

\bibitem[{Meituan LongCat Team} et~al.(2025){Meituan LongCat Team}, Cai, Huang, Kang, Li, Liang, Ma, Ren, Wei, Xie, and Zhang]{longcat2025video}
{Meituan LongCat Team}, Xunliang Cai, Qilong Huang, Zhuoliang Kang, Hongyu Li, Shijun Liang, Liya Ma, Siyu Ren, Xiaoming Wei, Rixu Xie, and Tong Zhang.
\newblock {LongCat-Video Technical Report}.
\newblock \emph{arXiv preprint arXiv:2510.22200}, 2025.

\bibitem[Meulemans et~al.(2020)Meulemans, Carzaniga, Suykens, Sacramento, and Grewe]{meulemans2020tp}
Alexander Meulemans, Francesco~S. Carzaniga, Johan A.~K. Suykens, João Sacramento, and Benjamin~F. Grewe.
\newblock {A Theoretical Framework for Target Propagation}.
\newblock In \emph{Advances in Neural Information Processing Systems (NeurIPS)}, volume~33, pages 20024--20036, 2020.

\bibitem[Meulemans et~al.(2021)Meulemans, Farinha, Ordóñez, Aceituno, Sacramento, and Grewe]{meulemans2021dfc}
Alexander Meulemans, Matilde~Tristany Farinha, Javier~García Ordóñez, Pau~Vilimelis Aceituno, João Sacramento, and Benjamin~F. Grewe.
\newblock {Credit Assignment in Neural Networks through Deep Feedback Control}.
\newblock In \emph{Advances in Neural Information Processing Systems (NeurIPS)}, volume~34, pages 4674--4687, 2021.

\bibitem[{MiniMax}(2026{\natexlab{a}})]{minimax2026h3}
{MiniMax}.
\newblock {MiniMax-H3}.
\newblock \url{https://huggingface.co/MiniMaxAI/MiniMax-H3}, 2026{\natexlab{a}}.
\newblock Model card.

\bibitem[{MiniMax}(2026{\natexlab{b}})]{minimax2026h3skill}
{MiniMax}.
\newblock {h3-prompt-writing}.
\newblock \url{https://github.com/MiniMax-AI/MiniMax-H3/tree/main/skills/h3-prompt-writing}, 2026{\natexlab{b}}.
\newblock Prompt-writing skill for T2VA, I2VA, FL2VA, L2VA and Ref2VA.

\bibitem[Nguyen et~al.(2026)Nguyen, Nguyen, Vu, Dao, Nguyen, Dao, Nguyen, Tran, Nguyen, Nguyen, Pham, Metaxas, Patel, and Tran]{nguyen2026crossspace}
Anh Nguyen, Ngan Nguyen, Duc Vu, Trung Dao, Viet Nguyen, Quan Dao, Kien Nguyen, Chi Tran, Phong Nguyen, Khoi Nguyen, Cuong Pham, Dimitris Metaxas, Vishal~M. Patel, and Anh Tran.
\newblock {Cross-Space Distillation: Teaching One-Step Students with Modern Diffusion Teachers}.
\newblock In \emph{Computer Vision -- {ECCV} 2026}, Lecture Notes in Computer Science, pages 275--292, 2026.
\newblock \doi{10.1007/978-3-032-37450-9_15}.

\bibitem[Nie et~al.(2026)Nie, Berner, Ma, Liu, Xie, and Vahdat]{nie2026tmd}
Weili Nie, Julius Berner, Nanye Ma, Chao Liu, Saining Xie, and Arash Vahdat.
\newblock {Transition Matching Distillation for Fast Video Generation}.
\newblock In \emph{Proceedings of the IEEE/CVF Conference on Computer Vision and Pattern Recognition (CVPR)}, pages 4645--4655, 2026.

\bibitem[{NVIDIA} et~al.(2025){NVIDIA}, Agarwal, Ali, Bala, Balaji, Barker, et~al.]{nvidia2025cosmos}
{NVIDIA}, Niket Agarwal, Arslan Ali, Maciej Bala, Yogesh Balaji, Erik Barker, et~al.
\newblock {Cosmos World Foundation Model Platform for Physical AI}.
\newblock \emph{arXiv preprint arXiv:2501.03575}, 2025.

\bibitem[Nützel et~al.(2026)Nützel, Dombrowski, Hagen, and Kainz]{nutzel2026bilipschitz}
Felix Nützel, Mischa Dombrowski, Luca Hagen, and Bernhard Kainz.
\newblock {Semantic Latent Alignment for Flow Matching with Invertible Bi-Lipschitz Adapters}.
\newblock \emph{arXiv preprint arXiv:2610.21332}, 2026.

\bibitem[O'Leary-Roseberry et~al.(2024)O'Leary-Roseberry, Chen, Villa, and Ghattas]{oleary2022dino}
Thomas O'Leary-Roseberry, Peng Chen, Umberto Villa, and Omar Ghattas.
\newblock {Derivative-Informed Neural Operator: An efficient framework for high-dimensional parametric derivative learning}.
\newblock \emph{Journal of Computational Physics}, 496:\penalty0 112555, 2024.
\newblock \doi{10.1016/j.jcp.2023.112555}.

\bibitem[{OpenAI}(2024)]{openai2024sora}
{OpenAI}.
\newblock {Video Generation Models as World Simulators}.
\newblock \url{https://openai.com/research/video-generation-models-as-world-simulators}, 2024.
\newblock Technical report.

\bibitem[Pan et~al.(2024)Pan, Lu, Zhu, and Zhang]{pan2023pgc}
Zijie Pan, Jiachen Lu, Xiatian Zhu, and Li~Zhang.
\newblock {Enhancing High-Resolution 3D Generation through Pixel-wise Gradient Clipping}.
\newblock In \emph{International Conference on Learning Representations (ICLR)}, 2024.

\bibitem[Peebles and Xie(2023)]{peebles2023dit}
William Peebles and Saining Xie.
\newblock {Scalable Diffusion Models with Transformers}.
\newblock In \emph{Proceedings of the IEEE/CVF International Conference on Computer Vision (ICCV)}, pages 4195--4205, 2023.

\bibitem[Polyak et~al.(2024)Polyak, Zohar, Brown, Tjandra, Sinha, Lee, et~al.]{polyak2024moviegen}
Adam Polyak, Amit Zohar, Andrew Brown, Andros Tjandra, Animesh Sinha, Ann Lee, et~al.
\newblock {Movie Gen: A Cast of Media Foundation Models}.
\newblock \emph{arXiv preprint arXiv:2410.13720}, 2024.

\bibitem[Poole et~al.(2023)Poole, Jain, Barron, and Mildenhall]{poole2022dreamfusion}
Ben Poole, Ajay Jain, Jonathan~T. Barron, and Ben Mildenhall.
\newblock {DreamFusion: Text-to-3D using 2D Diffusion}.
\newblock In \emph{International Conference on Learning Representations (ICLR)}, 2023.

\bibitem[{Qwen Team}(2025)]{qwen3next}
{Qwen Team}.
\newblock {Qwen3-Next: Towards Ultimate Training \& Inference Efficiency}.
\newblock Qwen Team blog, \url{https://qwen.ai/blog?id=qwen3-next}, 2025.

\bibitem[Richter-Powell et~al.(2025)Richter-Powell, Torralba, and Lorraine]{richterpowell2025audiosds}
Jessie Richter-Powell, Antonio Torralba, and Jonathan Lorraine.
\newblock {Score Distillation Sampling for Audio: Source Separation, Synthesis, and Beyond}.
\newblock \emph{arXiv preprint arXiv:2505.04621}, 2025.

\bibitem[Rouhani et~al.(2023)Rouhani, Zhao, More, Hall, Khodamoradi, Deng, et~al.]{rouhani2023mx}
Bita~Darvish Rouhani, Ritchie Zhao, Ankit More, Mathew Hall, Alireza Khodamoradi, Summer Deng, et~al.
\newblock {Microscaling Data Formats for Deep Learning}.
\newblock \emph{arXiv preprint arXiv:2310.10537}, 2023.

\bibitem[Roulet and Harchaoui(2023)]{roulet2021tp}
Vincent Roulet and Zaid Harchaoui.
\newblock {Target Propagation via Regularized Inversion for Recurrent Neural Networks}.
\newblock \emph{Transactions on Machine Learning Research}, 2023.

\bibitem[Ruan et~al.(2023)Ruan, Ma, Yang, He, Liu, Fu, Yuan, Jin, and Guo]{ruan2023mmdiffusion}
Ludan Ruan, Yiyang Ma, Huan Yang, Huiguo He, Bei Liu, Jianlong Fu, Nicholas~Jing Yuan, Qin Jin, and Baining Guo.
\newblock {MM-Diffusion: Learning Multi-Modal Diffusion Models for Joint Audio and Video Generation}.
\newblock In \emph{Proceedings of the IEEE/CVF Conference on Computer Vision and Pattern Recognition (CVPR)}, pages 10219--10228, 2023.

\bibitem[Salimans and Ho(2022)]{salimans2022progressive}
Tim Salimans and Jonathan Ho.
\newblock {Progressive Distillation for Fast Sampling of Diffusion Models}.
\newblock In \emph{International Conference on Learning Representations (ICLR)}, 2022.

\bibitem[{Sand.ai} et~al.(2025){Sand.ai}, Teng, Jia, Sun, Li, Li, et~al.]{sandai2025magi1}
{Sand.ai}, Hansi Teng, Hongyu Jia, Lei Sun, Lingzhi Li, Maolin Li, et~al.
\newblock {MAGI-1: Autoregressive Video Generation at Scale}.
\newblock \emph{arXiv preprint arXiv:2505.13211}, 2025.

\bibitem[Sauer et~al.(2024{\natexlab{a}})Sauer, Boesel, Dockhorn, Blattmann, Esser, and Rombach]{sauer2024ladd}
Axel Sauer, Frederic Boesel, Tim Dockhorn, Andreas Blattmann, Patrick Esser, and Robin Rombach.
\newblock {Fast High-Resolution Image Synthesis with Latent Adversarial Diffusion Distillation}.
\newblock In \emph{SIGGRAPH Asia 2024 Conference Papers}, pages 1--11, 2024{\natexlab{a}}.
\newblock \doi{10.1145/3680528.3687625}.

\bibitem[Sauer et~al.(2024{\natexlab{b}})Sauer, Lorenz, Blattmann, and Rombach]{sauer2023add}
Axel Sauer, Dominik Lorenz, Andreas Blattmann, and Robin Rombach.
\newblock {Adversarial Diffusion Distillation}.
\newblock In \emph{Computer Vision -- {ECCV} 2024}, Lecture Notes in Computer Science, pages 87--103, 2024{\natexlab{b}}.
\newblock \doi{10.1007/978-3-031-73016-0_6}.

\bibitem[Savani et~al.(2024)Savani, Finzi, and Kolter]{savani2024ddrep}
Yash Savani, Marc Finzi, and J.~Zico Kolter.
\newblock {Diffusing Differentiable Representations}.
\newblock In \emph{Advances in Neural Information Processing Systems (NeurIPS)}, volume~37, pages 44419--44443, 2024.

\bibitem[Singer et~al.(2023)Singer, Polyak, Hayes, Yin, An, Zhang, Hu, Yang, Ashual, Gafni, Parikh, Gupta, and Taigman]{singer2022makeavideo}
Uriel Singer, Adam Polyak, Thomas Hayes, Xi~Yin, Jie An, Songyang Zhang, Qiyuan Hu, Harry Yang, Oron Ashual, Oran Gafni, Devi Parikh, Sonal Gupta, and Yaniv Taigman.
\newblock {Make-A-Video: Text-to-Video Generation without Text-Video Data}.
\newblock In \emph{International Conference on Learning Representations (ICLR)}, 2023.

\bibitem[Skorokhodov(2026)]{skorokhodov2026pixsds}
Vsevolod Skorokhodov.
\newblock {PixSDS: Why Latent SDS Makes Noisy Pixels}.
\newblock \emph{arXiv preprint arXiv:2608.12997}, 2026.

\bibitem[Song et~al.(2021)Song, Meng, and Ermon]{song2021ddim}
Jiaming Song, Chenlin Meng, and Stefano Ermon.
\newblock {Denoising Diffusion Implicit Models}.
\newblock In \emph{International Conference on Learning Representations (ICLR)}, 2021.

\bibitem[Song et~al.(2023)Song, Dhariwal, Chen, and Sutskever]{song2023cm}
Yang Song, Prafulla Dhariwal, Mark Chen, and Ilya Sutskever.
\newblock {Consistency Models}.
\newblock In \emph{Proceedings of the 40th International Conference on Machine Learning (ICML)}, volume 202 of \emph{Proceedings of Machine Learning Research}, pages 32211--32252, 2023.

\bibitem[Su et~al.(2024)Su, Ahmed, Lu, Pan, Bo, and Liu]{su2021roformer}
Jianlin Su, Murtadha Ahmed, Yu~Lu, Shengfeng Pan, Wen Bo, and Yunfeng Liu.
\newblock {RoFormer: Enhanced Transformer with Rotary Position Embedding}.
\newblock \emph{Neurocomputing}, 568:\penalty0 127063, 2024.
\newblock \doi{10.1016/j.neucom.2023.127063}.

\bibitem[{Team Seawead} et~al.(2025){Team Seawead}, Yang, Lin, Zhao, Lin, Ma, et~al.]{seawead2025seaweed}
{Team Seawead}, Ceyuan Yang, Zhijie Lin, Yang Zhao, Shanchuan Lin, Zhibei Ma, et~al.
\newblock {Seaweed-7B: Cost-Effective Training of Video Generation Foundation Model}.
\newblock \emph{arXiv preprint arXiv:2504.08685}, 2025.

\bibitem[{Team Seedance} et~al.(2026){Team Seedance}, Chen, Chen, Chen, Chen, Chen, et~al.]{seedance2026seedance20advancingvideo}
{Team Seedance}, De~Chen, Liyang Chen, Xin Chen, Ying Chen, Zhuo Chen, et~al.
\newblock {Seedance 2.0: Advancing Video Generation for World Complexity}.
\newblock \emph{arXiv preprint arXiv:2604.14148}, 2026.

\bibitem[Tjandra et~al.(2025)Tjandra, Wu, Guo, Hoffman, Ellis, Vyas, Shi, Chen, Le, Zacharov, Wood, Lee, and Hsu]{tjandra2025audiobox}
Andros Tjandra, Yi-Chiao Wu, Baishan Guo, John Hoffman, Brian Ellis, Apoorv Vyas, Bowen Shi, Sanyuan Chen, Matt Le, Nick Zacharov, Carleigh Wood, Ann Lee, and Wei-Ning Hsu.
\newblock {Meta Audiobox Aesthetics: Unified Automatic Quality Assessment for Speech, Music, and Sound}.
\newblock \emph{arXiv preprint arXiv:2502.05139}, 2025.

\bibitem[Wallace et~al.(2024)Wallace, Dang, Rafailov, Zhou, Lou, Purushwalkam, Ermon, Xiong, Joty, and Naik]{wallace2023diffusiondpo}
Bram Wallace, Meihua Dang, Rafael Rafailov, Linqi Zhou, Aaron Lou, Senthil Purushwalkam, Stefano Ermon, Caiming Xiong, Shafiq Joty, and Nikhil Naik.
\newblock {Diffusion Model Alignment Using Direct Preference Optimization}.
\newblock In \emph{Proceedings of the IEEE/CVF Conference on Computer Vision and Pattern Recognition (CVPR)}, pages 8228--8238, 2024.

\bibitem[Wan et~al.(2025)Wan, Wang, Ai, Wen, Mao, Xie, et~al.]{wan2025wan}
Team Wan, Ang Wang, Baole Ai, Bin Wen, Chaojie Mao, Chen-Wei Xie, et~al.
\newblock {Wan: Open and Advanced Large-Scale Video Generative Models}.
\newblock \emph{arXiv preprint arXiv:2503.20314}, 2025.

\bibitem[Wang et~al.(2025{\natexlab{a}})Wang, Zuo, Li, Chen, Liao, Zhou, Yin, Dai, Jiang, and Yu]{wang2025universe1}
Duomin Wang, Wei Zuo, Aojie Li, Ling-Hao Chen, Xinyao Liao, Deyu Zhou, Zixin Yin, Xili Dai, Daxin Jiang, and Gang Yu.
\newblock {UniVerse-1: Unified Audio-Video Generation via Stitching of Experts}.
\newblock \emph{arXiv preprint arXiv:2509.06155}, 2025{\natexlab{a}}.

\bibitem[Wang et~al.(2023{\natexlab{a}})Wang, Du, Li, Yeh, and Shakhnarovich]{wang2022sjc}
Haochen Wang, Xiaodan Du, Jiahao Li, Raymond~A. Yeh, and Greg Shakhnarovich.
\newblock {Score Jacobian Chaining: Lifting Pretrained 2D Diffusion Models for 3D Generation}.
\newblock In \emph{Proceedings of the IEEE/CVF Conference on Computer Vision and Pattern Recognition (CVPR)}, pages 12619--12629, 2023{\natexlab{a}}.

\bibitem[Wang et~al.(2025{\natexlab{b}})Wang, Ma, Liu, Hou, Xu, Wang, Juefei-Xu, Luo, Zhang, Hou, Vajda, Jha, and Dai]{wang2024lingen}
Hongjie Wang, Chih-Yao Ma, Yen-Cheng Liu, Ji~Hou, Tao Xu, Jialiang Wang, Felix Juefei-Xu, Yaqiao Luo, Peizhao Zhang, Tingbo Hou, Peter Vajda, Niraj~K. Jha, and Xiaoliang Dai.
\newblock {LinGen: Towards High-Resolution Minute-Length Text-to-Video Generation with Linear Computational Complexity}.
\newblock In \emph{Proceedings of the IEEE/CVF Conference on Computer Vision and Pattern Recognition (CVPR)}, pages 2578--2588, 2025{\natexlab{b}}.

\bibitem[Wang et~al.(2023{\natexlab{b}})Wang, Lu, Wang, Bao, Li, Su, and Zhu]{wang2023prolificdreamer}
Zhengyi Wang, Cheng Lu, Yikai Wang, Fan Bao, Chongxuan Li, Hang Su, and Jun Zhu.
\newblock {ProlificDreamer: High-Fidelity and Diverse Text-to-3D Generation with Variational Score Distillation}.
\newblock In \emph{Advances in Neural Information Processing Systems (NeurIPS)}, volume~36, pages 8406--8441, 2023{\natexlab{b}}.

\bibitem[Wang et~al.(2026)Wang, Yuan, Wang, Yang, Zhang, Yuan, Huang, Liu, Wang, Yang, Ma, and Qian]{wang2026pdmd}
Zimo Wang, Junkun Yuan, Angtian Wang, Haotian Yang, Canyu Zhang, Siyuan Yuan, Xingchang Huang, Bo~Liu, Yizhi Wang, Yiding Yang, Chongyang Ma, and Gordon~Guocheng Qian.
\newblock {PDMD: Projected Distribution Matching Distillation for Video Diffusion Models}.
\newblock \emph{arXiv preprint arXiv:2609.35768}, 2026.

\bibitem[Wiedemer et~al.(2025)Wiedemer, Li, Vicol, Gu, Matarese, Swersky, Kim, Jaini, and Geirhos]{wiedemer2025veo3}
Thadd{\"a}us Wiedemer, Yuxuan Li, Paul Vicol, Shixiang~Shane Gu, Nick Matarese, Kevin Swersky, Been Kim, Priyank Jaini, and Robert Geirhos.
\newblock {Video models are zero-shot learners and reasoners}.
\newblock \emph{arXiv preprint arXiv:2509.20328}, 2025.

\bibitem[Wu et~al.(2025{\natexlab{a}})Wu, Zou, Li, Huang, Yang, Tan, et~al.]{wu2025hunyuanvideo15}
Bing Wu, Chang Zou, Changlin Li, Duojun Huang, Fang Yang, Hao Tan, et~al.
\newblock {HunyuanVideo 1.5 Technical Report}.
\newblock \emph{arXiv preprint arXiv:2511.18870}, 2025{\natexlab{a}}.

\bibitem[Wu et~al.(2026{\natexlab{a}})Wu, Xie, and Lv]{wu2026hdps}
Hongjie Wu, Yiping Xie, and Jiancheng Lv.
\newblock {Hybrid-Domain Posterior Sampling for Inverse Problems via Latent Flow Matching}.
\newblock In \emph{Proceedings of the 34th ACM International Conference on Multimedia (MM)}, pages 11526--11535, 2026{\natexlab{a}}.
\newblock \doi{10.1145/3767308.3835243}.

\bibitem[Wu et~al.(2026{\natexlab{b}})Wu, Li, Zhang, and Ma]{wu2026dpdmd}
Tianhe Wu, Ruibin Li, Lei Zhang, and Kede Ma.
\newblock {Diversity-Preserved Distribution Matching Distillation for Fast Visual Synthesis}.
\newblock In \emph{Proceedings of the 43rd International Conference on Machine Learning (ICML)}, volume 306 of \emph{Proceedings of Machine Learning Research}, pages 136351--136369, 2026{\natexlab{b}}.

\bibitem[Wu et~al.(2023)Wu, Chen, Zhang, Hui, Berg-Kirkpatrick, and Dubnov]{wu2023clap}
Yusong Wu, Ke~Chen, Tianyu Zhang, Yuchen Hui, Taylor Berg-Kirkpatrick, and Shlomo Dubnov.
\newblock {Large-Scale Contrastive Language-Audio Pretraining with Feature Fusion and Keyword-to-Caption Augmentation}.
\newblock In \emph{IEEE International Conference on Acoustics, Speech and Signal Processing (ICASSP)}, pages 1--5, 2023.
\newblock \doi{10.1109/ICASSP49357.2023.10095969}.

\bibitem[Wu et~al.(2026{\natexlab{c}})Wu, Gong, Yong, Huang, Fan, Yang, Lin, and Liu]{wu2026sgmd}
Zhuguanyu Wu, Ruihao Gong, Yang Yong, Yushi Huang, Xiangyu Fan, Lei Yang, Dahua Lin, and Xianglong Liu.
\newblock {SGMD: Score Gradient Matching Distillation for Few-Step Video Diffusion Distillation}.
\newblock In \emph{Proceedings of the 43rd International Conference on Machine Learning (ICML)}, volume 306 of \emph{Proceedings of Machine Learning Research}, pages 135898--135912, 2026{\natexlab{c}}.

\bibitem[Wu et~al.(2025{\natexlab{b}})Wu, Kag, Skorokhodov, Menapace, Mirzaei, Gilitschenski, Tulyakov, and Siarohin]{wu2025densedpo}
Ziyi Wu, Anil Kag, Ivan Skorokhodov, Willi Menapace, Ashkan Mirzaei, Igor Gilitschenski, Sergey Tulyakov, and Aliaksandr Siarohin.
\newblock {DenseDPO: Fine-Grained Temporal Preference Optimization for Video Diffusion Models}.
\newblock In \emph{Advances in Neural Information Processing Systems (NeurIPS)}, 2025{\natexlab{b}}.

\bibitem[Xi et~al.(2026)Xi, Xie, Zhao, Zhang, Liu, Creavin, Keutzer, Li, Lv, Xu, and Feng]{xi2026videodeltanet}
Haocheng Xi, Yiming Xie, Hexu Zhao, Yiwen Zhang, Michael Liu, Thomas Creavin, Kurt Keutzer, Xiuyu Li, Zhaoyang Lv, Chenfeng Xu, and Haiwen Feng.
\newblock {Video DeltaNet: A Video-Native Hybrid Attention for Livestream Video Generation}.
\newblock \emph{arXiv preprint arXiv:2609.20744}, 2026.

\bibitem[Xie et~al.(2025)Xie, Chen, Chen, Cai, Tang, Lin, Zhang, Li, Zhu, Lu, and Han]{xie2024sana}
Enze Xie, Junsong Chen, Junyu Chen, Han Cai, Haotian Tang, Yujun Lin, Zhekai Zhang, Muyang Li, Ligeng Zhu, Yao Lu, and Song Han.
\newblock {SANA: Efficient High-Resolution Text-to-Image Synthesis with Linear Diffusion Transformers}.
\newblock In \emph{International Conference on Learning Representations (ICLR)}, 2025.

\bibitem[Xing et~al.(2024)Xing, He, Tian, Wang, and Chen]{xing2024seeinghearing}
Yazhou Xing, Yingqing He, Zeyue Tian, Xintao Wang, and Qifeng Chen.
\newblock {Seeing and Hearing: Open-domain Visual-Audio Generation with Diffusion Latent Aligners}.
\newblock In \emph{Proceedings of the IEEE/CVF Conference on Computer Vision and Pattern Recognition (CVPR)}, pages 7151--7161, 2024.

\bibitem[Xu et~al.(2026)Xu, Huang, Lu, Zhong, Fan, Huang, and Liu]{xu2026dymd}
Haojun Xu, Jie Huang, Xin Lu, Mingchen Zhong, Zihao Fan, Linjiang Huang, and Si~Liu.
\newblock {DyMD: Preserving Interaction Dynamics through Distribution Matching Distillation in Few-Step Video World Models}.
\newblock \emph{arXiv preprint arXiv:2609.31349}, 2026.

\bibitem[Xu et~al.(2023)Xu, Liu, Wu, Tong, Li, Ding, Tang, and Dong]{xu2023imagereward}
Jiazheng Xu, Xiao Liu, Yuchen Wu, Yuxuan Tong, Qinkai Li, Ming Ding, Jie Tang, and Yuxiao Dong.
\newblock {ImageReward: Learning and Evaluating Human Preferences for Text-to-Image Generation}.
\newblock In \emph{Advances in Neural Information Processing Systems (NeurIPS)}, volume~36, pages 15903--15935, 2023.

\bibitem[Xue et~al.(2025)Xue, Chen, Wang, Huang, Shechtman, Xie, and Chen]{xue2025mogan}
Haotian Xue, Qi~Chen, Zhonghao Wang, Xun Huang, Eli Shechtman, Jinrong Xie, and Yongxin Chen.
\newblock {MoGAN: Improving Motion Quality in Video Diffusion via Few-Step Motion Adversarial Post-Training}.
\newblock \emph{arXiv preprint arXiv:2511.21592}, 2025.

\bibitem[Xue et~al.(2026)Xue, Xiong, Ma, and Zhang]{xue2026fpopd}
Leyan Xue, Feng Xiong, Mingjun Ma, and Changqing Zhang.
\newblock {Distill What the Student Can See: Fisher-Projected On-Policy Distillation for Vision-Language Models}.
\newblock \emph{arXiv preprint arXiv:2608.01263}, 2026.

\bibitem[Xue et~al.(2023)Xue, Yi, Luo, Zhang, Sun, Li, and Ma]{xue2023sasolver}
Shuchen Xue, Mingyang Yi, Weijian Luo, Shifeng Zhang, Jiacheng Sun, Zhenguo Li, and Zhi-Ming Ma.
\newblock {SA-Solver: Stochastic Adams Solver for Fast Sampling of Diffusion Models}.
\newblock In \emph{Advances in Neural Information Processing Systems (NeurIPS)}, volume~36, pages 77632--77674, 2023.

\bibitem[Yang et~al.(2025{\natexlab{a}})Yang, Huang, Chu, Xiao, Zhao, Wang, Li, Xie, Chen, Lu, Han, and Chen]{yang2025longlive}
Shuai Yang, Wei Huang, Ruihang Chu, Yicheng Xiao, Yuyang Zhao, Xianbang Wang, Muyang Li, Enze Xie, Yingcong Chen, Yao Lu, Song Han, and Yukang Chen.
\newblock {LongLive: Real-time Interactive Long Video Generation}.
\newblock \emph{arXiv preprint arXiv:2509.22622}, 2025{\natexlab{a}}.

\bibitem[Yang et~al.(2025{\natexlab{b}})Yang, Kautz, and Hatamizadeh]{yang2025gdn}
Songlin Yang, Jan Kautz, and Ali Hatamizadeh.
\newblock {Gated Delta Networks: Improving Mamba2 with Delta Rule}.
\newblock In \emph{International Conference on Learning Representations (ICLR)}, 2025{\natexlab{b}}.

\bibitem[Yang et~al.(2025{\natexlab{c}})Yang, Teng, Zheng, Ding, Huang, Xu, Yang, Hong, Zhang, Feng, Yin, Zhang, Wang, Cheng, Xu, Gu, Dong, and Tang]{yang2024cogvideox}
Zhuoyi Yang, Jiayan Teng, Wendi Zheng, Ming Ding, Shiyu Huang, Jiazheng Xu, Yuanming Yang, Wenyi Hong, Xiaohan Zhang, Guanyu Feng, Da~Yin, Yuxuan Zhang, Weihan Wang, Yean Cheng, Bin Xu, Xiaotao Gu, Yuxiao Dong, and Jie Tang.
\newblock {CogVideoX: Text-to-Video Diffusion Models with An Expert Transformer}.
\newblock In \emph{International Conference on Learning Representations (ICLR)}, 2025{\natexlab{c}}.

\bibitem[Ye et~al.(2023)Ye, Zhang, Liu, Han, and Yang]{ye2023ipadapter}
Hu~Ye, Jun Zhang, Sibo Liu, Xiao Han, and Wei Yang.
\newblock {IP-Adapter: Text Compatible Image Prompt Adapter for Text-to-Image Diffusion Models}.
\newblock \emph{arXiv preprint arXiv:2308.06721}, 2023.

\bibitem[Yin et~al.(2024{\natexlab{a}})Yin, Gharbi, Park, Zhang, Shechtman, Durand, and Freeman]{yin2024dmd2}
Tianwei Yin, Micha{\"e}l Gharbi, Taesung Park, Richard Zhang, Eli Shechtman, Fr{\'e}do Durand, and William~T. Freeman.
\newblock {Improved Distribution Matching Distillation for Fast Image Synthesis}.
\newblock In \emph{Advances in Neural Information Processing Systems (NeurIPS)}, volume~37, pages 47455--47487, 2024{\natexlab{a}}.

\bibitem[Yin et~al.(2024{\natexlab{b}})Yin, Gharbi, Zhang, Shechtman, Durand, Freeman, and Park]{yin2023dmd}
Tianwei Yin, Micha{\"e}l Gharbi, Richard Zhang, Eli Shechtman, Fr{\'e}do Durand, William~T. Freeman, and Taesung Park.
\newblock {One-step Diffusion with Distribution Matching Distillation}.
\newblock In \emph{Proceedings of the IEEE/CVF Conference on Computer Vision and Pattern Recognition (CVPR)}, pages 6613--6623, 2024{\natexlab{b}}.

\bibitem[Yin et~al.(2025)Yin, Zhang, Zhang, Freeman, Durand, Shechtman, and Huang]{yin2024causvid}
Tianwei Yin, Qiang Zhang, Richard Zhang, William~T. Freeman, Fredo Durand, Eli Shechtman, and Xun Huang.
\newblock {From Slow Bidirectional to Fast Autoregressive Video Diffusion Models}.
\newblock In \emph{Proceedings of the IEEE/CVF Conference on Computer Vision and Pattern Recognition (CVPR)}, pages 22963--22974, 2025.

\bibitem[You et~al.(2026)You, Li, Li, Mu, Chen, and Jiang]{you2026avd}
Yuyang You, Yongzhi Li, Jiahui Li, Yadong Mu, Quan Chen, and Peng Jiang.
\newblock {Adaptive Video Distillation: Mitigating Oversaturation and Temporal Collapse in Few-Step Generation}.
\newblock In \emph{Proceedings of the IEEE/CVF Conference on Computer Vision and Pattern Recognition (CVPR)}, pages 43429--43439, 2026.

\bibitem[Yu et~al.(2026)Yu, Yuan, Yang, Qian, Wang, Wang, Yang, Liu, Li, Wang, and Ma]{yu2026dmad}
Zhengming Yu, Junkun Yuan, Haotian Yang, Gordon~Guocheng Qian, Yizhi Wang, Angtian Wang, Yiding Yang, Bo~Liu, Xin Li, Wenping Wang, and Chongyang Ma.
\newblock {DMAD: Distribution Matching as Adversarial Distillation for Fast Visual Generation}.
\newblock \emph{arXiv preprint arXiv:2610.02188}, 2026.

\bibitem[Yuan et~al.(2025)Yuan, Huang, He, Ge, Shi, Chen, Luo, and Yuan]{yuan2025consisid}
Shenghai Yuan, Jinfa Huang, Xianyi He, Yunyang Ge, Yujun Shi, Liuhan Chen, Jiebo Luo, and Li~Yuan.
\newblock {Identity-Preserving Text-to-Video Generation by Frequency Decomposition}.
\newblock In \emph{Proceedings of the IEEE/CVF Conference on Computer Vision and Pattern Recognition (CVPR)}, pages 12978--12988, 2025.

\bibitem[Zhan et~al.(2026)Zhan, Wang, Ma, Xing, Yao, Liu, Zhao, and Xue]{zhan2026duomatching}
Jiahao Zhan, Yan Wang, Yongrui Ma, Qunliang Xing, Ruchang Yao, Runtao Liu, Shijie Zhao, and Tianfan Xue.
\newblock {DuoMatching: Joint-Marginal Distribution Matching for Few-Step Video Generation}.
\newblock \emph{arXiv preprint arXiv:2610.03543}, 2026.

\bibitem[Zhang et~al.(2026{\natexlab{a}})Zhang, Chen, Wang, Liu, Wang, and Qiao]{zhang2025accvideo}
Haiyu Zhang, Xinyuan Chen, Yaohui Wang, Xihui Liu, Yunhong Wang, and Yu~Qiao.
\newblock {AccVideo: Accelerating Video Diffusion Model with Synthetic Dataset}.
\newblock \emph{Transactions on Machine Learning Research}, 2026{\natexlab{a}}.

\bibitem[Zhang et~al.(2026{\natexlab{b}})Zhang, Cheng, Wang, Li, Zhang, Li, Zha, Zhang, Zhao, and Jia]{zhang2026codmd}
Wenhu Zhang, Kun Cheng, Changyuan Wang, Shiyao Li, Yuechen Zhang, Wenbo Li, Jiajun Zha, Jingyi Zhang, Kang Zhao, and Jiaya Jia.
\newblock {CoDMD: Copula-aware Distribution Matching Distillation for Fast Video Generation}.
\newblock In \emph{Advances in Neural Information Processing Systems (NeurIPS)}, 2026{\natexlab{b}}.

\bibitem[Zhao et~al.(2023{\natexlab{a}})Zhao, Bai, Rao, Zhou, and Lu]{zhao2023unipc}
Wenliang Zhao, Lujia Bai, Yongming Rao, Jie Zhou, and Jiwen Lu.
\newblock {UniPC: A Unified Predictor-Corrector Framework for Fast Sampling of Diffusion Models}.
\newblock In \emph{Advances in Neural Information Processing Systems (NeurIPS)}, volume~36, pages 49842--49869, 2023{\natexlab{a}}.

\bibitem[Zhao et~al.(2023{\natexlab{b}})Zhao, Gu, Varma, Luo, Huang, Xu, Wright, Shojanazeri, Ott, Shleifer, Desmaison, Balioglu, Damania, Nguyen, Chauhan, Hao, Mathews, and Li]{zhao2023fsdp}
Yanli Zhao, Andrew Gu, Rohan Varma, Liang Luo, Chien-Chin Huang, Min Xu, Less Wright, Hamid Shojanazeri, Myle Ott, Sam Shleifer, Alban Desmaison, Can Balioglu, Pritam Damania, Bernard Nguyen, Geeta Chauhan, Yuchen Hao, Ajit Mathews, and Shen Li.
\newblock {PyTorch FSDP: Experiences on Scaling Fully Sharded Data Parallel}.
\newblock \emph{Proceedings of the VLDB Endowment}, 16\penalty0 (12):\penalty0 3848--3860, 2023{\natexlab{b}}.
\newblock \doi{10.14778/3611540.3611569}.

\bibitem[Zheng et~al.(2025{\natexlab{a}})Zheng, Huang, Liu, Zou, He, Zhang, Gu, Zhang, He, Zheng, Qiao, and Liu]{zheng2025vbench2}
Dian Zheng, Ziqi Huang, Hongbo Liu, Kai Zou, Yinan He, Fan Zhang, Lulu Gu, Yuanhan Zhang, Jingwen He, Wei-Shi Zheng, Yu~Qiao, and Ziwei Liu.
\newblock {VBench-2.0: Advancing Video Generation Benchmark Suite for Intrinsic Faithfulness}.
\newblock \emph{arXiv preprint arXiv:2503.21755}, 2025{\natexlab{a}}.

\bibitem[Zheng et~al.(2026)Zheng, Wang, Ma, Chen, Zhang, Balaji, Chen, Liu, Zhu, and Zhang]{zheng2025rcm}
Kaiwen Zheng, Yuji Wang, Qianli Ma, Huayu Chen, Jintao Zhang, Yogesh Balaji, Jianfei Chen, Ming-Yu Liu, Jun Zhu, and Qinsheng Zhang.
\newblock {Large Scale Diffusion Distillation via Score-Regularized Continuous-Time Consistency}.
\newblock In \emph{International Conference on Learning Representations (ICLR)}, 2026.

\bibitem[Zheng et~al.(2024)Zheng, Peng, Yang, Shen, Li, Liu, Zhou, Li, and You]{zheng2024opensora}
Zangwei Zheng, Xiangyu Peng, Tianji Yang, Chenhui Shen, Shenggui Li, Hongxin Liu, Yukun Zhou, Tianyi Li, and Yang You.
\newblock {Open-Sora: Democratizing Efficient Video Production for All}.
\newblock \emph{arXiv preprint arXiv:2412.20404}, 2024.

\bibitem[Zheng et~al.(2025{\natexlab{b}})Zheng, Peng, Lou, Shen, Young, Guo, et~al.]{peng2025opensora2}
Zangwei Zheng, Xiangyu Peng, Yuxuan Lou, Chenhui Shen, Tom Young, Xinying Guo, et~al.
\newblock {Open-Sora 2.0: Training a Commercial-Level Video Generation Model in \$200k}.
\newblock \emph{arXiv preprint arXiv:2503.09642}, 2025{\natexlab{b}}.

\bibitem[Zhou et~al.(2024)Zhou, Zheng, Wang, Yin, and Huang]{zhou2024sid}
Mingyuan Zhou, Huangjie Zheng, Zhendong Wang, Mingzhang Yin, and Hai Huang.
\newblock {Score identity Distillation: Exponentially Fast Distillation of Pretrained Diffusion Models for One-Step Generation}.
\newblock In \emph{Proceedings of the 41st International Conference on Machine Learning (ICML)}, volume 235 of \emph{Proceedings of Machine Learning Research}, pages 62307--62331, 2024.

\bibitem[Zhu et~al.(2026)Zhu, Liu, Zhao, Ye, Chen, Yu, He, Han, and Xie]{zhu2026sanawm}
Haoyi Zhu, Haozhe Liu, Yuyang Zhao, Tian Ye, Junsong Chen, Jincheng Yu, Tong He, Song Han, and Enze Xie.
\newblock {SANA-WM: Efficient Minute-Scale World Modeling with Hybrid Linear Diffusion Transformer}.
\newblock \emph{arXiv preprint arXiv:2605.15178}, 2026.

\bibitem[Zou et~al.(2026)Zou, Zheng, Liu, Hang, Liu, and Yu]{zou2026hiar}
Kai Zou, Dian Zheng, Hongbo Liu, Tiankai Hang, Bin Liu, and Nenghai Yu.
\newblock {HiAR: Efficient Autoregressive Long Video Generation via Hierarchical Denoising}.
\newblock In \emph{Computer Vision -- {ECCV} 2026}, Lecture Notes in Computer Science, pages 225--242, 2026.
\newblock \doi{10.1007/978-3-032-37026-6_13}.

\end{thebibliography}
\end{document}